\documentclass{article}

\usepackage{microtype}
\usepackage{graphicx}
\usepackage{subcaption}
\usepackage{booktabs} 
\usepackage{diagbox}
\usepackage{booktabs}
\usepackage{multirow}
\usepackage{hyperref}

\usepackage[accepted]{icml2026}

\usepackage{amsmath}
\usepackage{amsthm}
\usepackage{amssymb}
\usepackage{mathtools}
\usepackage{amsthm}
\usepackage{amsfonts}
\usepackage{nicefrac}
\usepackage{xcolor}
\usepackage{algorithm}
\usepackage{algorithmic} 
\usepackage{enumerate}
\usepackage{comment}
\usepackage{thmtools}
\usepackage{thm-restate}

\usepackage[capitalize,noabbrev]{cleveref}

\theoremstyle{plain}
\newtheorem{theorem}{Theorem}[section]
\newtheorem{proposition}[theorem]{Proposition}

\theoremstyle{definition}

\theoremstyle{remark}
\newtheorem{remark}[theorem]{Remark}

\newcommand{\E}{\mathbb{E}}

\icmltitlerunning{Cost-aware Stopping for Bayesian Optimization}

\begin{document}

\twocolumn[
  \icmltitle{Cost-aware Stopping for Bayesian Optimization}


  \icmlsetsymbol{equal}{*}

  \begin{icmlauthorlist}
    \icmlauthor{Qian Xie}{equal,cornell}
    \icmlauthor{Linda Cai}{equal,ucb}
    \icmlauthor{Alexander Terenin}{cornell}
    \icmlauthor{Peter I. Frazier}{cornell}
    \icmlauthor{Ziv Scully}{cornell}
  \end{icmlauthorlist}

  \icmlaffiliation{cornell}{Cornell University, Ithaca, New York, USA}
  \icmlaffiliation{ucb}{University of California, Berkeley, Berkeley, California, USA}

  \icmlcorrespondingauthor{Qian Xie}{qx66@cornell.edu.} 

  \icmlkeywords{Bayesian Optimization, Stopping Rules, Machine Learning}

  \vskip 0.3in
]

\printAffiliationsAndNotice{\icmlEqualContribution}

\begin{abstract}
  In automated machine learning, scientific discovery, and other applications of Bayesian optimization, deciding when to stop evaluating expensive black-box functions in a cost-aware manner is an important but underexplored practical consideration.
  A natural performance metric for this purpose is the \emph{cost-adjusted simple regret}, which explicitly captures the trade-off between solution quality and cumulative evaluation cost.
  Existing stopping rules for Bayesian optimization are either heuristic, or are theoretically grounded but designed to optimize simple regret without accounting for evaluation costs; as a result, they provide no guarantees against unnecessary evaluations when costs are high. We propose a \emph{principled cost-aware stopping rule} for Bayesian optimization that adapts to varying evaluation costs without heuristic tuning. Our rule is grounded in a theoretical connection to state-of-the-art cost-aware acquisition functions, namely the Pandora's Box Gittins Index (PBGI) and log expected improvement per cost (LogEIPC). When paired with either acquisition function, we prove that the resulting policy satisfies a theoretical guarantee bounding the expected cost-adjusted simple regret. Across synthetic tasks and empirical benchmarks including hyperparameter optimization and neural architecture size search, pairing our stopping rule with PBGI or LogEIPC usually matches or outperforms other acquisition-function--stopping-rule pairs in terms of cost-adjusted simple regret. Code available at: \url{https://github.com/QianJaneXie/CostAwareStoppingBayesOpt}.
\end{abstract}

\section{Introduction}
\label{sec:intro}

Bayesian optimization is a framework designed to efficiently find approximate solutions to optimization problems involving expensive-to-evaluate black-box functions, where derivatives are unavailable. Such problems arise in applications like hyperparameter tuning \citep{snoek2012practical}, robot control optimization \citep{martinez2017bayesian}, and material design \citep{zhang2020bayesian}. 
It works by iteratively, (a)~forming a probabilistic model of the black-box objective function based on data collected thus far, then (b)~optimizing an acquisition function, which balances exploration-exploitation tradeoffs, to carefully choose a new point at which to observe the unknown function in the next iteration.

In this work, we consider the \emph{cost-aware} setting, where one must pay a cost to collect each data point, and study \emph{adaptive stopping rules} that choose when to stop the optimization process.
After stopping at some terminal time, we measure performance in terms of simple regret, which is the difference in value between the best solution found so far and the global optimum.
Collecting a data point can reduce simple regret, but incurs cost in order to do so.

As an example, consider using a cloud computing environment to tune the hyperparameters of a classifier in order to optimize a performance metric on a given test set.
Training and evaluating test error takes some number of CPU or GPU hours, that may depend on the hyperparamaters used. 
These come with a financial cost, billed by the cloud computing provider, which define our cost function.
The objective value is the business value of deploying the trained model under the given hyperparameters---a given function of the model's accuracy.
From this perspective, simple regret can be understood as the opportunity cost for deploying a suboptimal model instead of the optimal one. Motivated by the need to balance cost-performance tradeoff in examples such as above, we aim to design stopping rules that optimizes \emph{expected cost-adjusted simple regret}, defined as the sum of simple regret and the cumulative cost of data collection.

Several stopping rules have been proposed for Bayesian optimization.
Simple heuristics—such as fixing the number of iterations or stopping when the best value remains unchanged for a certain number of iterations—are widely used, but can either stop too early or lead to unnecessary evaluations.
Other approaches include acquisition-function-based rules that stop when metrics such as probability of improvement \citep{lorenz2015stopping} fall below a preset threshold, or when expected improvement (EI) \citep{nguyen2017regret, zhou2024corrected} or knowledge gradient (KG) \citep{frazier2008knowledge} fall below the cost per sample. Regret-bound-based rules instead aim to optimize simple regret \citep{makarova2022automatic, ishibashi2023stopping, wilson2024stopping, li2023not}.
These approaches are primarily developed for the classical uniform-cost setting and do not explicitly account for varying evaluation costs. Our work instead provides a principled stopping rule with theoretical guarantees that applies both to the uniform-cost setting and more generally to varying evaluation costs.

In this work, we study how to design cost-aware stopping rules, motivated by two primary factors.
First, state-of-the-art cost-aware acquisition functions such as the Pandora's Box Gittins Index (PBGI) \citep{xie2024cost} and log expected improvement per cost (LogEIPC) \citep{ament2023unexpected} have not yet been studied in the adaptive stopping setting.
As our experiments in \cref{sec:experiment} show, introducing adaptive stopping can further improve the performance of these acquisition functions compared to fixed-budget evaluations.
Second, while regret-based stopping rules such as UCB–LCB \citep{makarova2022automatic} guarantee low simple regret, they often incur excessive evaluation costs, resulting in high cost-adjusted regret, as reflected in our experiments.

Our work builds upon the Bayesian-optimal Pandora’s Box decision principle underlying the PBGI acquisition function and extends it to the adaptive stopping setting in correlated Bayesian optimization.
Furthermore, we show that an existing stopping rule proposed for the EI acquisition function in the classical cost-unaware setting \citep{nguyen2017regret,zhou2024corrected}, admits a principled choice of the stopping threshold as the evaluation cost per sample, rather than a heuristic tuning parameter. Under this choice, the resulting stopping condition is equivalent to the PBGI stopping rule, and naturally extends to the cost-aware setting.
This stopping rule can therefore naturally be paired with acquisition functions derived from either the PBGI or the EI design principles. Acquisition functions such as KG or MES, which are derived from alternative principles such as value-of-information or entropy search, inherently require their own matched stopping rules.

Our contributions are summarized as follows:
\begin{enumerate}
\item \emph{A Novel Cost-Aware Stopping Rule.} We present an adaptive stopping rule derived from Pandora's box theory, which also naturally extends to the expected improvement design principle, establishing a unified and principled framework applicable to both classic and cost-aware Bayesian optimization.
\item \emph{Theoretical Guarantees.} We prove in \Cref{thm:non_neg_util} that our stopping rule, when paired with the PBGI or LogEIPC acquisition function, satisfies a theoretical upper bound on the expected cost-adjusted regret, which constitutes the first theoretical guarantee of this type for any adaptive stopping rule for Bayesian optimization.
\item \emph{Empirical Validation.} We conduct a systematic empirical evaluation across multiple acquisition-function–-stopping-rule pairs in cost-aware Bayesian optimization. Our results on synthetic tasks and empirical benchmarks show that pairing our proposed stopping rule with the PBGI or LogEIPC acquisition function usually matches or outperforms other pairs.
\end{enumerate}
\section{Bayesian Optimization and Adaptive Stopping}
\label{sec:background}

In black-box optimization, the goal is to find an approximate optimum of an unknown objective function $f : X \to \mathbb{R}$ using a limited number of function evaluations at points $x_1, \ldots, x_T \in X$ where $X$ is the search space and $T$ is a given search budget, potentially chosen adaptively. 
The convention measures performance in terms of \emph{expected simple regret} \citep[Sec. 10.1]{garnett2023bayesian}, given by
\begin{equation}
\label{eqn:bayesopt_problem}
\mathcal{R} = \mathbb{E} \left[\min_{1\leq t\leq T} f(x_t) - \inf_{x\in X} f(x) \right],
\end{equation}
where the expectation is taken over all sources of randomness, including the sequence of points $x_1, \ldots, x_T$ selected by the algorithm.
Bayesian optimization approaches this problem by building a probabilistic model of $f$---typically a Gaussian process (GP)~\citep{rasmussen2006gaussian}, conditioned on the observed data points $(x_t, y_t)_{t=1}^T$, where $y_t = f(x_t)$. 
For each iteration $t=1,\ldots,T$, an acquisition function $\alpha_{t}: X \to \mathbb{R}$ then guides the selection of new samples by carefully balancing the exploration-exploitation tradeoff arising from uncertainty about $f$.

\subsection{Cost-aware Bayesian Optimization}
\label{sec:background:acquisition}
\emph{Cost-aware Bayesian optimization} \citep{lee2020cost,astudillo2021multi} extends the above setup to account for the fact that evaluation costs can vary across the search space.
For instance, in the hyperparameter tuning example of \Cref{sec:intro}, costs can vary according to the time needed to train a machine learning model under a given set of hyperparameters $x$.
Cost-aware Bayesian optimization handles this by introducing a cost function $c: X \to \mathbb{R}^+$, which may be known or unknown ahead.
The cost function, or observed costs, are then used to construct the acquisition function $\alpha_t$.

In this work, we focus primarily on the \emph{cost-per-sample} formulation \citep{chick2012sequential,cashore2016multi,xie2024cost} of cost-aware Bayesian optimization, which seeks methods with stopping time $\tau$, not necessarily fixed, that achieve a low \emph{expected cost-adjusted simple regret}
\begin{equation}
\mathcal{R}_c = \mathbb{E} \Bigg[\underbrace{\vphantom{\sum_{t=1}^{\tau}}\min_{1\le t \le \tau} f(x_t) - \inf_{x \in X} f(x)}_{\text{simple regret}} + \underbrace{\sum_{t=1}^{\tau} c(x_t)
    }_{\text{cumulative cost}}\Bigg] \label{eqn:cost_per_sample}
    .
\end{equation}
One can also work with the \emph{expected budget-constrained} formulation \citep{xie2024cost}, which incorporates budget constraints explicitly, and seeks algorithms which achieve a low expected simple regret under an expected evaluation budget.
Here, performance is evaluated in terms of
\begin{align}
\mathcal{R} &= \mathbb{E} \left[\min_{1\leq t\leq {\tau}} f(x_t) - \inf_{x\in X} f(x)\right] \nonumber \\
&\text{where } x_1,\ldots,x_{\tau} \text{ satisfy \quad } 
\mathbb{E}\sum\limits_{t=1}^{\tau} c(x_t) \leq B. \label{eqn:cost_aware_bayesopt_problem}
\end{align}
The stopping rules we study can be applied in this setting as well: we discuss this in \Cref{sec:method:theory}.

For both settings, we work with a few acquisition functions---chiefly, the \emph{Pandora's Box Gittins Index (PBGI)} \citep{xie2024cost} and \emph{log expected improvement per cost (LogEIPC)} \citep{xie2024cost} (the log variant \citep{ament2023unexpected} of the expected improvement per cost \citep{snoek2012practical}).
At iteration $t$, let $x_{1:t}:=\{x_1, \dots, x_t\}$ be the evaluated points, $y_{1:t}:=\{y_1, \dots, y_t\}$ be their observed values, $y^*_{1:t}=\min_{1\leq s \leq t} y_s$ be the current best observed value, and $f \mid x_{1:t}, y_{1:t}$ be the posterior distribution of $f$. Then the PBGI acquisition function is defined as
\begin{align}
\label{eqn:PBGI}
&\alpha_t^{\mathrm{PBGI}}(x) := \nonumber\\
&\begin{cases}
g \quad \text{s.t. } 
\mathrm{EI}_{f \mid x_{1:t}, y_{1:t}}(x; g) =  c(x), & x\notin x_{1:t}, \\
f(x), & x\in x_{1:t},
\end{cases}
\end{align}
where the value $g$ is the threshold such that the expected improvement over $g$ equals the cost\footnote{For an evaluated point $x$, we set $g = f(x)$ since the posterior at $x$ collapses to a point mass at the observed value $f(x)$ and its evaluation cost can be treated as $0$.} and $\mathrm{EI}_\psi(x; y) = \mathbb{E} \left[\max(y - \psi(x),0)\right]$ is the expected improvement at point $x$ with respect to some random function $\psi : X \to \mathbb{R}$ and a baseline value~$y$.
The LogEIPC acquisition function is defined as
\begin{align}
\label{eqn:EIC} \alpha_{t}^{\mathrm{LogEIPC}}(x) := \log\frac{\mathrm{EI}_{f \mid x_{1:t}, y_{1:t}}(x; y^*_{1:t})}{c(x)},
\end{align}
i.e., the logarithm of the expected improvement divided by the cost.
We also consider classical cost-unaware acquisition functions---\emph{lower confidence bound (LCB)} and \emph{Thompson sampling (TS)}; see \citet{garnett2023bayesian} for details.

\subsection{Adaptive Stopping Rules} \label{sec:background:stopping_rule_review}
To the best of our knowledge, stopping rules for Bayesian optimization typically do not incorporate cost explicitly into the stopping criterion.
We broadly categorize existing methods as follows:

\textbf{Criteria based on convergence or significance of improvement.} 
This includes empirical convergence, namely stopping when the best value remains unchanged for a fixed number of iterations, or the \emph{global stopping strategy (GSS)} \citep{bakshy2018ae}, which stops when the improvement is no longer statistically significant relative to the inter-quartile range (i.e., the range between the 25th percentile and the 75th percentile) of prior observations.
In the multi-fidelity setting, \citet{FourmaniB25} proposed stopping when the high-fidelity surrogate’s estimated optimum has stabilized over a window of iterations, which is related to cost-awareness but differs from our setting with explicitly varying evaluation costs.

\textbf{Acquisition-based criteria.} This includes stopping rules based on acquisition functions such as PI, EI, and KG. These approaches typically stop when the acquisition value falls below a threshold—such as a predetermined constant \citep{lorenz2015stopping,locatelli1997bayesian}, the median of the initial acquisition values \citep{ishibashi2023stopping}, or the cost of sampling \cite{nguyen2017regret,zhou2024corrected,frazier2008knowledge}. They are typically defined via fixed thresholds and are primarily designed for the uniform-cost setting, without explicitly accounting for varying evaluation costs.

\textbf{Regret-based criteria.} This includes stopping rules based on confidence bounds such as \emph{UCB-LCB} \citep{makarova2022automatic} and the \emph{gap of expected minimum simple regrets} \citep{ishibashi2023stopping}. These stop when certain estimated regret bounds fall below a preset or data-driven threshold. The related \emph{probabilistic regret bound (PRB)} stopping rule \citep{wilson2024stopping} stops when estimated simple regret is below a small threshold $\epsilon$ with confidence $1 - \delta$.
In contrast to these forward-looking rules, \citet{li2023not} propose a backward-looking stopping rule that detects when the search has entered a locally convex region and stops once an estimate of local regret falls below a predetermined threshold.

In settings beyond Bayesian optimization, \citet{chick2012sequential} have proposed a cost-aware stopping rules for finite-domain sequential sampling problems with independent values. 
Although this formulation allows for varying costs, it does not extend to general Bayesian optimization settings which use correlated GP models.

\section{A Stopping Rule Based on the Pandora's Box Gittins Index}
\label{sec:method}

In this work, we propose a new stopping rule tailored for two state-of-the-art acquisition functions used in cost-aware Bayesian optimization: PBGI and LogEIPC, introduced in \Cref{sec:background}.
As discussed by \citet{xie2024cost}, these acquisition functions are closely connected, arising from two different approximations of the intractable dynamic program which defines the Bayesian-optimal policy for cost-aware Bayesian optimization.
We now show that this connection can be used to obtain a principled stopping criterion to be paired with both acquisition functions.

\paragraph{A stopping rule for PBGI.}

To derive a stopping rule for PBGI, consider first the Pandora’s Box problem, from which it is derived.
Pandora’s Box can be seen as a special case of cost-per-sample Bayesian optimization, as introduced in \Cref{sec:background}, where $X = \{1,\dots,N\}$ is a discrete space and $f$ is assumed uncorrelated.
In this setting, the observed values do not affect the posterior distribution---meaning, we have $f(x') \mid x_{1:t}, y_{1:t} = f(x')$ at all unobserved points $x'$---and thus the acquisition function value at point $x$ is time-invariant and can be written simply as $\alpha^{\mathrm{PBGI}}(x)$, where $\mathrm{EI}_{f}(x; \alpha^{\mathrm{PBGI}}(x)) = c(x)$.
Following \citet{weitzman1979optimal}, one can show using a Gittins index argument that selecting $x_t$ to minimize $\alpha^{\mathrm{PBGI}}$ is Bayesian-optimal under minimization—the algorithm which does so achieves the smallest expected cost-adjusted regret, among all adaptive algorithms.

One critical detail applied in the optimality argument of \citet{weitzman1979optimal} is that the policy defined by $\alpha^{\mathrm{PBGI}}$ is \emph{not} Bayesian-optimal for any fixed deterministic $T$.
Instead, it is optimal only when the stopping time $T$ is chosen according to the condition
\begin{equation}
\label{eqn:Weitzman_stopping}
\min_{x\in X\backslash\{x_1,\dots,x_t\}}\alpha^{\mathrm{PBGI}}(x) \geq y^*_{1:t},  
\end{equation}
where $\geq$ can be replaced by $>$\footnote{Following \citet[Appendix B]{xie2024cost}, optimality holds under any tie-breaking rule in the cost-per-sample setting, but only under a carefully-chosen stochastic tie-breaking rule in the expected budget-constrained setting. For simplicity, we use the stopping-as-early-as-possible tie-breaking rule in this paper.}, and as before, $y^*_{1:t}$ is the best value observed so far.

In the correlated setting we study, \citet{xie2024cost} extend $\alpha^{\mathrm{PBGI}}$ to define an acquisition function by recomputing it at each iteration $t$ based on the posterior mean and variance, which defines $\alpha^{\mathrm{PBGI}}_t$ as in \Cref{eqn:PBGI}.
In order to also extend the associated stopping rule, a subtle design choice arises: should we use $\alpha_{t-1}^{\mathrm{PBGI}}$ (before posterior update) or $\alpha_t^{\mathrm{PBGI}}$ (after posterior update) in \Cref{eqn:Weitzman_stopping}?
While prior theoretical work \citep{gergatsouli2023weitzman} adopts the former choice, we instead use the latter (after posterior update), resulting in the stopping rule
\begin{equation}
\label{eqn:PBGI_stopping}
\min_{x\in X\backslash\{x_1,\dots,x_t\}}\alpha_t^{\mathrm{PBGI}}(x) \geq y^*_{1:t}.  
\end{equation}
This choice is motivated by two reasons. First, we argue it more faithfully reflects the intuition behind Weitzman’s original stopping rule. 
In the independent setting, $\alpha^{\mathrm{PBGI}}(x)$ represents a kind of \emph{fair value} for point $x$ \citep{kleinberg2016descending}.
For an evaluated point $x \in \{x_1, \dots, x_t\}$, this is simply the observed function value.
For an unevaluated point $x \not\in \{x_1, \dots, x_t\}$, the fair value reflects uncertainty in $f(x)$ and the cost $c(x)$.
The stopping rule \Cref{eqn:PBGI_stopping} then says to \emph{stop when the best fair value is among the already-evaluated points}---namely, when no point provides positive expected gain relative to its cost if evaluated in the next round, conditioned on current observations. 
In the correlated setting, the fair value naturally extends to $\alpha_t^{\mathrm{PBGI}}$, because this incorporates all known information at a given time.
Second, we show in \Cref{apdx:sec:order_posterior_update} that using $\alpha_t^{\mathrm{PBGI}}$ yields tangible empirical gains in cost-adjusted regret.

\paragraph{Connection with LogEIPC.}

The above reasoning may at first seem to be fundamentally tied to the Pandora's Box problem and its Gittins-index-theoretic properties.
We now show that it admits a second interpretation in terms of log expected improvement per cost, which arises from a completely different one-time-step approximation to the Bayesian-optimal dynamic program for cost-aware Bayesian optimization in the general correlated setting.

\begin{figure}[htbp]
  \centering
  \begin{subfigure}[t]{0.23\textwidth}
    \includegraphics[width=\linewidth, trim=0 0 0 0, clip]{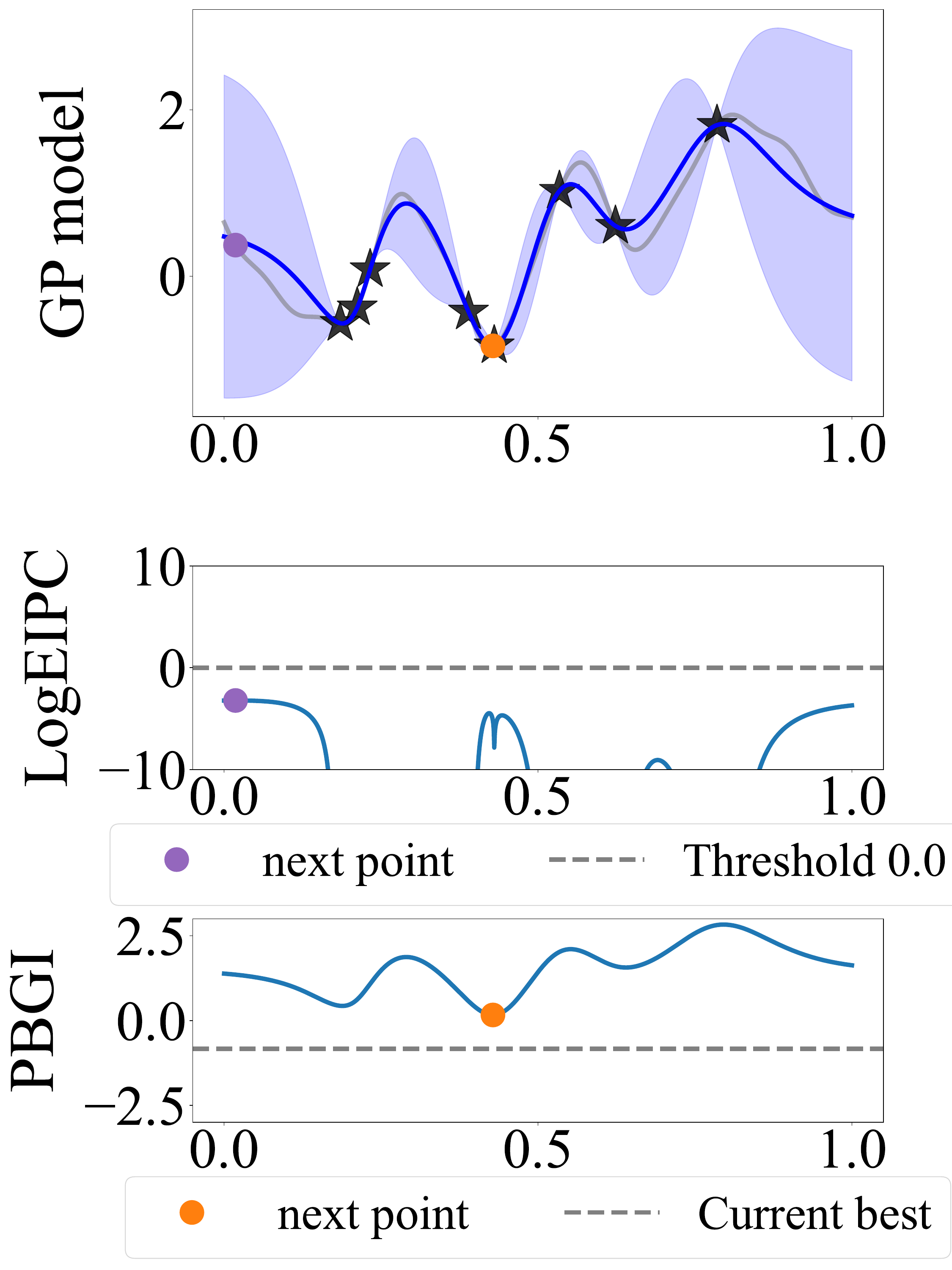}
    \caption{$c(x) \equiv 0.1$}
  \end{subfigure}
  \begin{subfigure}[t]{0.23\textwidth}
    \includegraphics[width=\linewidth, trim=0 0 0 0, clip]{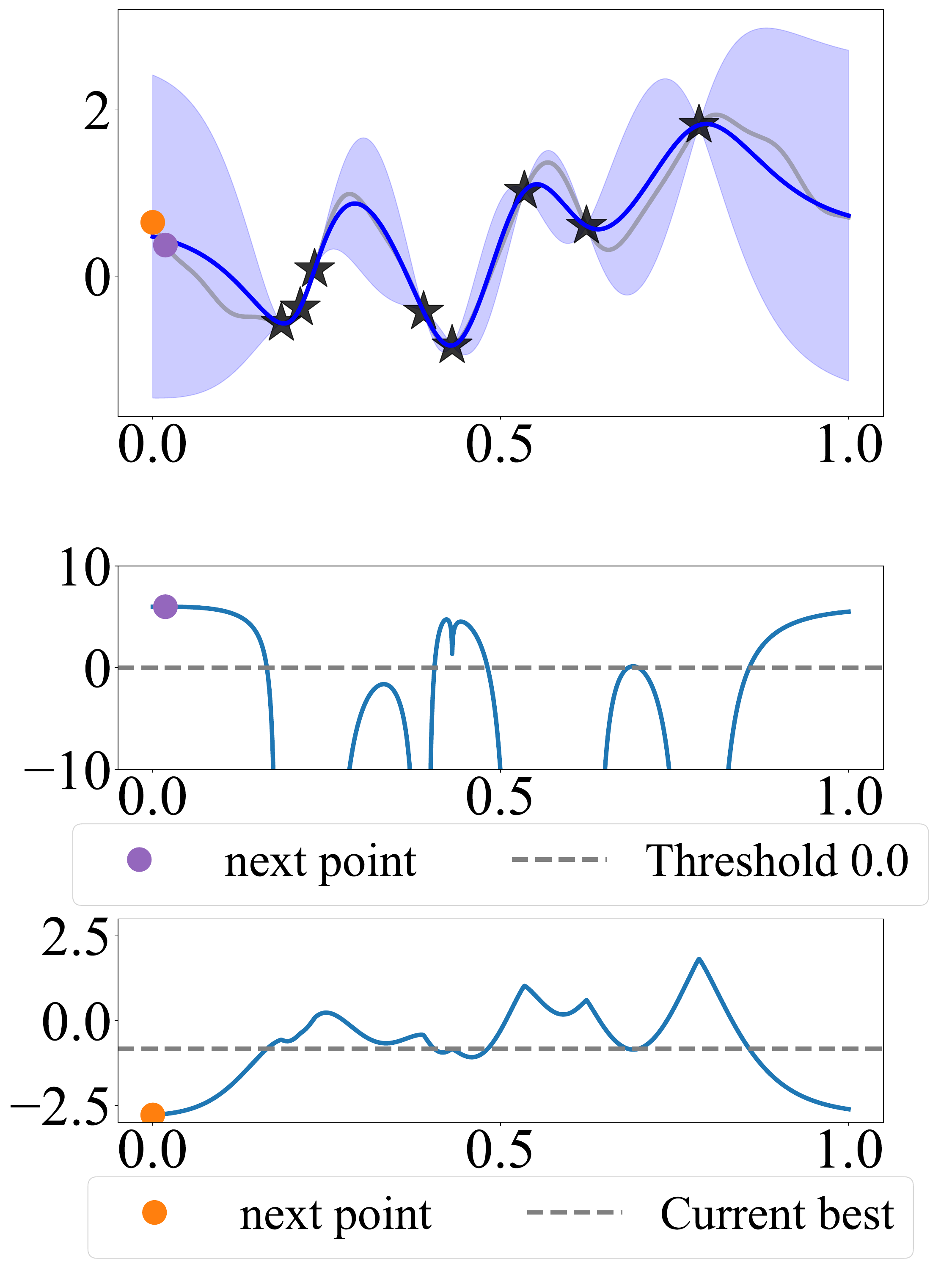}
    \caption{$c(x) \equiv 0.0001$}
  \end{subfigure}
  \caption{Illustration of the PBGI/LogEIPC stopping rule under a uniform-cost setting. When the cost-per-sample is large ($c(x) \equiv 0.1$), the minimum PBGI acquisition value exceeds the threshold (current best observed value) and the maximum LogEIPC acquisition value falls below the threshold $0.0$, indicating stopping; when the cost-per-sample is small ($c(x)\equiv 0.0001$), the minimum PBGI acquisition value is smaller than the threshold (current best observed value) and the maximum LogEIPC acquisition value remains above the threshold $0.0$, indicating no stopping.}
  \label{fig:stopping_illustration}
\end{figure}

To show this, we start with definitions above and apply a sequence of transformations.
Recall that for any unevaluated point $x \notin \{x_1, \dots, x_t\}$, $\alpha_t^{\mathrm{PBGI}}(x)$ is defined in \Cref{eqn:PBGI} to be the solution to
\begin{equation}
\mathrm{EI}_{f \mid x_{1:t}, y_{1:t}}(x; \alpha_t^{\mathrm{PBGI}}(x)) = c(x),    
\end{equation}
and since $\mathrm{EI}_{\psi}(x; y)$ is strictly increasing in $y$, any inequality involving $\alpha_t^{\mathrm{PBGI}}(x; c)$ can be lifted through the EI function without changing its direction, which means 
$\alpha_t^{\mathrm{PBGI}}(x) \geq y^*_{1:t}$ holds if and only if
$\mathrm{EI}_{f \mid x_{1:t}, y_{1:t}}\left(x; y^*_{1:t}\right) \leq c(x)$. 
This implies that the PBGI stopping rule from \Cref{eqn:PBGI_stopping} is equivalent to stopping when
\begin{equation}
    \label{eqn:EI_c_stopping}
\mathrm{EI}_{f \mid x_{1:t}, y_{1:t}}\left(x; y^*_{1:t}\right) \leq c(x),
\quad
\forall x \in X \backslash \{x_1, .., x_t\},
\end{equation}
meaning \emph{stop when no point's expected improvement is worth its evaluation cost}.
Rearranging the inequality and taking logs, we can rewrite this condition using the LogEIPC acquisition function\footnote{Excluding evaluated points is not theoretically required but often beneficial in practice, as numerical stability adjustments can cause their expected improvement to remain positive.}
:
\begin{align}
    \label{eqn:LogEIPC_stopping}
    \max_{x\in X\backslash \{x_1, \dots, x_t\}} \alpha_t^{\mathrm{LogEIPC}}(x; y^*_{1:t}) \leq 0
    .
\end{align}
When \(c(x)\equiv c_0\), Equation (8) reduces to the EI thresholding stopping rule \cite{nguyen2017regret,zhou2024corrected} 
\begin{align}
    \label{eqn:EI_stopping}
    \max_{x\in X\backslash \{x_1, \dots, x_t\}} \alpha_t^{\mathrm{EI}}(x; y^*_{1:t}) \leq c_0
    .
\end{align}
We call the stopping rule given by the equivalent conditions (\Cref{eqn:PBGI_stopping,eqn:EI_c_stopping,eqn:LogEIPC_stopping}) the \emph{PBGI/LogEIPC stopping rule}.
\Cref{fig:stopping_illustration} gives an illustration of how the rule behaves in a simple setting, demonstrating that it is more conservative---preferring to stop earlier---when the cost is high.

\subsection{Theoretical Guarantees on Cost-Adjusted Regret} \label{sec:method:theory}
The connection between PBGI and LogEIPC in fact goes beyond a shared stopping rule. In \Cref{lem:EI_lower_bound}, we prove that when paired with this stopping rule, both acquisition functions guarantee that at each iteration before stopping, the expected improvement at the selected point is at least its evaluation cost. 


\begin{restatable}[Expected improvement exceeds cost before stopping]{lemma}{LemEIlb} \label{lem:EI_lower_bound}
Let $X$ be compact, and let $f : X \to \mathbb{R}$ be a random function with prior mean $\mu(\cdot)$. 
Consider a Bayesian optimization algorithm that begins at some initial point $x_1\in X$ with cost $C = c(x_1)$, acquires subsequent points using either the PBGI or LogEIPC acquisition function, and terminates according to the PBGI/LogEIPC stopping rule.
Let $\tau = \min_{t \geq 1} \{\sup_{x \in X} \alpha_{t}^{\mathrm{LogEIPC}}(x) \leq 0\}$ be the algorithm's stopping time, and denote the posterior expected improvement function by $\alpha_{t}^{\mathrm{EI}}(x) = \mathrm{EI}_{f \mid x_{1:t}, y_{1:t}}\left(x; y^*_{1:t}\right)$. Then, for all $t < \tau$, $\alpha_{t}^{\mathrm{EI}}(x_{t+1})\ge c(x_{t+1})$.
\end{restatable}
 
As a consequence of this claim, we show in \Cref{thm:non_neg_util} that the PBGI/LogEIPC stopping rule, when paired with PBGI or LogEIPC, achieves cost-adjusted simple regret no worse than stopping immediately after the initial evaluation.
Notably, many acquisition function–stopping rule pairs fail to satisfy this \emph{no-worse-than-immediate} property in practice (see \Cref{fig:synthetic,fig:empirical}), largely because most stopping rules are designed with simple regret alone in mind and do not explicitly account for evaluation costs.
Moreover, this guarantee is tight in the worst case: when evaluation costs are sufficiently large relative to the maximum achievable improvement, immediate stopping is optimal (see \Cref{prop:immediate_optimal}).
To our knowledge, this is the first formal guarantee on cost-adjusted simple regret for Bayesian optimization, generalizing beyond simpler settings such as Pandora’s~Box, which assumes independent and discrete evaluations.

\begin{restatable}[No worse than immediate stopping]{theorem}{ThmPositiveUtil} 
\label{thm:non_neg_util}
Consider the setting and algorithm specified in \Cref{lem:EI_lower_bound}. Let $U:= \mu(x_1) - \mathbb{E}[\min_{x\in X}f(x)]<\infty$, 
then the algorithm's expected cost-adjusted simple regret is bounded by
\begin{align}&\mathbb{E}\left[y_{1:\tau}^* - \min_{x\in X}f(x) + \sum_{t=1}^{\tau} c(x_t)\right]
    \nonumber\\
    &\leq \mathbb{E}\big[
        \underbrace{
        y_{1} - \min_{x\in X}f(x) + c(x_1)}_{\textnormal{cost-adjusted regret of immediate stopping}}
    \big]
    = U+C.\label{eqn:cost_adjusted_regret_bound}
\end{align}
\end{restatable}
\vspace{-6pt}
This result yields a bound on expected cumulative cost and, under the natural assumption that practical evaluations have a positive minimum cost, a high-probability finite-time termination guarantee.
\begin{restatable}[Finite-time termination under positive costs]{corollary}{ThmTermination}
    Consider the setting and algorithm specified in \Cref{lem:EI_lower_bound}. Then the expected cumulative cost of the algorithm is bounded by   $\mathbb{E}\left[\sum_{t=1}^{\tau} c(x_t)\right] \leq U+ C$. Further, if evaluation costs are uniformly bounded below by a constant $c_0>0$, i.e., $c(x)\geq c_0, \forall x \in X$, then for any $\delta \in (0, 1)$, the algorithm terminates in at most $\frac{U +C}{\delta \cdot c_0}$ iterations with probability $1 - \delta$.     
\end{restatable}

\begin{remark}
\Cref{lem:EI_lower_bound} and \Cref{thm:non_neg_util} are stated for simplicity under initialization from a single point $x_1$. The results extend directly to arbitrary initial datasets $(x_{1:t_0},y_{1:t_0})$ by conditioning on the observations and treating the resulting posterior as the effective prior. The same guarantees then hold for subsequent iterations $t\geq t_0$, with all quantities defined with respect to this updated posterior.
\end{remark}

\subsubsection{Implication in the Budgeted Setting}
\label{subsubsec:budget-constrained}
We first note that in the discrete Pandora's Box setting, under an expected budget constraint $B$, minimizing $\alpha_t^{\mathrm{PBGI}}(x)$ is Bayesian-optimal: it achieves the lowest \emph{expected simple regret, among all algorithms} satisfying the expected cumulative cost within the budget (i.e., 
$\mathbb{E}\left[\sum_{t=1}^{\tau} c(x_t)\right] \leq B$), and the cost function used in defining $\alpha_t^{\mathrm{PBGI}}(x)$ is $\lambda c(x)$ for some \emph{cost-scaling factor} $\lambda$ which depends on $B$ \citep[Theorem 2]{xie2024cost}.

This result, at first, appear to be completely Pandora's-Box-theoretic: it requires $X$ to be discrete and $f$ to be independent. 
In the more general correlated setting of cost-aware Bayesian optimization, however, Bayesian optimality of PBGI may no longer hold, and the relationship between $B$ and the choice of $\lambda$ is not immediately clear. \Cref{thm:non_neg_util} helps bridge this gap: it provides an upper bound on the expected cumulative cost up to the stopping time, which in turn yields the following principled choice of $\lambda$ that ensures compliance with the budget constraint.

\begin{restatable}[Choosing $\lambda$ for expected budget compliance]{corollary}{ThmCosts}
\label{thm:lambda_for_budget}
Consider the setting, algorithm, and notation specified in \Cref{lem:EI_lower_bound,thm:non_neg_util}, but with costs rescaled by a factor $\lambda > 0$: both the acquisition values and the stopping conditions are computed using $\lambda c(\cdot)$. 
For some budget $B>C$, if the cost-scaling factor is set to $\lambda = \frac{U}{B - C}$, 
then the algorithm’s expected cumulative unscaled cost satisfies $\E\left[\sum_{t=1}^{\tau} c(x_t)\right] \leq B$.
\end{restatable}
If this choice of $\lambda$ proves overly conservative—leading to underspending within the budget—it can be paired with the PBGI-D variant of \citet{xie2024cost}, which starts with $\lambda=\lambda_0$ and halves it each time stopping is triggered. Choosing $\lambda_0 = U/(B-C)$ aligns the initial fixed-$\lambda$ phase with the budget in expectation, while ensuring that the adaptive decay is activated as designed. In \Cref{fig:budgeted} in \Cref{apdx:experiment_results}, we show PBGI-D with this choice of $\lambda_0$ is competitive.

All proofs in this section are deferred to \Cref{apdx:theory}.

\subsection{Practical Implementation Considerations for the PBGI/LogEIPC Stopping Rule}

\label{subsec:practical-consideration}
\paragraph{Expressing objective and cost in common units.} Evaluation costs are often measured in different units than the objective, such as time versus accuracy.
To compare them directly, we rescale costs by a constant $\lambda > 0$, the conversion rate between objective improvement and cost.
For instance, if one is willing to spend $1000$ seconds to improve test accuracy by $0.01$, then $\lambda = 10^{-5}$.
Importantly, in the cost-per-sample setting we mainly study, $\lambda$ is a fixed constant set by the problem provider, rather than a parameter tuned heuristically by the user. The budgeted setting without a natural unit conversion is discussed in \Cref{subsubsec:budget-constrained}.


\paragraph{Unknown costs.}
In practice, evaluation costs are often not known in advance. They can be modeled either (i) deterministically, using domain knowledge (e.g., proportional to model size in hyperparameter tuning), or (ii) stochastically, via assumptions on the distribution of $c(x)$ (our theoretical guarantees still hold after replacing $c(x)$ with $\E[c(x)]$). In \Cref{sec:experiment}, we present empirical results under both modeling approaches. For the latter, we follow \citet[Proposition~2]{astudillo2021multi} to model $\ln c(x)$ as a GP with posterior mean $\mu_{\ln c}$ and variance $\sigma^2_{\ln c}$. To compute PBGI or LogEIPC, as well as their related stopping rules (including those in the prior work and ours), we replace $c(x)$ in \Cref{eqn:EIC,eqn:PBGI,eqn:EI_c_stopping} with $\mathbb{E}[c(x)]= \exp(\mu_{\ln c}(x)+(\sigma_{\ln c}(x))^2 / 2 )$. The empirical performance is preserved. See \Cref{apdx:sec:empirical_results} for further discussion.

\paragraph{Preventing spurious stops.}
Although our stopping rule has theoretical guarantees, in practice, it---like other adaptive stopping rules---can still suffer from spurious stops caused by two main sources.
First, early in the optimization process, the GP model parameters often fluctuate significantly as new data points are collected, causing unstable stopping signals.
To mitigate this, we enforce a stabilization period consisting of the first few evaluations, during which we do not allow any stopping rule to trigger.
Second, imperfect optimization of the acquisition function---which is especially common in continuous higher-dimensional search spaces---can lead to misleading stopping signals.
To handle this, we smooth the stopping signal in this setting by applying a moving average over a window of $W$ iterations, and require the stopping condition to hold consistently under this smoothed signal before stopping.
See \Cref{fig:moving_average} for an illustration.
We use the same window size $W=20$ for both the stabilization period and smoothing to minimize the number of parameters in our main experiments. A ablation study on the effect of window size can be found in Appendix~\ref{sec:ma_ablation}. 

\section{Experiments}
\label{sec:experiment}
To evaluate our proposed PBGI/LogEIPC stopping rule, we design three complementary sets of experiments that progressively evaluate its performance. 
First, we consider an idealized, low-dimensional Bayesian regret setting in which the GP model exactly matches the true objective. This controlled setting allows us to isolate the effect of different stopping rules without interference from model misspecification and acquisition function optimization errors. 
Then, we move to a higher-dimensional Bayesian regret setting where acquisition-function optimization is imperfect. 
Finally, we evaluate in practical settings with objective model misspecification, using the LCBench hyperparameter tuning benchmark and the NATS neural architecture size search benchmark. 
In each case, we compare pairs of acquisition functions and stopping rules, which we describe next.

\paragraph{Acquisition functions.} We consider four common acquisition functions that were discussed in \Cref{sec:background}: PBGI, LogEIPC, LCB, and TS---chosen for their competitive performance, computational efficiency, and close connections to existing stopping rules. 

\paragraph{Baselines.} We compare the proposed PBGI/LogEIPC stopping rule against several stopping rules from prior work: \emph{UCB–LCB}  \citep{makarova2022automatic}, \emph{LogEIPC-med} \citep{ishibashi2023stopping}, \emph{SRGap-med} \citep{ishibashi2023stopping}, and \emph{PRB} \citep{wilson2024stopping}. We also include two simple heuristics used in practice: \emph{Convergence} and \emph{GSS}. 
\emph{UCB–LCB} stops once the gap between upper and lower confidence bounds falls below a configurable threshold $\theta$. \emph{LogEIPC-med} stops when the log expected improvement per cost drops beneath $\log(\eta)$ plus the median of its initial $I$ values. 
\emph{SRGap-med} stops when the gap of the expected minimum simple regret falls below $\chi$ times the median of its initial $I$ values.
\emph{PRB} triggers once a probabilistic regret bound satisfies the regret tolerance $\epsilon$ and confidence $\delta$ parameters. \emph{Convergence} stops as soon as the best observed value remains unchanged for $k$ iterations.
\emph{GSS} stops if the recent improvement is less than $\phi \times \mathrm{IQR}$ over the past $k$ trials where $\mathrm{IQR}$ denotes the inter-quartile range of current observations. 
Finally, we include two reference baselines: \emph{Immediate}, which stops right after the initial evaluation (see \Cref{sec:method:theory}), and \emph{Hindsight}, which, for each trial, selects the stopping time that yields the lowest cost-adjusted simple regret in hindsight, thus providing a lower bound on achievable performance. 

In our experiments, unless specified otherwise, we follow parameter values recommended in the literature, and set $\theta = 0.01$, $\eta=0.01$, $\chi = 0.01$ (except $\chi=0.1$ for LCBench), $I = 20$, $\epsilon = 0.1$ for Bayesian regret experiments, $\epsilon = 0.5\%$ of the best test error for empirical experiments, $\delta=0.05$, $k=5$, and $\phi = 0.01$.
For the stabilization and smoothing window, we use $W=20$, aligned with the initial window size $I=20$ used in EI-med \citep{ishibashi2023stopping}.

Each experiment is repeated with 50 random seeds to assess variability, and we report the mean with error bars (2 times the standard error) for each stopping rule. 
Each trial, in the sense of a run with a distinct random seed, is capped at a fixed number of iterations; if a stopping rule is not triggered within this limit, the stopping time is set to the cap. Details are provided in \Cref{apdx:experiment_setup}.
\subsection{Bayesian Regret}
We first evaluate our PBGI/LogEIPC stopping rule on random functions sampled from prior. 
Specifically, objective functions are sampled from Gaussian processes with Matérn 5/2 kernels with length scale $0.1$, defined over spaces of dimension $d=1$ and $d=8$.
We consider a variety of evaluation cost function types, including uniform costs, linearly increasing costs in terms of parameter magnitude, and periodic costs that vary non-monotonically over the domain. 

In the 1-dimensional experiments, we perform an exhaustive grid search over $[0,1]$ to isolate the effect of stopping behavior from numerical optimization.
In the 8-dimensional setting, due to instability from higher dimensionality, we apply moving average with a window size of 20 when computing the PBGI/LogEIPC stopping condition. See \Cref{fig:moving_average} in \Cref{apdx:experiment_setup} for an illustration.
More details on our experiment setup, and computational considerations are provided in \Cref{apdx:experiment_setup,apdx:experiment_results}.

\begin{figure}[!t]
    \centering
    \includegraphics[width=0.48\textwidth, trim=0 0.2cm 0 0.2cm, clip]{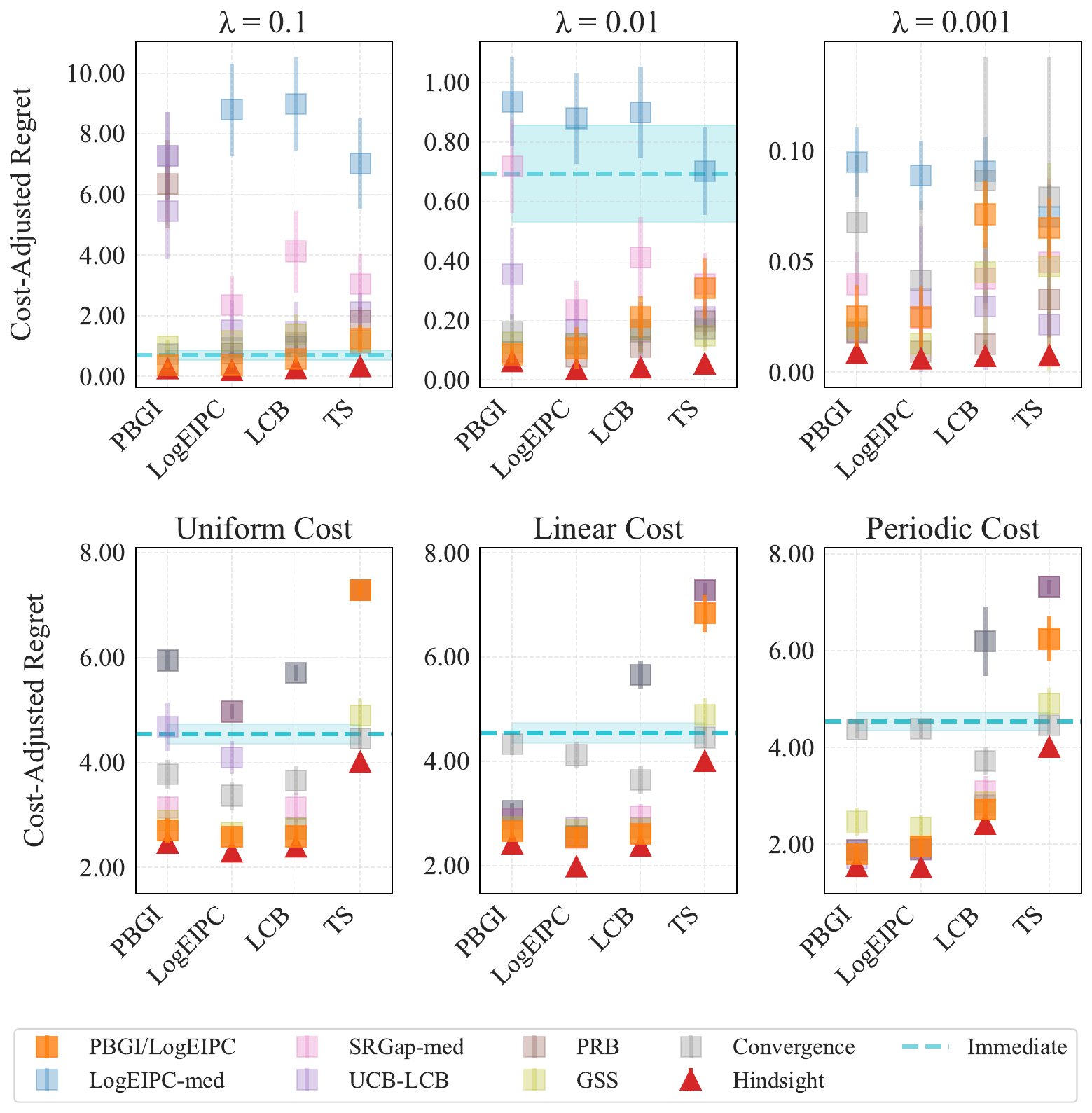}
    \caption{Cost-adjusted simple regret across acquisition-stopping rule pairs in 1D and 8D Bayesian regret setting. In 1D, objective functions are sampled from a GP with a Matérn-5/2 kernel and a linear cost function scaled by $\lambda = 0.1, 0.01, 0.001$. The Immediate baseline is omitted at $\lambda = 0.001$ due to its much higher regret (mean 0.6942, error bar [0.5314, 0.8570]). In 8D, objective functions are also drawn from a GP with a Matérn-5/2 kernel, using three cost functions scaled by $\lambda = 0.01$.}
    \label{fig:synthetic}
\end{figure}

\Cref{fig:synthetic} shows a comparison of cost-adjusted regret for pairings of acquisition functions and stopping rules under different cost-scaling factors $\lambda = 0.1, 0.01, 0.001$ in the 1-dimensional setting, and under different cost regimes (uniform, linear, periodic) in 8-dimensional setting. 

In the 1-dimensional setting, the optimization problem is relatively straightforward and all acquisition functions have nearly identical hindsight optima, independent of cost scale or cost type. From the plots, our PBGI/LogEIPC stopping rule, when paired separately with the PBGI and LogEIPC acquisition functions, delivers competitive performance for each $\lambda$, and is particularly strong in handling high-cost scenarios---those with large $\lambda$.

In the 8-dimensional experiments, however, clear gaps emerge. Under uniform and linear costs, PBGI, LogEIPC, and LCB exhibit similar hindsight optima, substantially better than TS, while in the periodic case, PBGI and LogEIPC achieve much better hindsight optima than LCB and TS. In every cost regime, combining our stopping rule with the PBGI  or LogEIPC acquisition function yields cost-adjusted regret that nearly matches the hindsight optimal. 

This demonstrates that our PBGI/LogEIPC stopping rule is relatively robust to the increased optimization difficulty. 
Further discussions of our experiments across multiple cost types and cost-scaling factors can be found in \Cref{apdx:experiment_results}.

\subsection{Automated Machine Learning Benchmarks}

We then evaluate on two empirical automated machine learning benchmarks based on use cases from hyperparameter optimization and neural architecture search: LCBench \citep{ZimLin2021a} and NATS-Bench \citep{dong2021nats}.
Unless otherwise specified, results in this section assume a known cost function; results under the unknown-cost setting are provided in \Cref{apdx:experiment_results}.

LCBench~\cite{ZimLin2021a} provides training data of 2,000 configurations over a 7-dimensional hyperparameter space, evaluated on 35 OpenML \cite{vanschoren2014openml} datasets. 
In the \emph{unknown-cost} experiments, the cost is the full 51-epoch training time. 
For the known-cost setting, we observe that training time scales approximately linearly with the number of model parameters. 
Based on a linear regression fit (see \Cref{apdx:experiment_setup}), we adopt $\alpha p$ as a proxy for training time, where $p$ is the number of model parameters and $\alpha$ is a dataset-specific estimated coefficient.

NATS-Bench~\cite{dong2021nats} provides a search space of 32,768 neural architectures varying in channel sizes across layers, evaluated on three datasets. As in LCBench, we similarly approximate the full 90-epoch (training and validation) runtime using a linear function $\alpha F + \beta$ of the number of floating point operations $F$; see \Cref{apdx:experiment_setup} for details.

We report the mean and error bars (2 times standard error) of the \emph{cost-adjusted simple regret}, where we consider the evaluation costs to be some representative cost-scaling factor $\lambda$ (see \cref{subsec:practical-consideration}) multiplied with cumulative runtime.
Simple regret is computed as the difference between (a)~test error of the configuration with best validation error at the stopping time and (b)~the best test error over all configurations.
We present the results using proxy runtime here, and defer to \cref{apdx:experiment_results} the additional results under (i) the \emph{cost model misspecification} scenario (proxy runtime used during Bayesian optimization but actual runtime used for final performance evaluation), and (ii) the \emph{unknown-cost} scenario (actual runtime used during Bayesian optimization).

\begin{figure}[!t]
    \centering
    \includegraphics[width=0.48\textwidth]{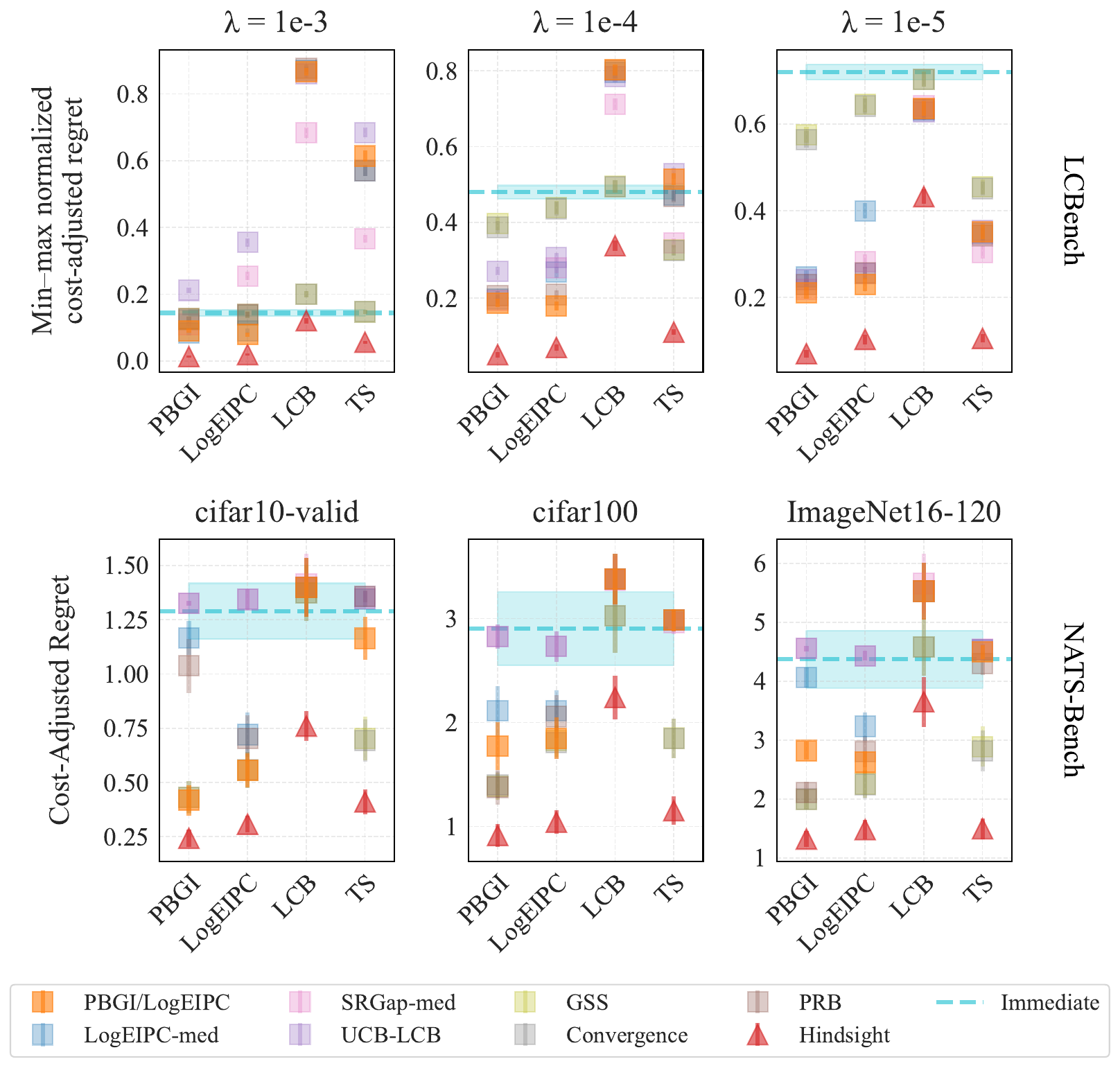}
    \caption{Cost-adjusted simple regret across acquisition function--stopping rule pairs on LCBench and NATS-Bench. The objective is to minimize validation error on classification tasks, with scaled proxy runtime as evaluation cost, scaled by representative values of $\lambda$ ($10^{-3}, 10^{-4}, 10^{-5}$ for LCBench and $10^{-5}$ for NATS-Bench).
    For LCBench, results are normalized within each dataset and random seed, then aggregated across the 35 datasets. Both matched cost-aware combinations are competitive in aggregate, with their relative performance varying with $\lambda$. Per-dataset results and within-acquisition stopping-rule ranks are reported in \Cref{fig:lcbench_all_1e-3,fig:lcbench_all_1e-4,fig:lcbench_all_1e-5,fig:ranking_full_dataset}.}
    \label{fig:empirical}
\end{figure}

\Cref{fig:empirical} presents performance of acquisition function--stopping rule pairs on LCBench and NATS-Bench. For LCBench, we report min–max normalized cost-adjusted simple regret (see definition in \Cref{apdx:experiment_results}) aggregated across 35 datasets in \Cref{fig:empirical} evaluated under three representative\footnote{These $\lambda$ values are chosen to avoid degenerate cases—neither so large that the policy stops after only a few evaluations (e.g., $\lambda = 10^{-4}$ on NATS-Bench, see \Cref{fig:cifar10-valid}) nor so small that evaluations are effectively free.} cost-scaling values. For NATS-Bench, we report cost-adjusted simple regret under $\lambda = 10^{-5}$, with additional results under two other representative cost-scaling factors shown in \Cref{fig:cifar10-valid,fig:cifar100,fig:ImageNet16-120} in \Cref{apdx:experiment_results}.

Per-dataset LCBench results are provided in \Cref{fig:lcbench_all_1e-3,fig:lcbench_all_1e-4,fig:lcbench_all_1e-5} in the appendix. Across the three cost-scaling values, our PBGI/LogEIPC stopping rule ranks among the top three of the seven non-trivial stopping rules on 21--26 of the 35 tasks when paired with PBGI, and on 24--28 tasks when paired with LogEIPC. These ranks compare stopping rules within the same acquisition function. \Cref{fig:ranking_full_dataset} reports the corresponding Top-$k$ frequencies for all 35 tasks and for the 20 datasets with $>10{,}000$ instances. Performance varies across datasets; \Cref{sec:val_test_diagnostic} examines the association between underperformance and benchmark-intrinsic validation--test mismatch.

Additional results under the cost model misspecification setting and the unknown-cost setting are provided in \Cref{fig:cost_misspecification_lcbench,fig:unknown_cost} in \Cref{apdx:experiment_results}, along with comparisons between our adaptive stopping rule and fixed-iteration baselines.
Overall, pairing our PBGI/LogEIPC stopping rule with PBGI or LogEIPC yields competitive cost-adjusted simple regret across a broad range of these empirical tasks.
We also report, in \Cref{tab:nonstops} of \Cref{apdx:experiment_results}, how often each acquisition--stopping rule pair does not stop before the 200-iteration cap. At the cost scales reported in the table, our rule almost always stops before the cap when paired with PBGI or LogEIPC: 584 of 600 trials (97.3\%) across the six datasets. Our rule frequently fails to stop before the iteration cap when paired with LCB or TS. SRGap-med and UCB--LCB also often reach the cap. These results highlight the importance of the acquisition--stopping pairing.

On LCBench, both matched cost-aware combinations remain competitive under objective misspecification, and their relative performance depends on the cost-scaling factor. In the aggregate normalized results, the LogEIPC combination has lower mean cost-adjusted regret at $\lambda=10^{-3}$ and $10^{-4}$, while the PBGI combination has lower mean regret at $\lambda=10^{-5}$. On NATS-Bench at $\lambda=10^{-5}$, the matched pairs are competitive on cifar10-valid, whereas GSS and Convergence yield lower mean cost-adjusted regret on cifar100 and ImageNet16-120 for both acquisition functions. Separately, \Cref{fig:budgeted} in \Cref{apdx:experiment_results} compares acquisition functions under fixed-budget evaluations without stopping and shows that PBGI attains lower simple regret than LogEIPC in those experiments. These results support the use of both cost-aware acquisition functions with our stopping rule, without implying a uniform ordering across cost scales and datasets.
\begin{table}[t]
\centering
\small
\setlength{\tabcolsep}{2.5pt}
\renewcommand{\arraystretch}{1.1}

\begin{tabular}{c l cc cc cc}
\toprule
\multirow{2}{*}{$\lambda$} &
\multirow{2}{*}{Acq} &
\multicolumn{2}{c}{Top 1} & \multicolumn{2}{c}{Top 2} & \multicolumn{2}{c}{Top 3} \\
\cmidrule(lr){3-4}\cmidrule(lr){5-6}\cmidrule(lr){7-8}
& & Large & All & Large & All & Large & All \\
\midrule

\multirow{2}{*}{$10^{-3}$}
& PBGI
& $30.0\%$ & $25.7\%$ & $75.0\%$ & $51.4\%$ & $80.0\%$ & $60.0\%$ \\
& LogEIPC
& $40.0\%$ & $37.1\%$ & $80.0\%$ & $65.7\%$ & $80.0\%$ & $68.6\%$ \\
\midrule

\multirow{2}{*}{$10^{-4}$}
& PBGI
& $40.0\%$ & $28.6\%$ & $70.0\%$ & $51.4\%$ & $75.0\%$ & $60.0\%$ \\
& LogEIPC
& $60.0\%$ & $40.0\%$ & $75.0\%$ & $60.0\%$ & $85.0\%$ & $74.3\%$ \\
\midrule

\multirow{2}{*}{$10^{-5}$}
& PBGI
& $35.0\%$ & $20.0\%$ & $70.0\%$ & $48.6\%$ & $90.0\%$ & $74.3\%$ \\
& LogEIPC
& $50.0\%$ & $37.1\%$ & $70.0\%$ & $62.9\%$ & $85.0\%$ & $80.0\%$ \\
\bottomrule
\end{tabular}

\caption{Percentage of LCBench tasks on which our PBGI/LogEIPC stopping rule ranks in the Top-$k$ ($k=1,2,3$) by mean cost-adjusted simple regret among the seven non-trivial stopping rules paired with the same acquisition function. Immediate and Hindsight are excluded. Results are shown for PBGI and LogEIPC on the 20 large datasets with $>10{,}000$ instances and on all 35 datasets, for $\lambda \in \{10^{-3},10^{-4},10^{-5}\}$.}
\label{fig:ranking_full_dataset}
\end{table}

\section{Conclusion}
We develop the \emph{PBGI/LogEIPC stopping rule} for Bayesian optimization with varying evaluation costs.
Paired with either the PBGI or LogEIPC acquisition function, it (a)~satisfies a theoretical guarantee bounding the expected cost-adjusted simple regret (\Cref{sec:method:theory}), and (b)~shows strong empirical performance in terms of cost-adjusted simple regret (\Cref{sec:experiment}).
We believe our framework can be extended to settings involving noisy, multi-fidelity or batched evaluations, as well as alternative objective formulations ---for instance, applying a sigmoid transformation to test error rather than a linear one, to reflect real-world user preferences that shift sharply once error falls below a threshold.

\section*{Impact Statement}
This paper presents work whose goal is to advance the field of machine learning. There are many potential societal consequences of our work, none of which we feel must be specifically highlighted here.

\section*{Acknowledgments}
QX thanks Nairen Cao for helpful discussions on hindsight optimal, Theodore Brown for helpful discussions on LCBench runtime proxy, and Raul Astudillo for helpful discussions on unknown cost handling for LogEIPC. AT thanks James T. Wilson for helpful comments on stopping rules for Bayesian optimization.
LC is funded by the European Union (ERC-2022-SYG-OCEAN-101071601).
Views and opinions expressed are however those of the author(s) only and do not necessarily reflect those of the European Union or the European Research Council Executive Agency. Neither the European Union nor the granting authority can be held responsible for them.
AT was supported by Cornell University, jointly via the Center for Data Science for Enterprise and Society, the College of Engineering, and the Ann S. Bowers College of Computing and Information Science.
PF was supported by AFOSR FA9550-20-1-0351.
ZS was supported by the NSF under grant no.\ CMMI-2307008.

\bibliography{CostAwareStopping}
\bibliographystyle{icml2026}

\newpage
\appendix
\onecolumn 

\section{Theoretical Analysis and Calculations} \label{apdx:theory}
In the following lemma, we prove a point-wise lower bound on the expected improvement before stopping for our recommended pairing of PBGI/LogEIPC stopping rule paired with either PBGI or LogEIPC acquisition function.\footnote{In fact, any acquisition function (e.g., expected improvement-cost (EIC) \citep{hu2025adjusted}) that ensures the one-step expected improvement is worth the evaluation cost before the stopping time $\tau$ can apply here.}

\LemEIlb*

\begin{proof}
While stopping has not occurred, meaning $t<\tau$, by the stopping criteria definition we have $\max_{x\in X}\alpha_{t}^{\mathrm{EI}}(x)/c(x)\ge1$.
Hence, there exists at least one point with
$\alpha_{t}^{\mathrm{EI}}(x)/c(x)\ge1$.
We now argue for each algorithm.

\smallskip
\textit{PBGI.}
For each $x$ we defined a threshold $\alpha_{t}^{\mathrm{PBGI}}(x)$ by
$\mathrm{EI}_{f \mid x_{1:t}, y_{1:t}}\left(x; \alpha_{t}^{\mathrm{PBGI}}(x)\right)=c(x)$.
Since $\mathrm{EI}_{f \mid x_{1:t}, y_{1:t}}$ is increasing in its second argument,
\begin{equation}
\;y_{1:t}^*\;\ge\;\alpha_{t}^{\mathrm{PBGI}}(x)\;\Longleftrightarrow\;
\alpha_{t}^{\mathrm{EI}}(x)/c(x)\ge1.
\end{equation}
The existence of a point with ratio at least~$1$ therefore implies that the set
$
\mathcal S_{t}=\{x: y_{1:t}^*\ge \alpha_{t}^{\mathrm{PBGI}}(x)\}
$
is non-empty. PBGI chooses $x_{t+1}$ with the \emph{smallest} threshold,
and thus
\begin{equation}
\alpha_{t}^{\mathrm{EI}}(x_{t+1})=\mathrm{EI}_{f \mid x_{1:t}, y_{1:t}}\left(x_{t+1}; y_{1:t}^*\right)\ge\mathrm{EI}_{f \mid x_{1:t}, y_{1:t}}\left(x_{t+1}; \alpha_{t}^{\mathrm{PBGI}}(x_{t+1})\right)=c(x_{t+1}).
\end{equation}

\smallskip
\textit{(Log)EIPC.}
By definition $x_{t+1}$ maximizes $\log(\alpha_{t}^{\mathrm{EI}}(x)/c(x))$ and  $\alpha_{t}^{\mathrm{EI}}(x)/c(x)$, hence 
$\alpha_{t}^{\mathrm{EI}}(x_{t+1})\ge c(x_{t+1})$.

\smallskip
Thus for \emph{both} algorithms, we have
\begin{equation}
\label{eq:key-LB}
\alpha_{t}^{\mathrm{EI}}(x_{t+1})\ge c(x_{t+1}),
\qquad\text{for all }t < \tau.
\end{equation}
\end{proof}
We can now use \cref{lem:EI_lower_bound} to prove the following theorem, where we show that our PBGI/LogEIPC stopping rule (paired with the PBGI or LogEIPC acquisition function) also achieves cost-adjusted simple regret no worse than a naive baseline---stopping-immediately (\emph{Immediate}).

\ThmPositiveUtil*

\begin{proof}
We treat the two algorithms---meaning, the two acquisition function and stopping rule pairs---together and write
$
\mathcal F_{t}=\sigma\!\left(\{x_i,y_i\}_{i=1}^{t}\right)
$
for the filtration generated by the observations.
Since we are minimizing, the one–step improvement after
iteration~$t$ is
$y_{1:t-1}^*-y_{1:t}^*$, where recall that $y_{1:t}^* = \min_{1 \leq i \leq t} y_i$.
By our assumption about the random function $f$, 
$\E[y_{1:1}^*] = \mu(x_1)$ and
the quantity $y_{1:1}^*-\min\nolimits_{x\in X}f(x)$ has finite expectation $U<\infty$. Denote the posterior expected improvement function as $\alpha_{t}^{\mathrm{EI}}(x) = \mathrm{EI}_{f \mid x_{1:t}, y_{1:t}}\left(x; y^*_{1:t}\right)$.

By \Cref{lem:EI_lower_bound}, $c(x_t)\le \alpha_{t-1}^{\mathrm{EI}}(x_t)$ for all $t\leq\tau$. Set
$
\Delta_t = y_{1:t-1}^* - y_{1:t}^* \ge 0
$
for $2\leq t\leq\tau$.
Conditional on $\mathcal F_{t-1}$ and the choice of $x_t$, we have
\begin{equation}
\E[\Delta_t \mid \mathcal F_{t-1},x_t]
      = \alpha_{t-1}^{\mathrm{EI}}(x_t).
\end{equation}      
Taking expectations and summing up from 2 to $\tau$, 
\begin{align}
\E\!\left[\sum_{t=2}^{\tau}\alpha^{\mathrm{EI}}_{t-1}(x_t)\right]
&= \E\!\left[\sum_{t=2}^{\infty}\mathbf{1}_{\{t\le \tau\}}\,
      \alpha^{\mathrm{EI}}_{t-1}(x_t)\right] \nonumber\\
&= \E\!\left[\sum_{t=2}^{\infty}\mathbf{1}_{\{t\le \tau\}}\,
      \E\!\left[\Delta_t \mid \mathcal F_{t-1}, x_t\right]\right] \nonumber
\\
&= \E\!\left[\sum_{t=2}^{\infty}\mathbf{1}_{\{t\le \tau\}}\,\Delta_t\right]       \quad(\text{tower property}) 
 \nonumber\\
&= \E\!\left[\sum_{t=2}^{\tau}\Delta_t\right]= \E\!\left[y^*_{1:1} - y^*_{1:\tau}\right]. \label{eq:max_total_improvment}
\end{align}

Summing \eqref{eq:key-LB} over $t\leq\tau$, taking expectations, and applying \eqref{eq:max_total_improvment}, we have
\begin{equation}
\mathbb{E}\left[\sum_{t=2}^{\tau}c(x_t)\right]
      \le
\mathbb{E}\left[\sum_{t=2}^{\tau}\alpha_{t-1}^{\mathrm{EI}}(x_t)\right]
      \le \E\!\left[y^*_{1:1} - y^*_{1:\tau}\right]. \label{eq:bound_on_cost} 
\end{equation}
Adding the term $y^*_{1:\tau}$ on both sides of \eqref{eq:bound_on_cost} gives
\[
\mathbb{E}\!\left[y^*_{1:\tau}+\sum_{t=2}^{\tau} c(x_t)\right]
\le \mathbb{E}[\,y^*_{1:1}\,].
\]
Finally, adding $c(x_1)$ to both sides and subtracting $\mathbb{E}\!\left[\min_{x\in X} f(x)\right]$ yields
\begin{align*}
\mathbb{E}\!\left[y^*_{1:\tau}-\min_{x\in X} f(x)+\sum_{t=1}^{\tau} c(x_t)\right]
&\le
\mathbb{E}\!\left[y^*_{1:1}-\min_{x\in X} f(x)+c(x_1)\right]
\\
&=
\mathbb{E}\!\left[y_1-\min_{x\in X} f(x)+c(x_1)\right]
\\
&=
\mu(x_1) - \mathbb{E}\!\left[\min_{x\in X} f(x)\right] + \mathbb{E}[c(x_1)] \\
&= U+C,
\end{align*}
since $\mathbb{E}[y_1]=\mu(x_1)$, $C=c(x_1)$, and $U=\mu(x_1) - \mathbb{E}\left[\min_{x\in X} f(x)\right]$. This is exactly~\eqref{eqn:cost_adjusted_regret_bound}.
\end{proof}

\begin{remark}
    The quantities $U$ and $C$ depend on the choice of the initial point $x_1$. Choosing $x_1$ to minimize $\mu(x_1)+c(x_1)$ yields a tighter bound.
\end{remark}

\begin{remark}
In practice, one may center the objective by replacing $f(x)$ with $f(x) - \mu(x)$ 
and then using a zero-mean GP prior.  
This changes only the parameterization of the posterior mean (which can be shifted back by $\mu(x)$) 
and does not affect the implementation of the acquisition functions or stopping rules.
\end{remark}

Notably, this guarantee may not hold for other acquisition–stopping rule pairings. Moreover, in the worst case, this is the best guarantee we can hope for---for instance, the evaluation costs can be uniformly high and no point is worth evaluating.
We now formalize this worst-case regime.

\begin{proposition}[Immediate stopping is optimal under large costs]
\label{prop:immediate_optimal}
Suppose the objective function is bounded as $f(x) \in [a,b]$ for all $x \in \mathcal X$, and define $B := b-a$. Consider any Bayesian optimization policy that starts from an initial evaluation $x_1$ and may take additional evaluations. If the evaluation cost satisfies
\[
c(x) \ge B \quad \text{for all } x \in \mathcal X,
\]
then the policy that stops immediately after the initial evaluation is optimal with respect to expected cost-adjusted regret.
\end{proposition}

\begin{proof}
Let $\pi_{\mathrm{imm}}$ denote the policy that stops immediately after the initial evaluation $x_1$, and let $\pi$ be any other policy with stopping time $\tau \ge 1$. Recall that the cost-adjusted simple regret is
\[
R_c(\pi)
:=
\mathbb E\!\left[
\min_{1 \le t \le \tau} f(x_t) - \min_{x \in \mathcal X} f(x)
+ \sum_{t=1}^\tau c(x_t)
\right].
\]

Since $y_1 = f(x_1)$ and $f(x) \in [a,b]$ for all $x \in \mathcal X$, we have
\[
0 \le y_1 - \min_{x \in \mathcal X} f(x) \le b-a = B.
\]

The immediate-stopping policy has $\tau=1$, so
\[
R_c(\pi_{\mathrm{imm}})
=
\mathbb E\!\left[
y_1 - \min_{x \in \mathcal X} f(x)
+ c(x_1)
\right].
\]

Now consider an arbitrary policy $\pi$.

If $\tau = 1$, then $\pi$ coincides with immediate stopping and incurs the same cost-adjusted simple regret.

If $\tau \ge 2$, then the simple regret term is nonnegative, and the policy incurs at least one additional evaluation beyond $x_1$. Hence,
\[
\min_{1 \le t \le \tau} f(x_t) - \min_{x \in \mathcal X} f(x)
+ \sum_{t=1}^\tau c(x_t)
\ge
c(x_1) + c(x_2).
\]
Using $c(x) \ge B$ for all $x$,
\[
c(x_1) + c(x_2)
\ge
c(x_1) + B
\ge
c(x_1) + \bigl(y_1 - \min_{x \in \mathcal X} f(x)\bigr).
\]

Therefore, in all cases,
\[
\min_{1 \le t \le \tau} f(x_t) - \min_{x \in \mathcal X} f(x)
+ \sum_{t=1}^\tau c(x_t)
\ge
y_1 - \min_{x \in \mathcal X} f(x) + c(x_1).
\]
Taking expectations yields
\[
R_c(\pi) \ge R_c(\pi_{\mathrm{imm}}),
\]
which proves optimality of immediate stopping.
\end{proof}

\ThmTermination*
\begin{proof}
    Immediately by \cref{thm:non_neg_util} and the fact that $y^*_{1:\tau}-\min_{x\in X} f(x) \geq 0$ pointwise, we have that the expected cumulative cost of the algorithm 
    \begin{equation}
        \mathbb{E}\left[\sum_{t=1}^{\tau} c(x_t)\right] \leq U + C. 
    \end{equation}
    By Markov's inequality, for any $\delta \in (0, 1)$,
    \begin{align*}
        \Pr\left[\sum_{t=1}^{\tau} c(x_t) \leq \frac{U + C}{\delta} \right] \geq 1 - \delta. 
    \end{align*}
    Since $c(x_t)\geq c_0$ for all $t<\tau$, it follows that 
    \begin{align*}
        \tau = \sum_{t=1}^\tau 1 \leq  \sum_{t=1}^\tau \frac{c(x_t)}{c_0} = \frac{1}{c_0} \sum_{t=1}^\tau c(x_t).  
    \end{align*}
    Therefore, 
    \begin{align*}
        \Pr\left[\tau \leq \frac{U+C}{\delta \cdot c_0}\right] \geq   \Pr\left[\frac{1}{c_0} \sum_{t=1}^{\tau} c(x_t) \leq \frac{U + C}{\delta \cdot c_0} \right] = \Pr\left[\sum_{t=1}^{\tau} c(x_t) \leq \frac{U + C}{\delta} \right]  \geq 1 - \delta. 
    \end{align*}
\end{proof}

\ThmCosts*

\begin{proof}
Since the Bayesian optimization algorithm considers the post-scaling cost, by \cref{thm:non_neg_util} and the fact that $y^*_{1:\tau}-\min_{x\in X} f(x) > 0$ pointwise, we have 
\begin{equation}
\mathbb{E}\left[\sum_{t=1}^{\tau}\lambda \cdot c(x_t)\right] \leq \mathbb{E}\!\left[y^*_{1:\tau}-\min_{x\in X} f(x)+\sum_{t=1}^{\tau} \lambda \cdot c(x_t)\right] \le \mu(x_1)-\mathbb{E}\left[\min_{x\in X}f(x)\right] + \lambda C = U + \lambda C. 
\end{equation}

Since $\lambda = U/(B-C)$, we have \[\E\left[\sum_{t=1}^{\tau} c(x_t)\right] \leq \frac{U + \lambda C}{\lambda} = \frac{U}{\lambda} + C = B.\] 
\end{proof}

\begin{remark}
In the Bayesian regret setting where $f$ is drawn from a Gaussian process, i.e., $f \sim \mathcal{GP}(\mu(\cdot), K)$, the term $U = \mu(x_1) - \mathbb{E}\left[\min_{x\in X} f(x)\right]$ can be further bounded above using classical results on the expected supremum/infimum of Gaussian processes; see, for example, \citet[Theorem 10.1]{lifshits2012lectures}.
In many practical settings, the objective is naturally bounded. For example, in our AutoML experiments the objective is accuracy, which lies in $[0,100]$. This implies
\[
U = \mu(x_1) - \mathbb{E}\left[\min_{x\in X} f(x)\right] \le 100,
\]
providing a simple and interpretable upper bound on $U$.
\end{remark}

\section{Experimental Setup}
\label{apdx:experiment_setup}
All experiments are implemented based on BoTorch \citep{balandat2020botorch}. Each
Bayesian optimization procedure is initialized with $2(d + 1)$ random samples, where $d$ is
the dimension of the search domain. For Bayesian regret experiments, we follow the standard practice to generate the initial random samples using a quasirandom Sobol sequence. For empirical experiments, we randomly sample configuration IDs from a fixed pool of candidates—2,000 for LCBench and 32,768 for NATS-Bench. All computations are performed on CPU.

Each experiment is repeated with 50 random seeds, and we report the mean with error bars, given by two times the standard error, for each stopping rule. We also impose a cap on the number of iterations: 100 for 1D Bayesian regret, 500 for 8D Bayesian regret, and 200 for empirical experiments. If a stopping rule is not triggered before reaching this cap, we treat the stopping time as equal to the cap.

\paragraph*{Gaussian process models.}
For all experiments, we follow the standard practice to apply Matérn kernels with smoothness $5/2$ and length scales learned from data via maximum marginal likelihood optimization, and standardize input variables to be in $[0,1]^d$. For empirical experiments, we standardize output variables to be zero-mean and unit-variance, but not for Bayesian regret experiments.
In this work, we consider the noiseless setting and set the fixed noise level to be $10^{-6}$.
In the unknown-cost experiments, we follow \citet{astudillo2021multi} to model the objective and the logarithm of the cost function using independent Gaussian processes.

\paragraph*{Acquisition function optimization.} For the 1D Bayesian regret experiments, we optimize over 10,001 grid points. For the 8D Bayesian regret experiment, we use BoTorch’s `gen\_candidates\_torch', a gradient-based optimizer for continuous acquisition function maximization, as it avoids reproducibility issues caused by internal randomness in the default scipy optimizer. For the empirical experiments, since LCBench and NATS-Bench provide only 2,000 and 32,768 configurations respectively, we optimize the acquisition function by simply applying an argmin/argmax over the acquisition values of the unevaluated configurations, without using any gradient-based methods.

\paragraph*{Acquisition function and stopping rule parameters.}
For PBGI, we follow \citet{xie2024cost} to compute the Gittins indices using $100$ iterations of bisection search without any early stopping or other performance and reliability optimizations.

For UCB/LCB-based acquisition functions and stopping rules, we follow the original GP-UCB paper \citet{srinivas2009gaussian} and the choice in UCB/LCB based stopping rules \citep{makarova2022automatic,ishibashi2023stopping} to use the schedule $\beta_t=2\log(d t^2\pi^2/6\delta)$, where $d$ is the dimension.
We also adopt their choice of $\delta=10^{-1}$ and a scale-down factor of $5$.

For PRB, we follow \citet{wilson2024stopping} to use the schedule $N_t=\max(\lceil 64*1.5^{t-1}\rceil, 1000)$ for number of posterior samples, risk tolerance $\delta=0.05$. The error bound $\epsilon$ is set to be 0.1 for Bayesian regret experiments and 0.5\% of the best test error (here, the misclassification rate) among all configurations for the empirical experiments.

\begin{figure}[t!]
    \centering   \includegraphics[width=\textwidth]{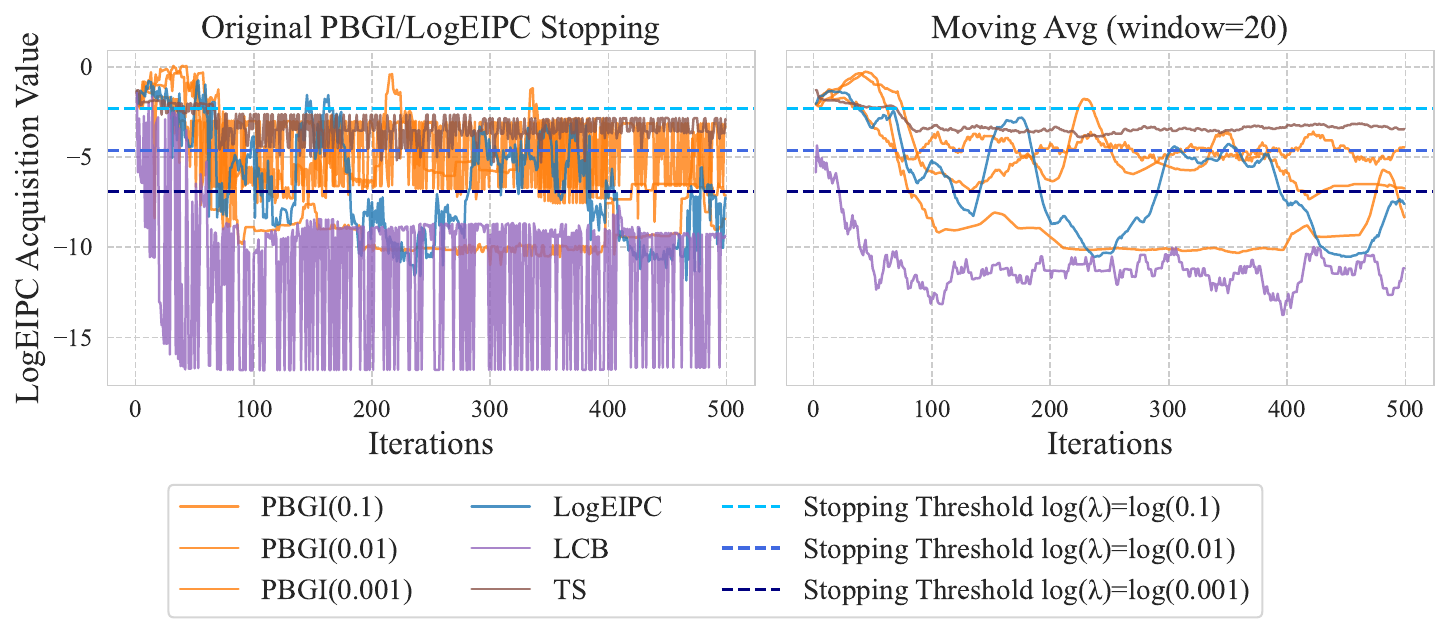}
    \caption{Comparison of the raw and moving-averaged PBGI/LogEIPC stopping rule signals (i.e., the LogEIPC acquisition values) in Bayesian regret 8D experiments for multiple acquisition functions, with linear cost and stopping thresholds at $\log(0.1)$, $\log(0.01)$ and $\log(0.001)$. \textit{Left:} The unaveraged signals exhibits large, high-frequency fluctuations due to the difficulty of acquisition optimization in high dimensions. \textit{Right:} Applying moving average (window=20) smooths these wiggles, yielding more stable stopping signals.}
    \label{fig:moving_average}
\end{figure}

For the 8D Bayesian regret experiments, for all acquisition-value-based stopping rules, we apply moving average over 20 iterations to mitigate the fluctuations due to the imperfect acquisition function optimization. \Cref{fig:moving_average} illustrates the challenges these oscillations pose when computing stopping rule statistics and shows the improvement with moving average. For consistency, we also apply the 20-iteration averaging to the non-acquisition-value-based GSS and Convergence baselines.

\paragraph{Omitted baselines.} We omit the KG acquisition function and stopping rule due to its high computational cost, as they are shown to be computationally intensive in the runtime experiments of \citet{xie2024cost}.

\paragraph{Objective functions: Bayesian regret.}
In all Bayesian regret experiments, each objective function $f$ is sampled from a Gaussian process prior with a Matérn 5/2 kernel and a length scale of $0.1$, using a different seed in each of the 50 trials.

\paragraph*{Objective functions: empirical.} 
In the empirical experiments, the validation error (scaled out of 100) is used as the objective function during the Bayesian optimization procedure, while the cost-adjusted simple regret is reported based on the corresponding test error.

\paragraph*{Cost functions: Bayesian regret.}  
In Bayesian regret experiments, we consider three types of evaluation costs: uniform, linear, and periodic. These costs are normalized such that $\E_{x \in [0,1]^d}[c(x)]$ is approximately $1$ and their expressions (prior to cost scaling) are given below.

In the \textit{uniform} cost setting, each evaluation incurs a constant cost of $1$.

In the \textit{linear} cost setting, the cost increases proportionally with the average coordinate value of the input:
\[
\text{linear\_cost}(x) = \frac{1 + 20 \cdot \left( \frac{1}{d} \sum_{i=1}^{d} x_i \right)}{11}.
\]
In the \textit{periodic} cost setting, the evaluation cost fluctuates across the domain. Following \cite{astudillo2021multi}, we define the periodic cost as
\[
\text{periodic\_cost}(x) = \frac{ \exp\left( \frac{\alpha}{d} \sum_{i=1}^{d} \cos\left( 2\pi \beta (x_i - x^*_i) \right) \right) }{ \left[ I_0\left( \frac{\alpha}{d} \right) \right]^d },
\]
where $x^*_i$ denotes the coordinate of the global optimum of $f$, and $I_0$ is the modified Bessel function of the first kind, which acts as a normalization constant. We set $\alpha = 2$ and $\beta = 2$ to induce noticeable variation in cost across the domain, while ensuring that costly evaluations can still be worthwhile. 

\begin{figure}[t!]
    \centering   \includegraphics[width=0.9\textwidth]{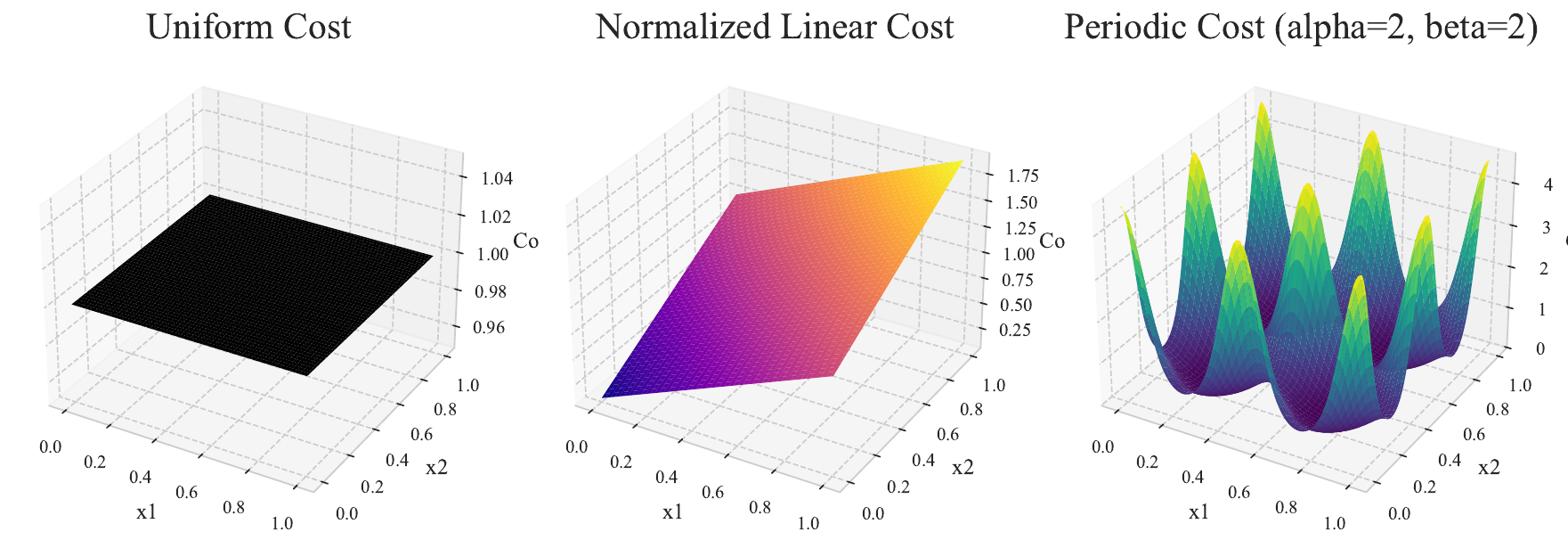}
    \caption{Surface plots of cost functions over $[0, 1]^2$: (Left) Uniform cost of $1$ across domain. (Middle) Normalized linear cost increasing with the mean of $x_1$ and $x_2$. (Right) Periodic cost with $\alpha=2$, $\beta=2$, normalized by Bessel-based factor. }
    \label{fig:cost-surfaces}
\end{figure}

A visualization of the three cost functions is provided in \Cref{fig:cost-surfaces}.

\paragraph{Cost functions: empirical.}
In the \emph{unknown-cost} experiments, we treat runtime---meaning, the provided full model training time (200 epochs for LCBench and 90 epochs for NATS)---as evaluation costs (prior to cost scaling).

\begin{figure}[t!]
    \centering   \includegraphics[width=\textwidth]{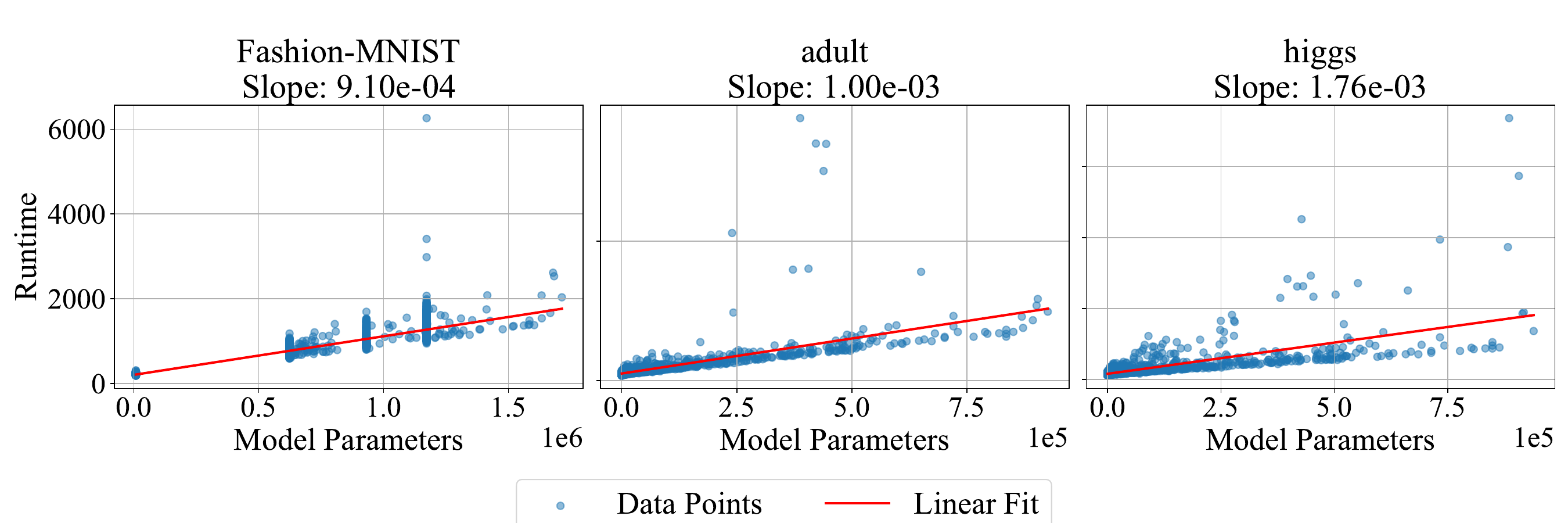}
    \caption{Empirical relationship between the number of model parameters and runtime for three LCBench datasets. Each subplot shows a scatter plot of actual runtime ($y$-axis) against number of model parameters ($x$-axis), along with a fitted linear regression line. The observed linear trend supports using 0.001 times the number of model parameters as a proxy for runtime. For \emph{Fashion-MNIST} and \emph{adult}, the fitted slopes are close to 0.001. The slope for higgs is slightly higher, possibly due to a few outliers.
    }
    \label{fig:runtime_proxy}
\end{figure}

In the \emph{known-cost} experiments, for LCBench, we estimate the runtime cost from the number of model parameters $p$. Specifically, for the first three datasets in the six-dataset lite version, we use 0.001 times the number of model parameters as a proxy for runtime. This proxy is motivated by our observation of an approximately linear relationship between the number of model parameters and the actual runtime, with slope close to 0.001 (see Figure~\ref{fig:runtime_proxy}). For the full 35-dataset version, where the slope varies across datasets, we instead use a regression-derived coefficient $\alpha$ and the proxy cost is $\alpha p$. Importantly, the number of model parameters can be computed in advance, before the Bayesian optimization procedure, based on the network structure and classification task, as we explain in detail below.

In a feedforward neural network like shapedmlpnet with shape `funnel', the model parameters are determined by input size (number of features), output size (e.g., number of classes), number of layers, size of each layer.
The input size and output size are given by:
\begin{itemize}
    \item Fashion-MNIST:
    \begin{itemize}
        \item Input dimension: 784 (each image has 28×28 pixels, flattened into a vector of length 784)
        \item Number of Output Features (output\_feat): 10 (corresponding to 10 clothing categories)
    \end{itemize}
    \item Adult:
    \begin{itemize}
        \item Input Dimension: 14 (the dataset comprises 14 features, including both numerical and categorical attributes)
        \item Number of Output Features: 1 (binary classification: income $>50K$ or $\leq50K$)
    \end{itemize}
    \item Higgs: 
    \begin{itemize}
        \item Input Dimension: 28 (each instance has 28 numerical features)
        \item Number of Output Features: 1 (binary classification: signal or background process)
    \end{itemize}
\end{itemize}
The number of layers (num\_layers) and size of each layer (max\_unit) can be obtained from the configuration. With these information, we can compute the total number of model parameters (weights and biases) based on the layer-wise structure as follows:
\begin{align}
\text{layer\_params}_{i \to i+1} &= \text{layer}_i \cdot \text{layer}_{i+1} + \text{layer}_{i+1} \quad \text{(weights + biases)} \\
\text{model\_params} &= \sum_{i=0}^{L - 1} \text{layer\_params}_{i \to i+1} \\
\text{where} \quad &\text{layer}_0 = \text{input\_dim}, \quad \text{layer}_L = \text{output\_feat}
\end{align}

\begin{figure}[t!]
    \centering   \includegraphics[width=\textwidth]{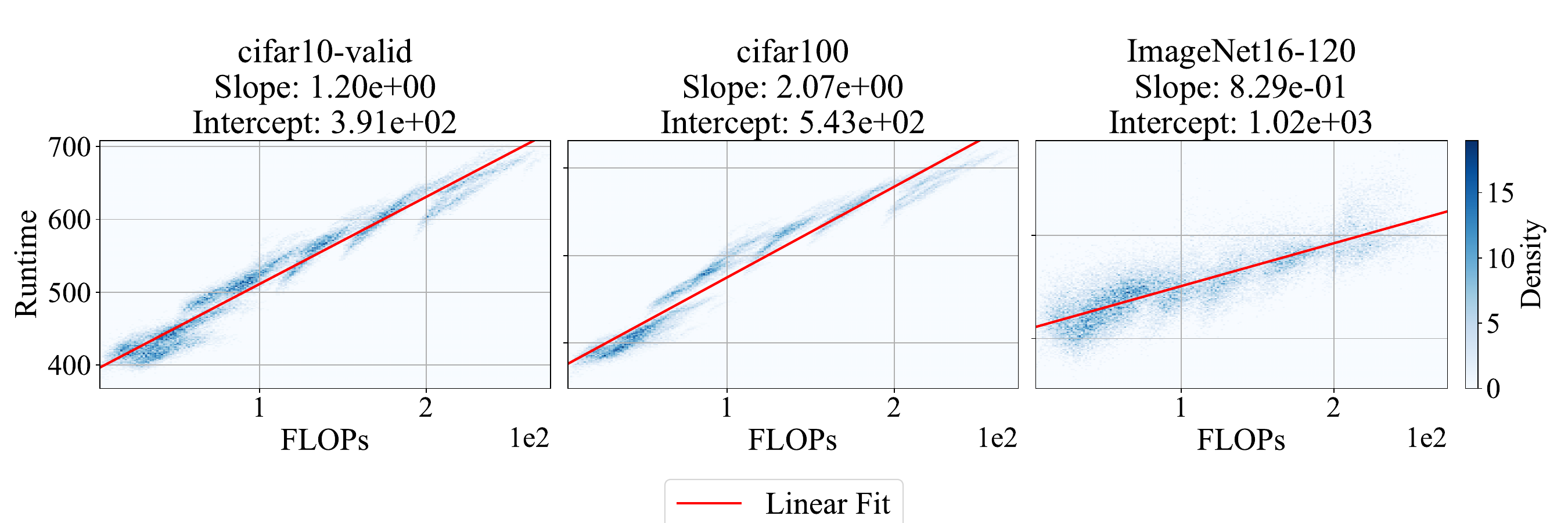}
    \caption{Empirical relationship between the number of FLOPs and runtime for the three NATS-Bench datasets. Each subplot shows a heatmap of actual runtime ($y$-axis) against number of FLOPs ($x$-axis), along with a fitted linear regression line. The observed linear trend supports using $\alpha \times \#\text{FLOPs} + \beta$ as a proxy for runtime.
    }
    \label{fig:NATS_runtime_proxy}
\end{figure}

Similarly for NATS-Bench, we use $\alpha \times F + \beta$ as a proxy for runtime (see Figure~\ref{fig:NATS_runtime_proxy} for a visualization of the linear relationship), where $F$ is the number of floating point operations (FLOPs).
Specifically, for \emph{cifar10-valid}, we set $\alpha=1$, $\beta=400$; for \emph{cifar100}, we set $\alpha=2$, $\beta=550$; and for \emph{ImageNet16-120}, we set $\alpha=1$, $\beta=1000$.

FLOPs can also be computed in advance, as it is determined solely by the architecture's structure and the fixed input shape. Specifically, they are precomputed and stored for each architecture. Since each architecture corresponds to a deterministic computational graph and all inputs (e.g., CIFAR-10 images) have a fixed shape, the FLOPs required for a forward pass can be calculated analytically—without executing the model on data.

\section{Additional Experiment Results}
\label{apdx:experiment_results}
In this section, we present additional experimental results to further evaluate the performance of different acquisition-function--stopping-rule pairs across various settings. We also include alternative visualizations to aid interpretation of the results. 

\subsection{Runtime Comparison}
First, we compare the runtime between our PBGI/LogEIPC stopping rule with several baselines. We measure the CPU time
(in seconds) of the computation of the stopping rule, excluding the acquisition function computation and optimization.

\begin{figure}[t!]
    \centering
    \includegraphics[width=\textwidth]{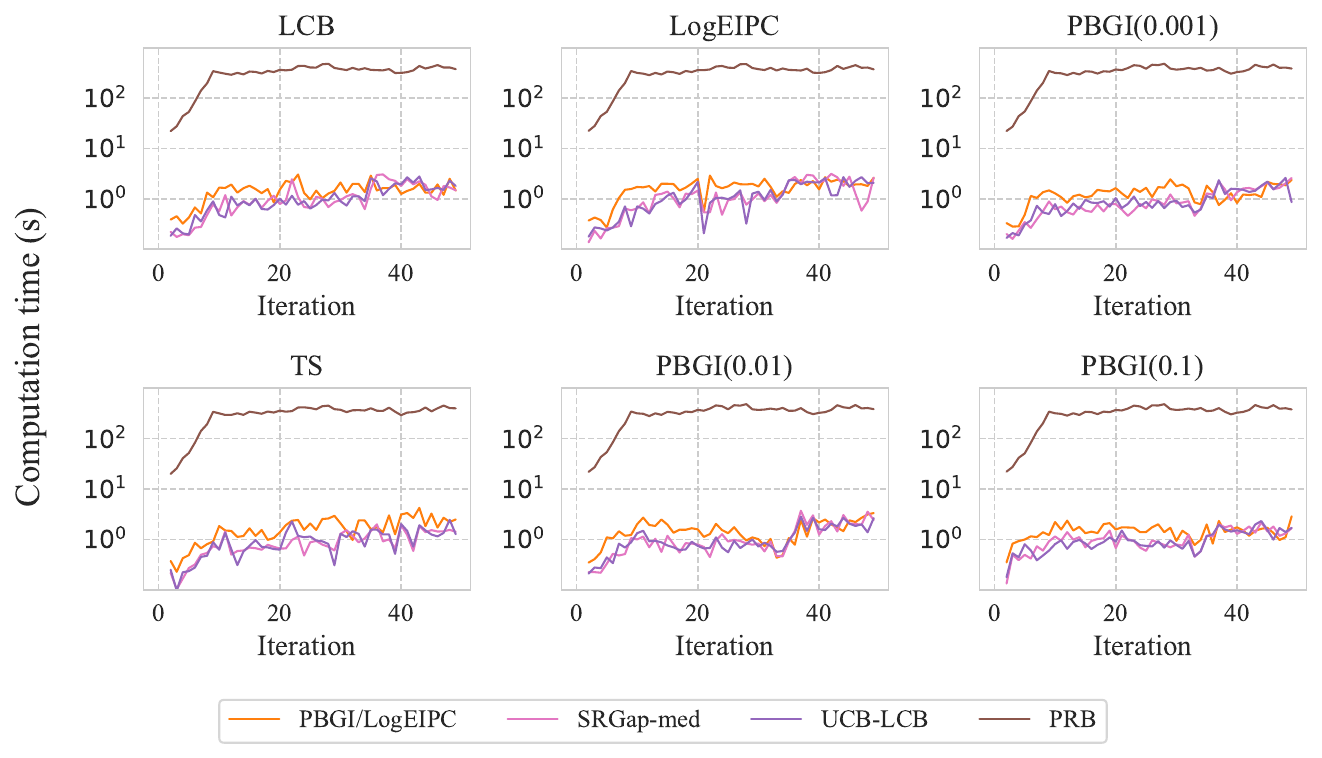}
    \caption{Evolution of per-iteration computation time (in log scale) for different stopping stopping rules when paired with six acquisition policies on the 8-dimensional Bayesian regret benchmark. Each subplot shows the average runtime (in seconds) over 50 iterations under one acquisition function—LCB, TS, LogEIPC, PBGI ($\lambda=10^{-1}$), PBGI ($\lambda=10^{-2}$), and PBGI ($\lambda=10^{-3}$). Curves correspond to four stopping criteria: our PBGI/LogEIPC stopping rule, SRGap-med, UCB–LCB, and PRB. Convergence and GSS can be applied using only the best observed value and thus require no additional computation time, thus they are omitted here. LogEIPC-med relies on the same underlying statistical computations as the LogEIPC rule, and therefore its runtime is not measured separately. The results should that PRB incurs significant computational overhead compared to other stopping rules.}
    \label{fig:runtime}
\end{figure}

From results in Figure \ref{fig:runtime} we can see that our PBGI/LogEIPC stopping rule is roughly as efficient as SRGap-med and UCB–LCB. In contrast, PRB is significantly more time-consuming, as it involves optimizing over up to 1000 samples.

\subsection{Order of Stopping and Posterior Updates} \label{apdx:sec:order_posterior_update}

\begin{figure}[t!]
    \centering
    \includegraphics[width=\textwidth]{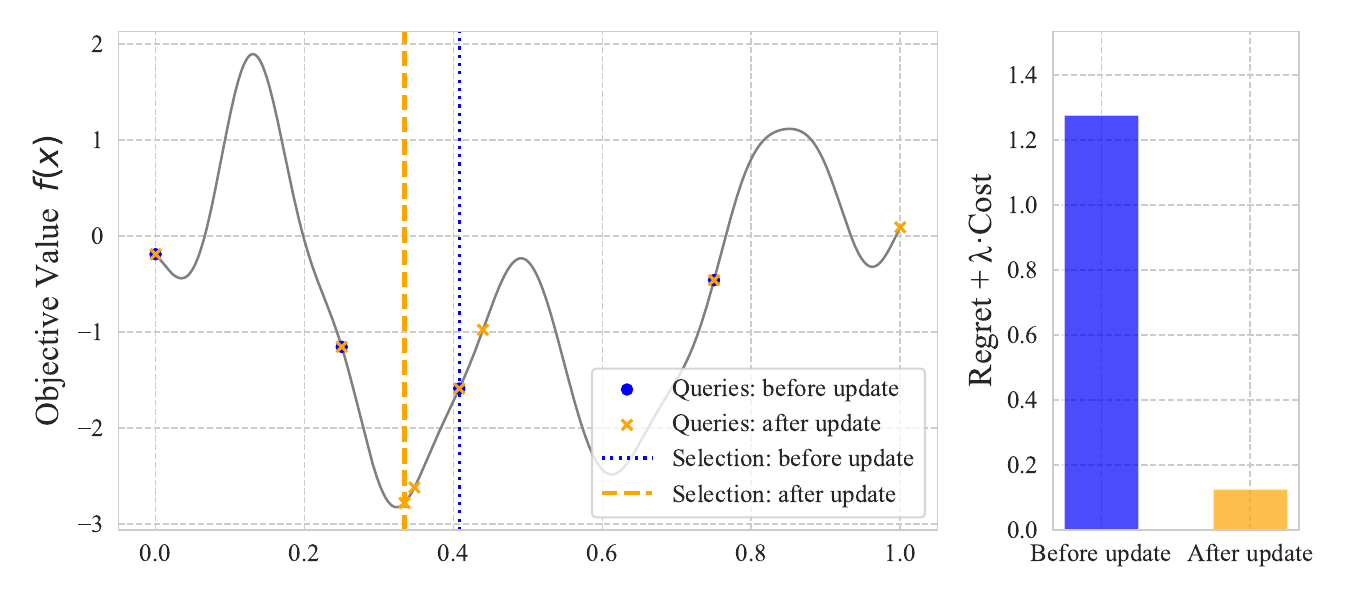}
    \caption{Illustration of a single draw from a Mat\'ern-$5/2$ Gaussian process on $[0,1]$ with lengthscale $0.1$, optimized using PBGI acquisition function under uniform cost and cost-scaling factor $\lambda=0.01$. We compare two variants of PBGI stopping rules: the \textit{before-posterior-update} (this-round) stopping rule and the \textit{after-posterior-update} (next-round) stopping rule.  \textbf{Left:} The latent objective function (solid gray) and evaluation sequences for \textit{before-posterior-update} stopping (blue circles) and \textit{after-posterior-update} stopping (orange crosses). The dotted blue line and the dashed orange line mark the best observed value under each respective rule. \textbf{Right:} Cost‐adjusted regret for each stopping rule. In this example, \emph{after-posterior-update} stopping achieves strictly lower cost-adjusted regret despite performing more evaluations.}
    \label{fig:pbgi_stopping_discussion}
\end{figure}

Following the discussions in \Cref{sec:method}, we always compute our proposed stopping rule with respect to the optimal acquisition function value of the \emph{next round}---namely, the one which is obtained after posterior updates have been performed.
One could alternatively consider checking the stopping criteria \emph{before} posterior updates, which is backward-looking rather than forward-looking.
\Cref{fig:pbgi_stopping_discussion} provides an empirical comparison between the two choices, showing that stopping \emph{after} the posterior update leads to stronger empirical performance.

This suggests that the theoretical guarantee for the Gittins index policy in the correlated Pandora's Box setting by \citet{gergatsouli2023weitzman}, which is based on the \emph{before-posterior-update} stopping, could potentially be improved by adopting the \emph{after-posterior-update} stopping.

\subsection{Additional Experiment Results: Empirical}
\label{apdx:sec:empirical_results}
This subsection presents additional results for hyperparameter optimization on the LCBench datasets and neural architecture–size search on the three NATS-Bench datasets. For LCBench, we provide results for the first three datasets from the six-dataset lite version of the benchmark: Fashion-MNIST, adult, and higgs under alternative settings (fixed budget, actual runtime, and unknown cost), along with an alternative visualization. We also include per-dataset results under the proxy-runtime setting for the full set of 35 datasets.

\subsubsection{Simple regret under the fixed-budget setting.} To isolate the effect of the acquisition function on cost-adjusted regret, we report the simple regret of several acquisition functions in the fixed-budget setting. We compare PBGI, LogEIPC, LCB, and TS, and additionally include PBGI-D with our recommended choice $\lambda_0 = U/(B-C)$ from \Cref{subsubsec:budget-constrained}, where $U=50$ (reflecting the [0,100] range of classification accuracy) and $B-C=10{,}000$ (corresponding to a budget after initial evaluation of 10,000 seconds, or roughly 3 hours). Cost-aware methods (the PBGI variants and LogEIPC) outperform cost-unaware ones (UCB and TS), with PBGI at smaller $\lambda$ consistently better than LogEIPC. This mirrors the findings of \citet{xie2024cost} and helps explain the strong performance of the matched PBGI combination in our main experiments.
\begin{figure}[t!]
    \centering
    \includegraphics[width=\textwidth]{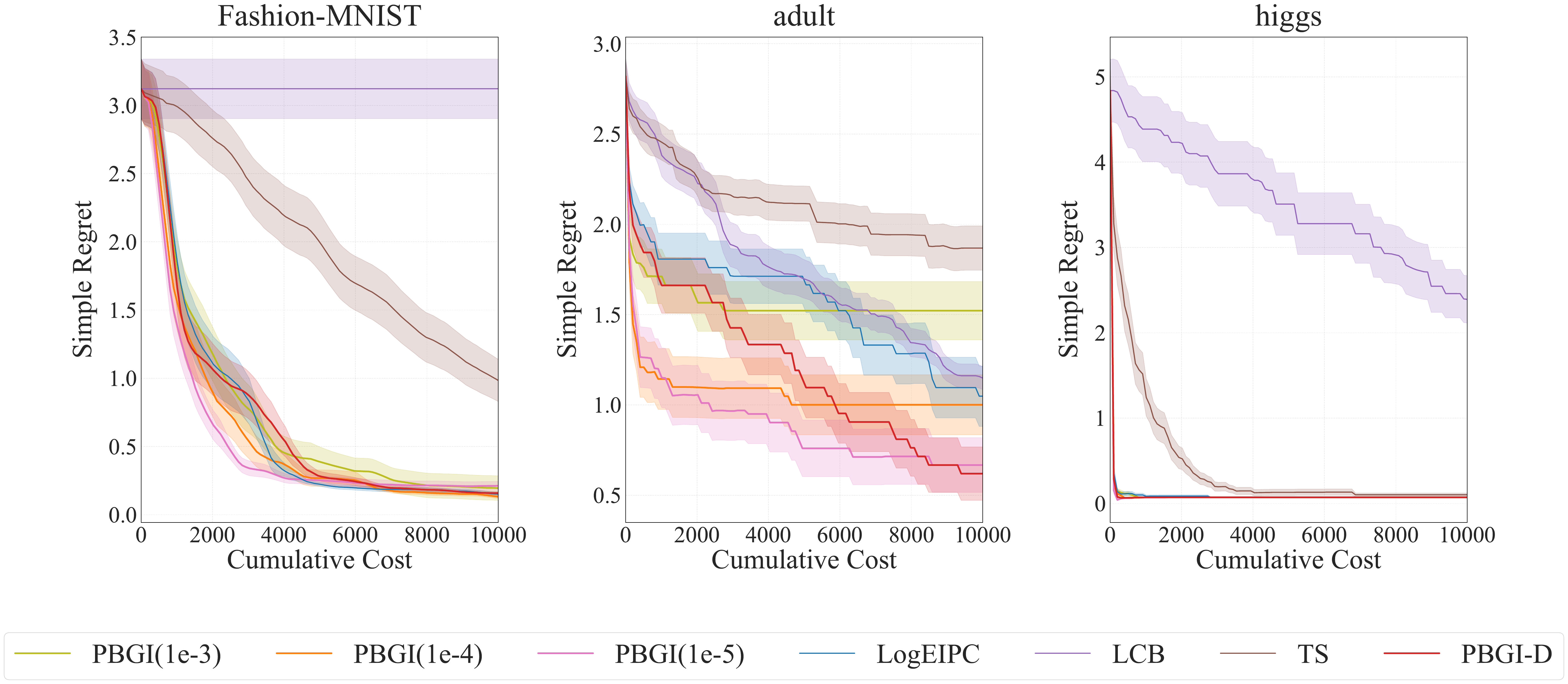}
    \caption{Comparison of simple regret of seven acquisition functions:  PBGI ($\lambda=10^{-3}$), PBGI ($\lambda=10^{-4}$), PBGI ($\lambda=10^{-4}$), LogEIPC, LCB, TS, and PBGI-D on LCBench datasets, using scaled proxy runtime as evaluation cost. We can see that indeed the cost-aware PBGI and LogEIPC outperform the cost-unaware UCB and TS. PBGI-D with our recommended $\lambda_0 = U/(B-C) = 50 / 10000 = 0.005$ is also competitive.}
    \label{fig:budgeted}
\end{figure}

\subsubsection{Number of trials where stopping fails.}
\Cref{tab:nonstops} reports the number of trials that do not stop before iteration 200, separately for each acquisition--stopping rule pair. Each rule is evaluated using the observations and model diagnostics of its paired acquisition function. The counts depend on this pairing. On NATS-Bench, our rule reaches the cap in 0, 3, and 0 of the 50 trials on cifar10-valid, cifar100, and ImageNet16-120, respectively, with either PBGI or LogEIPC, while it frequently fails to stop before the iteration cap with LCB or TS. Several baseline rules, including SRGap-med and UCB--LCB, also frequently reach the cap. Early stopping observed with the matched cost-aware acquisition functions should therefore not be generalized to arbitrary acquisition pairings.

\begin{table*}[t]
\centering
\caption{Number of trials (out of 50) that do not stop before iteration 200, for each acquisition--stopping rule pair. Online stopping is evaluated from iteration 20. LCBench uses $\lambda=10^{-4}$ (first three datasets), and NATS-Bench uses $\lambda=10^{-5}$ (last three). ``Ours'' denotes the PBGI/LogEIPC stopping rule.}
\label{tab:nonstops}
\small
\setlength{\tabcolsep}{4pt}
\renewcommand{\arraystretch}{1.05}
\resizebox{\textwidth}{!}{%
\begin{tabular}{@{}llrrrrrrr@{}}
\toprule
\textbf{Dataset} & \textbf{Acq} & \textbf{Ours} &
\shortstack{\textbf{LogEIPC-}\\\textbf{med}} &
\shortstack{\textbf{SRGap-}\\\textbf{med}} &
\textbf{UCB--LCB} & \textbf{GSS} & \textbf{Convergence} & \textbf{PRB} \\
\midrule
\multirow{4}{*}{Fashion-MNIST} & PBGI & 0 & 0 & 0 & 50 & 0 & 0 & 0 \\
 & LogEIPC & 0 & 0 & 17 & 50 & 0 & 0 & 0 \\
 & LCB & 50 & 50 & 25 & 50 & 0 & 0 & 50 \\
 & TS & 30 & 7 & 0 & 50 & 0 & 0 & 6 \\
\midrule
\multirow{4}{*}{adult} & PBGI & 5 & 0 & 7 & 44 & 0 & 0 & 6 \\
 & LogEIPC & 5 & 0 & 16 & 41 & 0 & 0 & 7 \\
 & LCB & 50 & 50 & 31 & 50 & 0 & 0 & 50 \\
 & TS & 48 & 25 & 7 & 45 & 0 & 0 & 28 \\
\midrule
\multirow{4}{*}{higgs} & PBGI & 0 & 0 & 31 & 50 & 0 & 0 & 0 \\
 & LogEIPC & 0 & 0 & 46 & 50 & 0 & 0 & 1 \\
 & LCB & 50 & 42 & 42 & 50 & 0 & 0 & 50 \\
 & TS & 50 & 26 & 7 & 50 & 0 & 0 & 7 \\
\midrule
\multirow{4}{*}{cifar10-valid} & PBGI & 0 & 32 & 50 & 50 & 0 & 0 & 34 \\
 & LogEIPC & 0 & 3 & 50 & 50 & 0 & 0 & 2 \\
 & LCB & 50 & 50 & 44 & 50 & 0 & 0 & 50 \\
 & TS & 30 & 47 & 49 & 50 & 0 & 0 & 50 \\
\midrule
\multirow{4}{*}{cifar100} & PBGI & 3 & 17 & 50 & 50 & 0 & 0 & 3 \\
 & LogEIPC & 3 & 11 & 50 & 50 & 0 & 0 & 10 \\
 & LCB & 50 & 50 & 49 & 50 & 0 & 0 & 50 \\
 & TS & 50 & 50 & 48 & 50 & 0 & 0 & 50 \\
\midrule
\multirow{4}{*}{ImageNet16-120} & PBGI & 0 & 23 & 50 & 50 & 0 & 0 & 1 \\
 & LogEIPC & 0 & 5 & 50 & 50 & 0 & 0 & 1 \\
 & LCB & 50 & 50 & 48 & 50 & 0 & 0 & 50 \\
 & TS & 49 & 50 & 49 & 50 & 0 & 0 & 42 \\
\bottomrule
\end{tabular}%
}
\end{table*}

\subsubsection{Alternative visualization: cost-adjusted simple regret vs iteration.}
We provide an alternative visualization of cost-adjusted simple regret by plotting its mean and error bars at fixed iterations, along with the mean and error bars of the stopping iterations for each rule. This allows us to compare the performance of adaptive stopping rules not only against the hindsight-optimal adaptive stopping but also against the hindsight-optimal fixed-iteration stopping.

\begin{figure}[t!]
    \centering
    \includegraphics[width=0.9\textwidth, clip, trim={0pt 0pt 0pt 5.5em}]{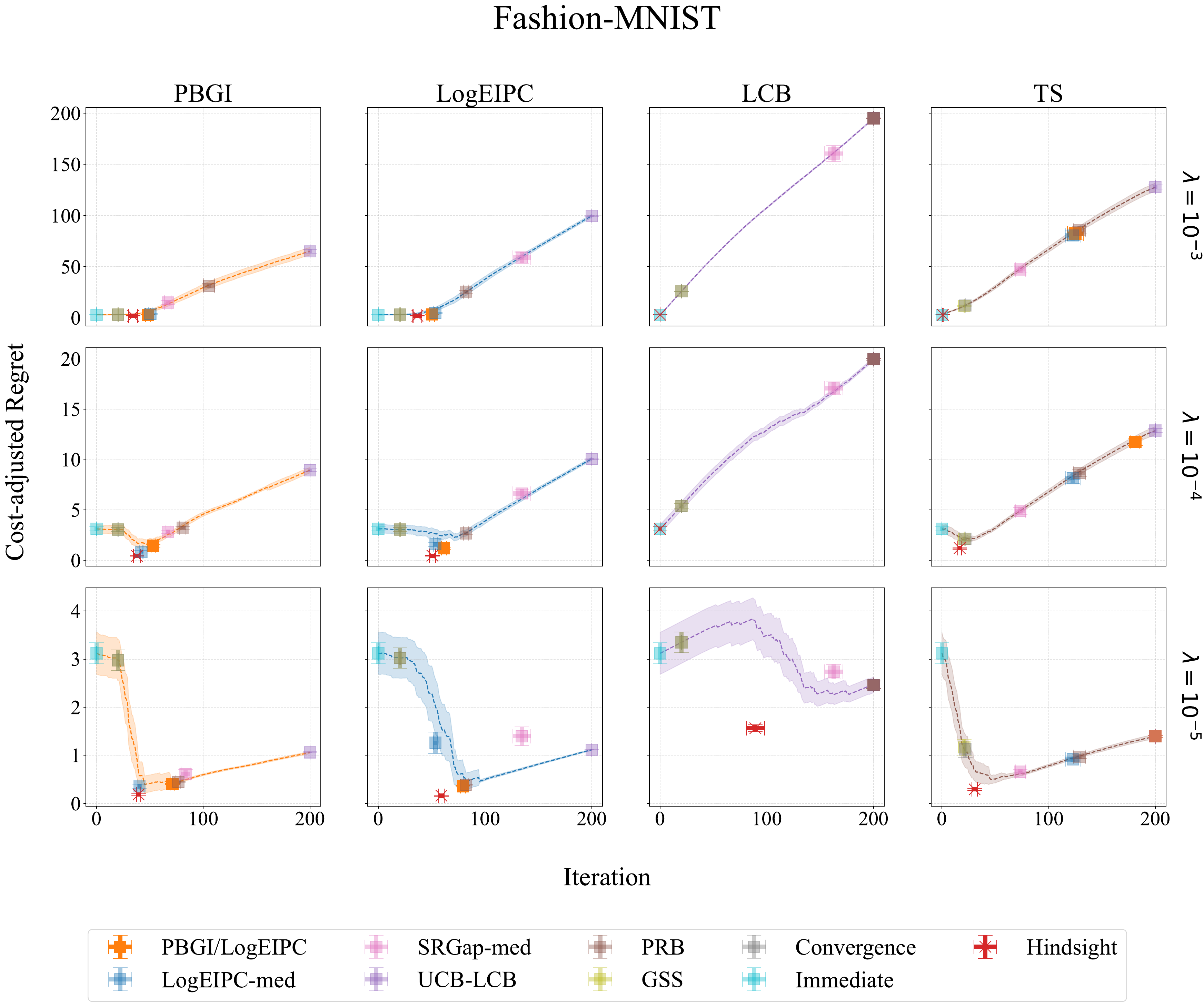}
    \caption{Comparison of cost-adjusted simple regret across acquisition function and stopping rule pairs on the \emph{Fashion-MNIST} dataset. The objective function is the validation error, and the evaluation cost is the proxy runtime, scaled by $\lambda = 10^{-3}, 10^{-4}, 10^{-5}$. Each stopping rule is evaluated on the trajectory of its paired acquisition function. The red marker shows hindsight-optimal adaptive stopping, while the minimum of the fixed-iteration curve gives the hindsight-optimal fixed iteration. These references distinguish the effects of acquisition, stopping, and cost scaling.}
    \label{fig:Fashion-MNIST}
\end{figure}

\begin{figure}[t!]
    \centering
    \includegraphics[width=0.9\textwidth, clip, trim={0pt 0pt 0pt 5.5em}]{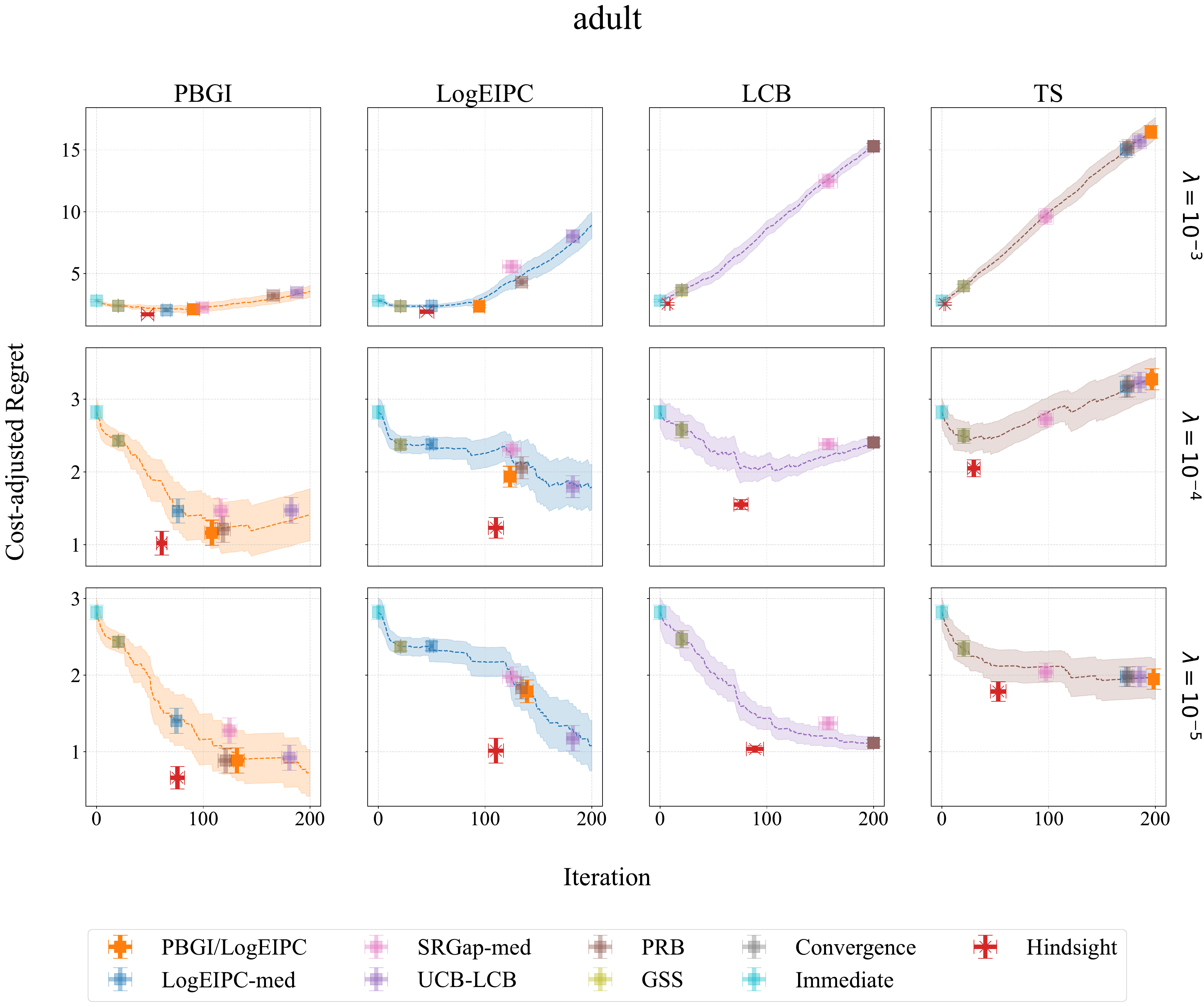}
    \caption{Comparison of cost-adjusted simple regret across acquisition function and stopping rule pairs on the \emph{adult} dataset. The objective function is the validation error, and the evaluation cost is the proxy runtime, scaled by $\lambda = 10^{-3}, 10^{-4}, 10^{-5}$. Each stopping rule is evaluated on the trajectory of its paired acquisition function. The red marker shows hindsight-optimal adaptive stopping, while the minimum of the fixed-iteration curve gives the hindsight-optimal fixed iteration. These references distinguish the effects of acquisition, stopping, and cost scaling.}
    \label{fig:adult}
\end{figure}

\begin{figure}[t!]
    \centering
    \includegraphics[width=0.9\textwidth, clip, trim={0pt 0pt 0pt 5.5em}]{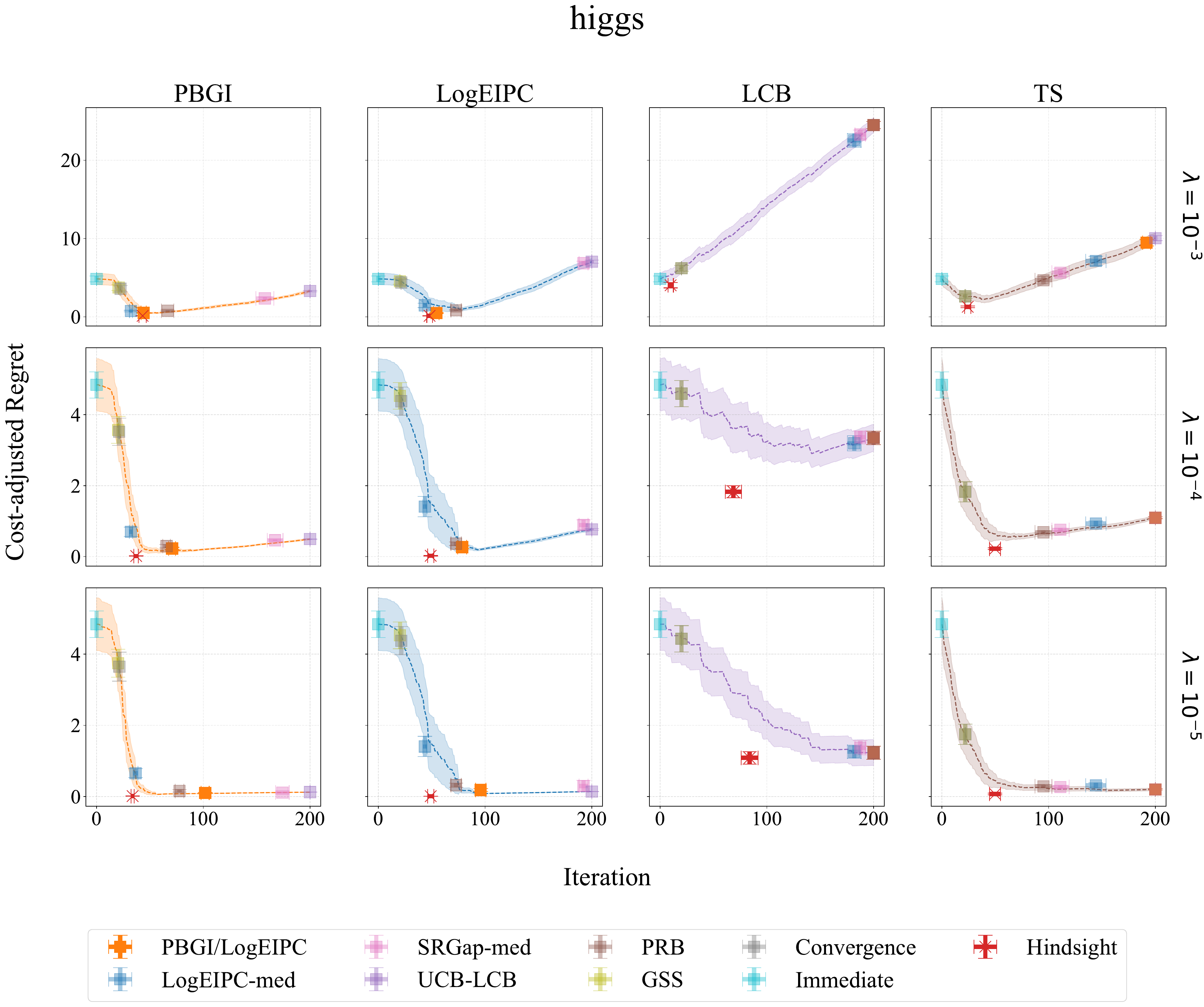}
    \caption{Comparison of cost-adjusted simple regret across acquisition function and stopping rule pairs on the \emph{higgs} dataset. The objective function is the validation error, and the evaluation cost is the proxy runtime, scaled by $\lambda = 10^{-3}, 10^{-4}, 10^{-5}$. Each stopping rule is evaluated on the trajectory of its paired acquisition function. The red marker shows hindsight-optimal adaptive stopping, while the minimum of the fixed-iteration curve gives the hindsight-optimal fixed iteration. These references distinguish the effects of acquisition, stopping, and cost scaling.}
    \label{fig:higgs}
\end{figure}

\begin{figure}[t!]
    \centering
    \includegraphics[width=0.9\textwidth, clip, trim={0pt 0pt 0pt 5em}]{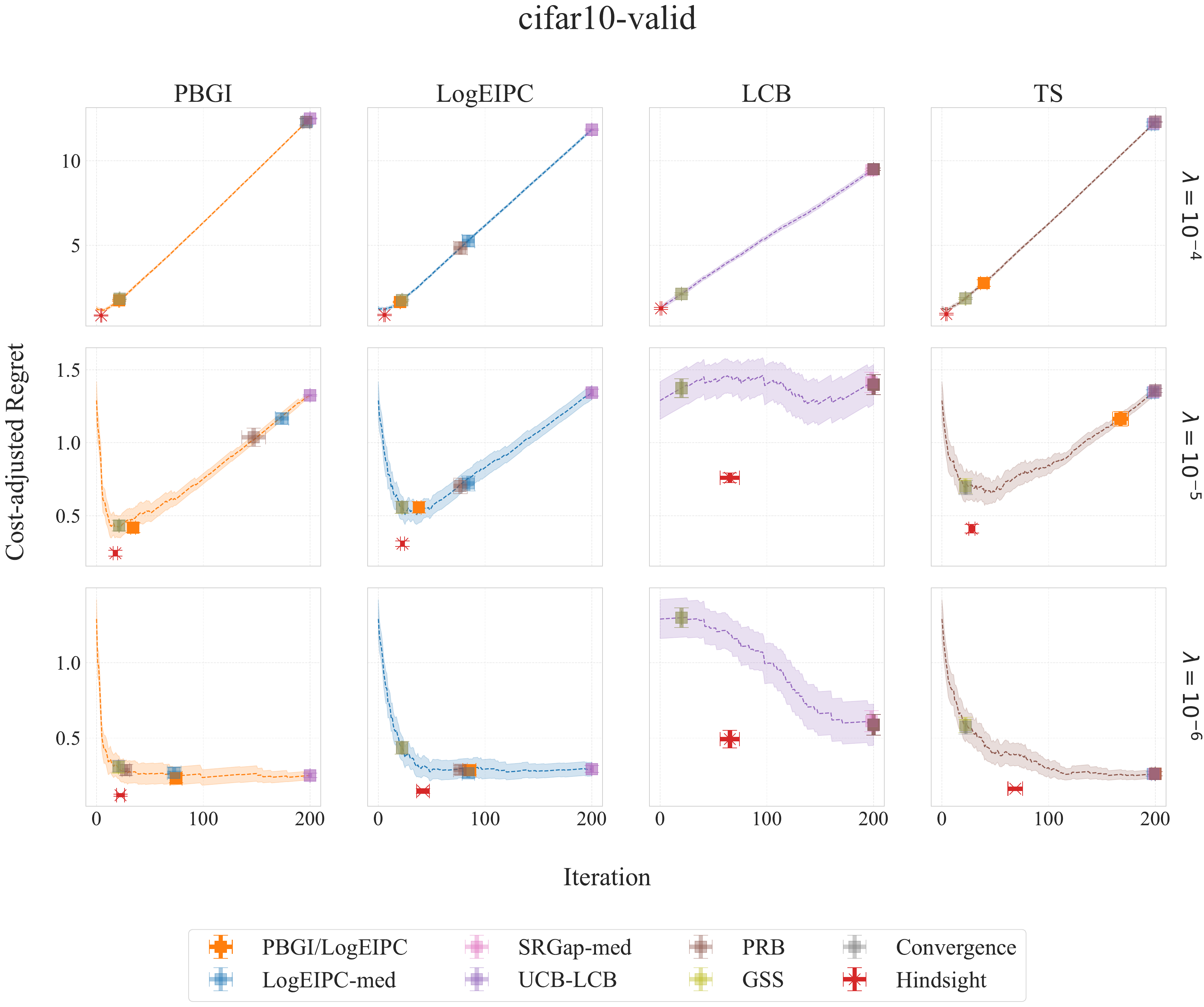}
    \caption{Comparison of cost-adjusted simple regret across acquisition function and stopping rule pairs on the \emph{cifar10-valid} dataset. The objective function is the validation error, and the evaluation cost is the proxy runtime, scaled by $\lambda = 10^{-4}, 10^{-5}, 10^{-6}$. Each stopping rule is evaluated on the trajectory of its paired acquisition function. The red marker shows hindsight-optimal adaptive stopping, while the minimum of the fixed-iteration curve gives the hindsight-optimal fixed iteration. These references distinguish the effects of acquisition, stopping, and cost scaling.}
    \label{fig:cifar10-valid}
\end{figure}

\begin{figure}[t!]
    \centering
    \includegraphics[width=0.9\textwidth, clip, trim={0pt 0pt 0pt 5.5em}]{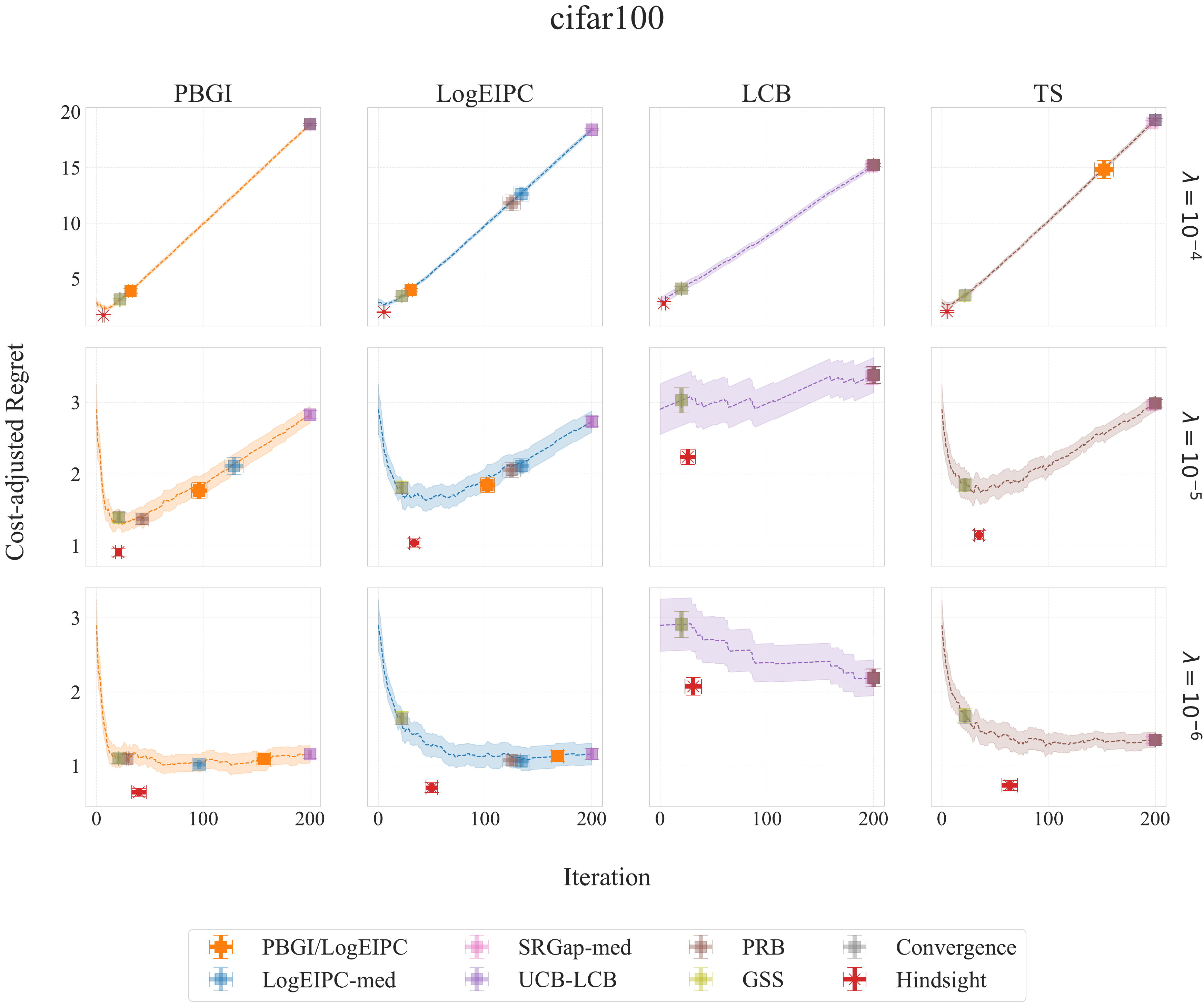}
    \caption{Comparison of cost-adjusted simple regret across acquisition function and stopping rule pairs on the \emph{cifar100} dataset. The objective function is the validation error, and the evaluation cost is the proxy runtime, scaled by $\lambda = 10^{-4}, 10^{-5}, 10^{-6}$. Each stopping rule is evaluated on the trajectory of its paired acquisition function. The red marker shows hindsight-optimal adaptive stopping, while the minimum of the fixed-iteration curve gives the hindsight-optimal fixed iteration. These references distinguish the effects of acquisition, stopping, and cost scaling. At $\lambda=10^{-5}$, GSS and Convergence have lower mean cost-adjusted regret than our rule when paired with either PBGI or LogEIPC.}

    \label{fig:cifar100}
\end{figure}

\begin{figure}[t!]
    \centering
    \includegraphics[width=0.9\textwidth, clip, trim={0pt 0pt 0pt 5.5em}]{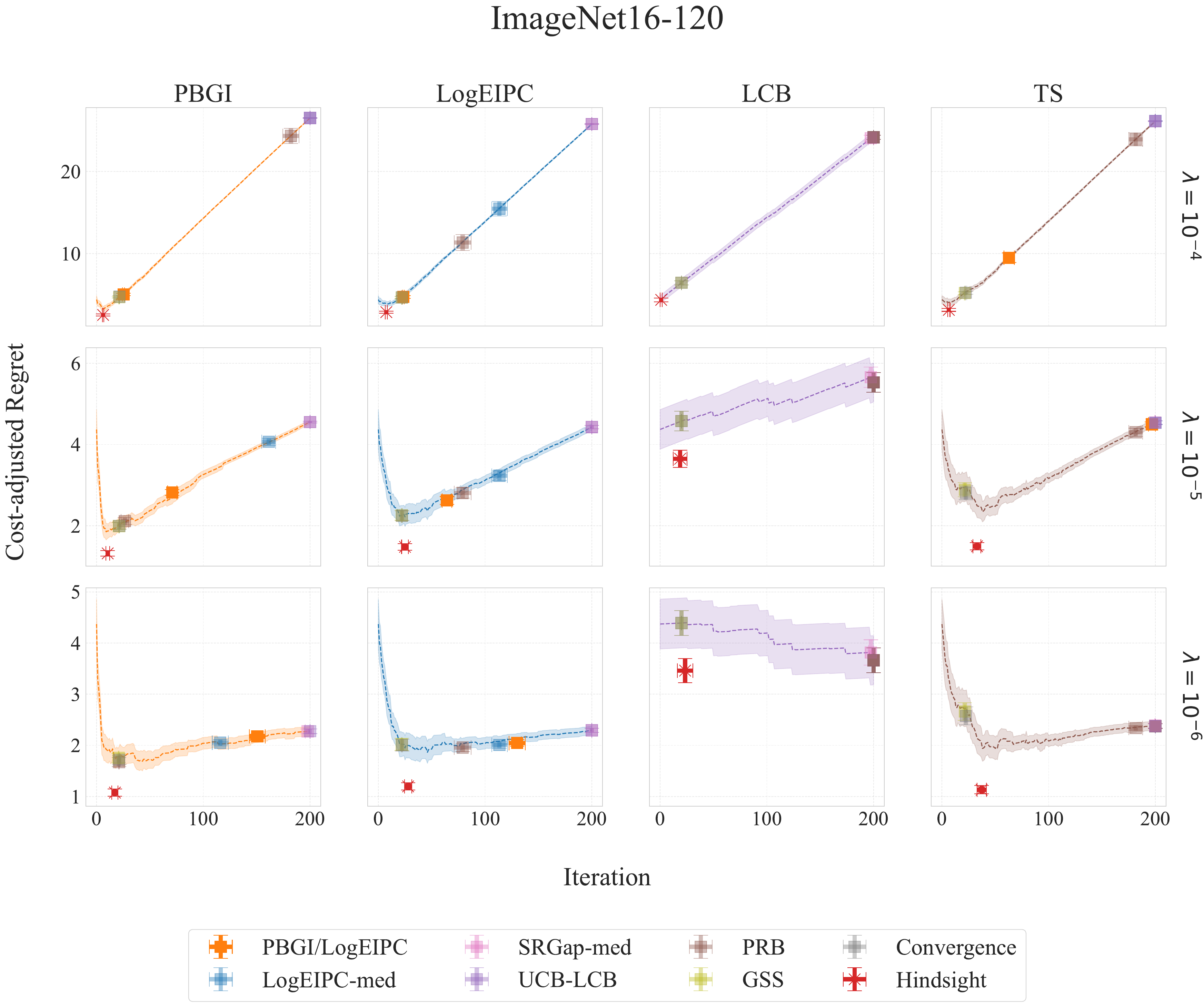}
    \caption{Comparison of cost-adjusted simple regret across acquisition function and stopping rule pairs on the \emph{ImageNet16-120} dataset. The objective function is the validation error, and the evaluation cost is the proxy runtime, scaled by $\lambda = 10^{-4}, 10^{-5}, 10^{-6}$. Each stopping rule is evaluated on the trajectory of its paired acquisition function. The red marker shows hindsight-optimal adaptive stopping, while the minimum of the fixed-iteration curve gives the hindsight-optimal fixed iteration. These references distinguish the effects of acquisition, stopping, and cost scaling. At $\lambda=10^{-5}$, GSS and Convergence have lower mean cost-adjusted regret than our rule when paired with either PBGI or LogEIPC.}
    \label{fig:ImageNet16-120}
\end{figure}

\Cref{fig:Fashion-MNIST,fig:adult,fig:higgs,fig:cifar10-valid,fig:cifar100,fig:ImageNet16-120} compare each stopping rule with the cost-adjusted regret trajectory of its corresponding acquisition function. The minimum of the mean fixed-iteration curve gives the hindsight-optimal fixed iteration, while the hindsight adaptive baseline selects a stopping iteration separately for each trial. The resulting comparisons depend on the dataset, acquisition function, and cost-scaling factor.

\subsubsection{Cost model misspecification: proxy runtime vs actual runtime.}
\begin{figure}[t!]
    \centering
    \includegraphics[width=0.8\textwidth]{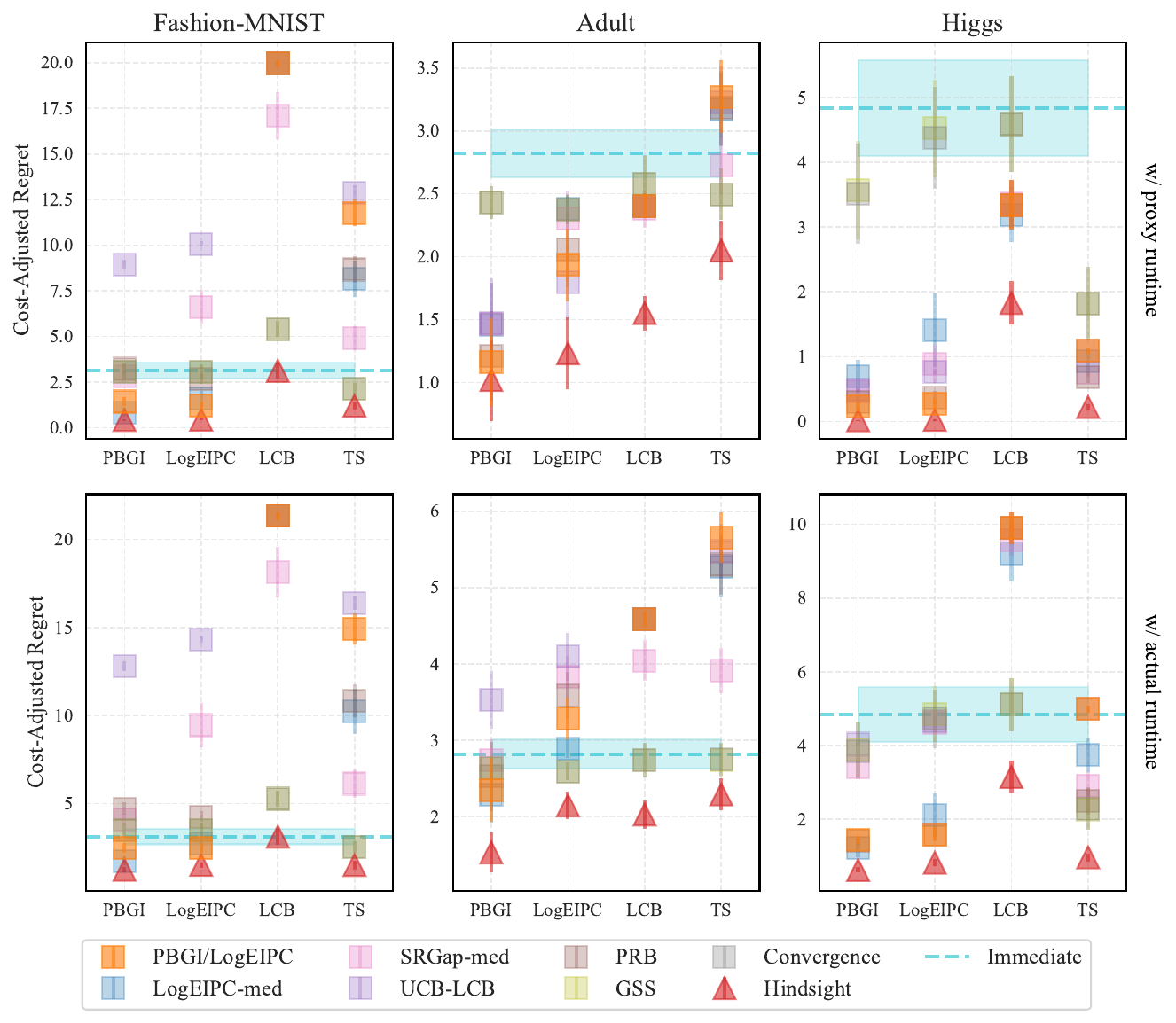}
    \caption{Comparison of cost-adjusted simple regret on three LCBench datasets with $\lambda = 10^{-4}$, using scaled proxy runtime versus scaled actual runtime for reporting evaluation cost. Online stopping decisions use proxy runtime in both cases; Hindsight is recomputed for each reporting cost. Changing the reporting cost can alter stopping-rule rankings, as illustrated by the PBGI and LogEIPC combinations on \emph{higgs}.}
    \label{fig:cost_misspecification_lcbench}
\end{figure}

In the known-cost setting of hyperparameter tuning, a practical approach is to use a proxy for runtime as the evaluation cost during the Bayesian optimization procedure. In our case, we use the number of model parameters scaled by a constant factor, which can be known in advance and has been shown to correlate well with the actual runtime. However, for reporting performance, one may prefer to use the actual runtime to better reflect real-world cost. To assess the impact of this cost model misspecification, we compare the cost-adjusted simple regret obtained when evaluation costs are computed using either the proxy runtime or the actual runtime. Figure~\ref{fig:cost_misspecification_lcbench} shows that changing the cost used for reporting can alter the relative ranking of stopping rules. On \emph{higgs}, our rule changes from first to third among the seven non-trivial rules paired with PBGI, and from first to second when paired with LogEIPC. Thus, the cost model used to report regret affects empirical comparisons even when the online stopping decisions are unchanged.

\begin{figure}[t!]
    \centering
    \includegraphics[width=0.8\textwidth]{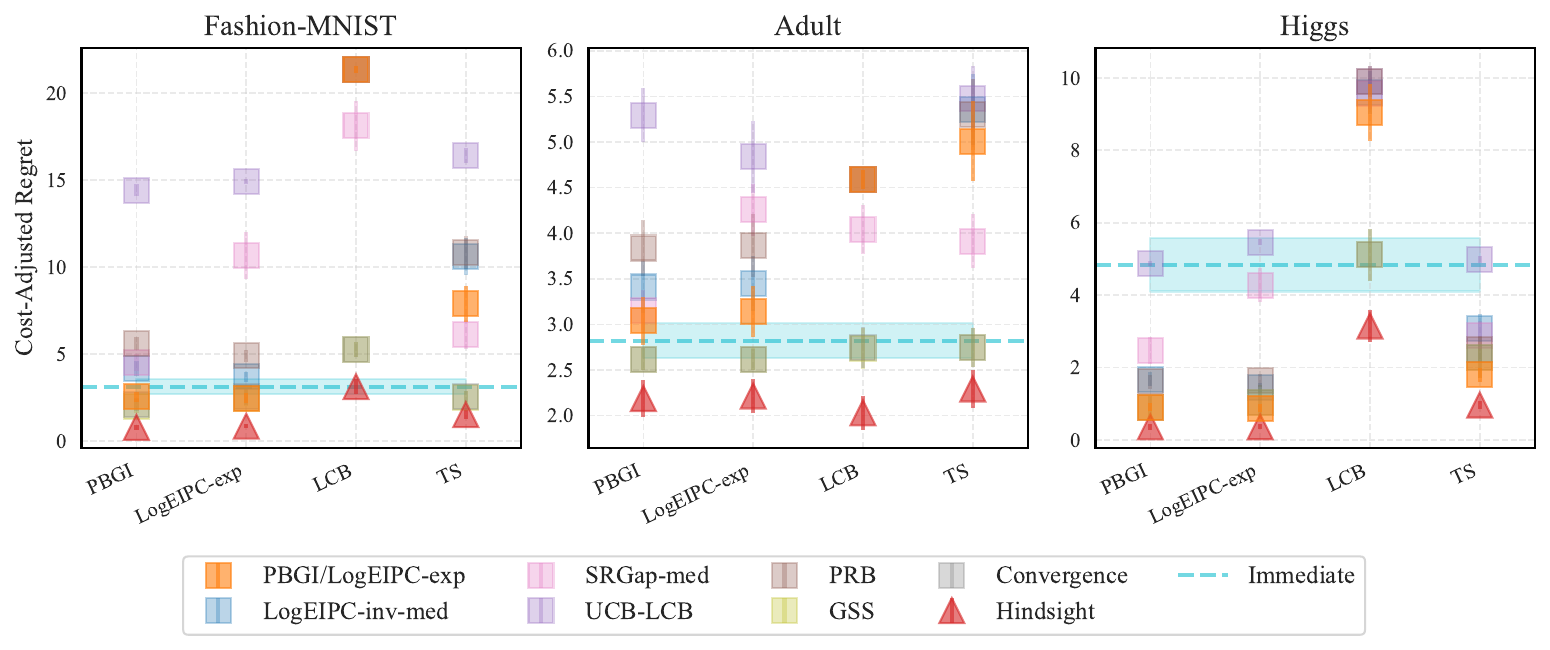}
    \caption{Cost-adjusted simple regret across acquisition function and stopping rule pairs under the unknown-cost setting on LCBench with $\lambda=10^{-4}$. The acquisition functions are PBGI, LogEIPC-exp, LCB, and TS. The primary PBGI/LogEIPC-exp stopping criterion uses the expected-cost diagnostic for every acquisition, while LogEIPC-inv-med denotes the median baseline using the inverse-cost diagnostic. The primary rule does not uniformly improve on simple baselines; on \emph{adult}, its PBGI and LogEIPC-exp combinations have higher mean regret than Immediate and GSS.}
    \label{fig:unknown_cost}
\end{figure}

\paragraph{Unknown-cost.}
\citet[Proposition~2]{astudillo2021multi} proposed modeling unknown cost $c(x)$ via
\begin{equation}
\label{eqn:-lnE[1/c]}
    \mathbb{E}[1/c(x)]^{-1}= \exp(\mu_{\ln c}(x)-(\sigma_{\ln c}(x))^2 / 2).
\end{equation}
An alternative is
\begin{equation}
\label{eqn:lnEc}
\mathbb{E}[c(x)]= \exp(\mu_{\ln c}(x)+(\sigma_{\ln c}(x))^2 / 2 ).
\end{equation}
The difference in sign before the variance term reflects how each formulation handles predictive uncertainty: \eqref{eqn:-lnE[1/c]} encourages more exploration than \eqref{eqn:lnEc}. For PBGI under the unknown-cost setting, it is more natural to replace $c(x)$ in \eqref{eqn:PBGI} with $\mathbb{E}[c(x)]$ using \eqref{eqn:lnEc}, as this aligns with how costs enter the root-finding problem. For LogEIPC, both variants are possible—we refer to the \eqref{eqn:-lnE[1/c]} version as \emph{LogEIPC-inv} and the \eqref{eqn:lnEc} version as \emph{LogEIPC-exp}. However, equivalence between PBGI and LogEIPC stopping rules and our theoretical guarantees hold only with \eqref{eqn:lnEc} but not with \eqref{eqn:-lnE[1/c]}. Accordingly, our primary stopping criterion uses the PBGI diagnostic based on \eqref{eqn:lnEc}, equivalently the LogEIPC-exp criterion, on the trajectory of every paired acquisition function. In \Cref{fig:unknown_cost}, we compare the PBGI, LogEIPC-exp, LCB, and TS acquisition functions, using the expected-cost LogEIPC-exp variant consistently with this criterion. The LogEIPC median baseline uses the inverse-cost diagnostic and is labeled LogEIPC-inv-med in this figure. The theoretical equivalence applies to the expected-cost criterion and its matching acquisition functions. Empirical performance depends on the dataset and acquisition: for example, on \emph{adult}, the primary rule has higher mean cost-adjusted regret than Immediate and GSS when paired with PBGI or LogEIPC-exp.

\subsubsection{Cost-adjusted simple regret of all LCBench datasets.} Due to space constraints, for LCBench, \Cref{fig:empirical} in the main text reports min–max normalized cost-adjusted simple regret, defined as 
\[\frac{r-r_{\min}}{r_{\max}-r_{\min}},\]
For each dataset and random seed separately, $r$ denotes the cost-adjusted regret of a given acquisition function--stopping rule pair, and $r_{\min}$ and $r_{\max}$ are the minimum (including Hindsight) and maximum regrets across all pairs on that same dataset and seed. We normalize each trial in this way before averaging across seeds and datasets. This normalization enables aggregation across datasets with different regret scales.

\begin{figure}[t!]
    \centering
    \includegraphics[width=\textwidth]{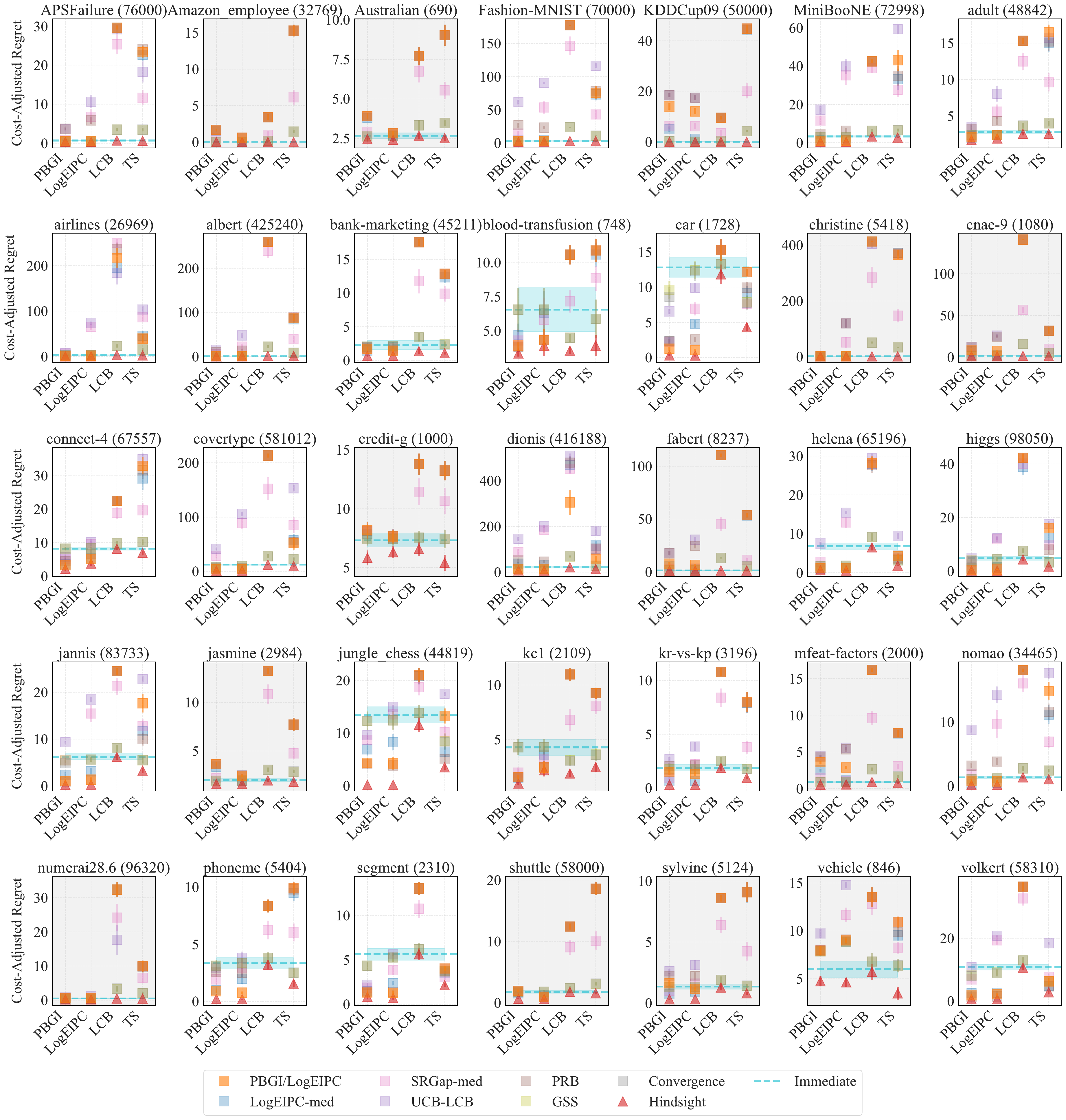}
    \caption{Cost-adjusted simple regret of all 35 LCBench datasets when $\lambda=10^{-3}$. Light grey highlights datasets where the matched PBGI pair does not rank among the top three of the seven non-trivial stopping rules paired with the PBGI acquisition function; Immediate and Hindsight are excluded from this ranking.}
    \label{fig:lcbench_all_1e-3}
\end{figure}

\begin{figure}[t!]
    \centering
    \includegraphics[width=\textwidth]{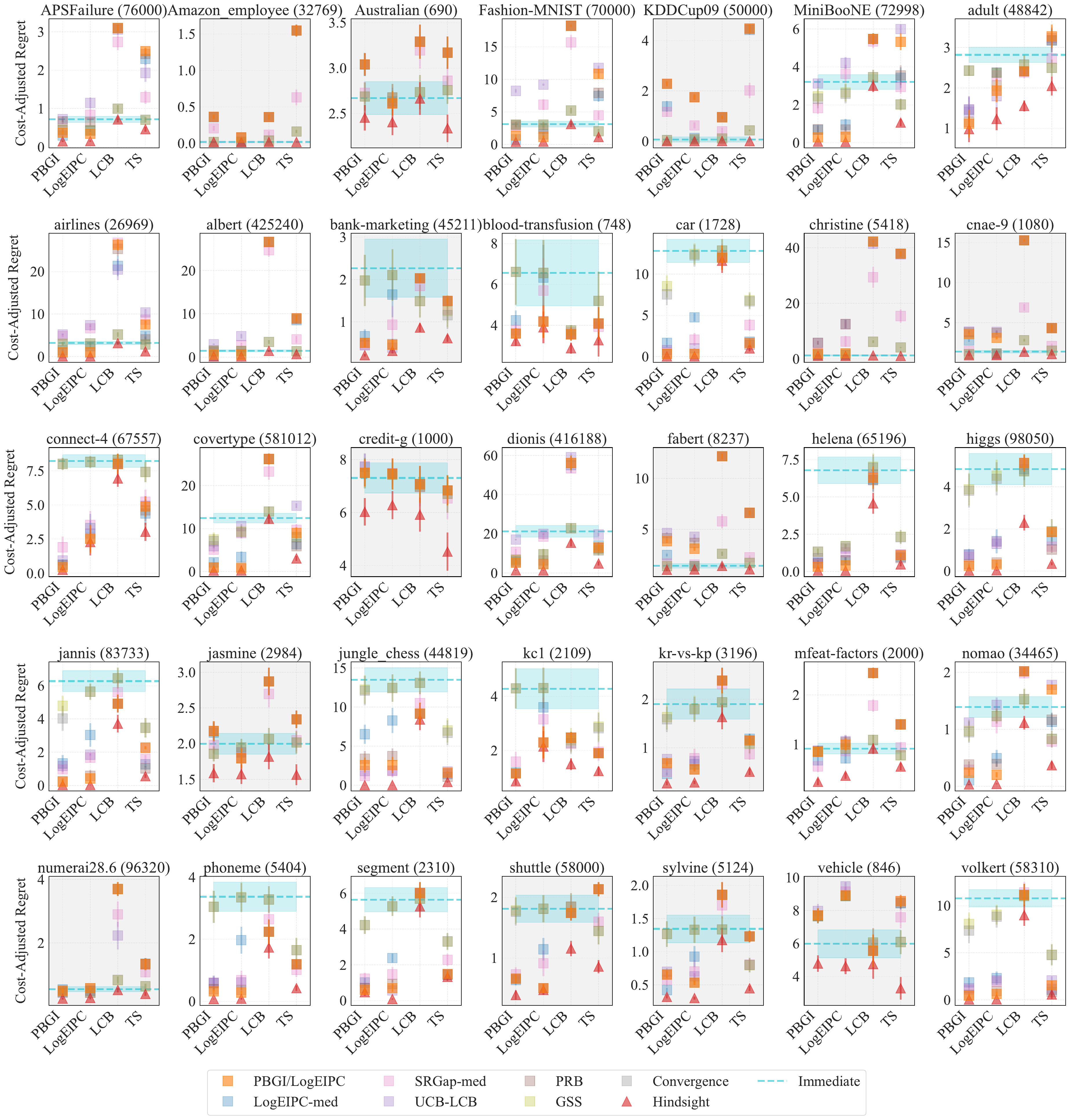}
    \caption{Cost-adjusted simple regret of all 35 LCBench datasets when $\lambda=10^{-4}$. Light grey highlights datasets where the matched PBGI pair does not rank among the top three of the seven non-trivial stopping rules paired with the PBGI acquisition function; Immediate and Hindsight are excluded from this ranking.}
    \label{fig:lcbench_all_1e-4}
\end{figure}

\begin{figure}[t!]
    \centering
    \includegraphics[width=\textwidth]{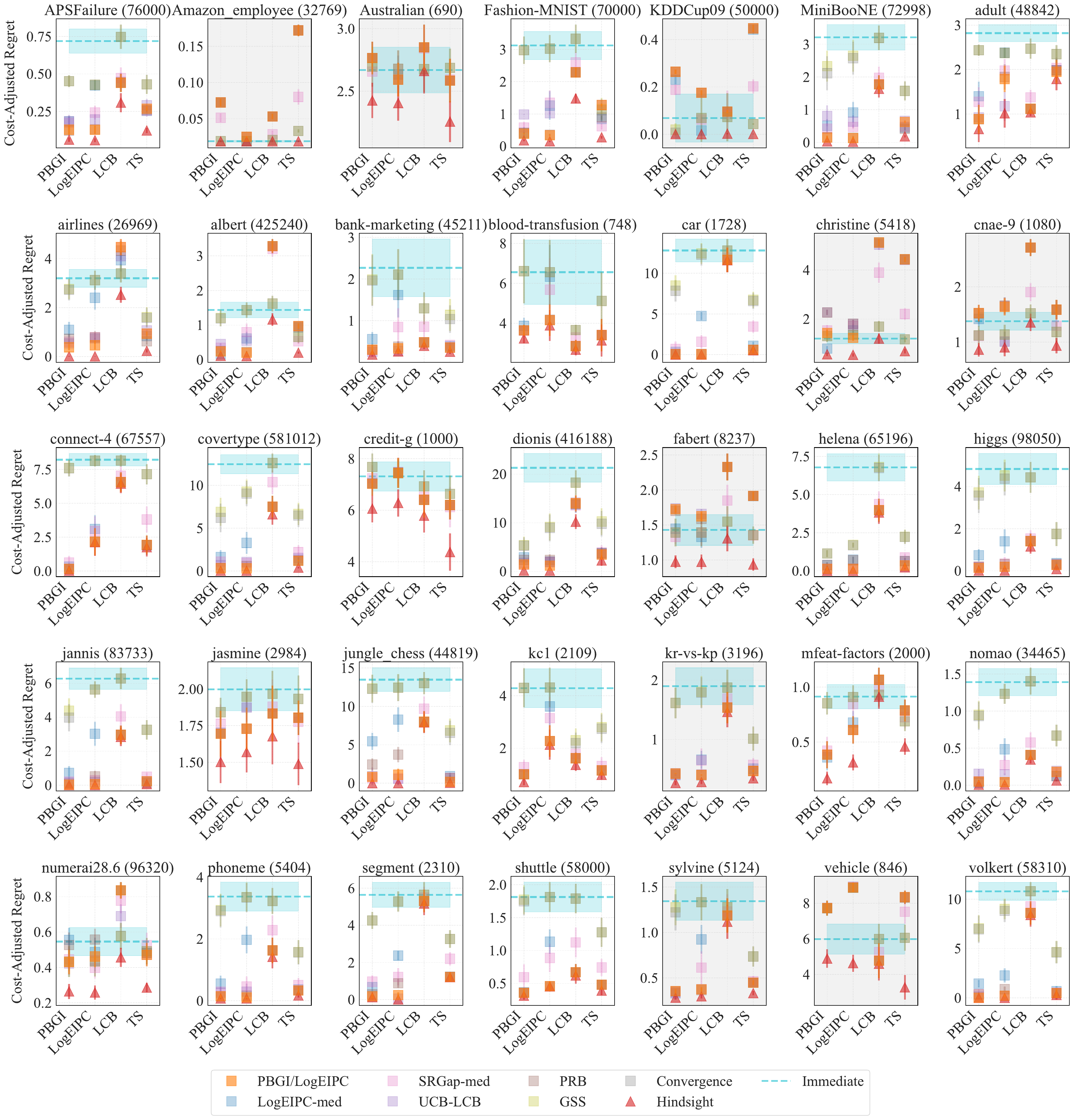}
    \caption{Cost-adjusted simple regret of all 35 LCBench datasets when $\lambda=10^{-5}$. Light grey highlights datasets where the matched PBGI pair does not rank among the top three of the seven non-trivial stopping rules paired with the PBGI acquisition function; Immediate and Hindsight are excluded from this ranking.}
    \label{fig:lcbench_all_1e-5}
\end{figure}

Here we present the \emph{unnormalized} bar-plot results for each LCBench dataset (OpenML dataset size in parentheses) under all three cost-scaling parameters in \Cref{fig:lcbench_all_1e-3,fig:lcbench_all_1e-4,fig:lcbench_all_1e-5}. Among the seven non-trivial stopping rules paired with the same acquisition function, our PBGI/LogEIPC rule ranks in the top three on 21, 21, and 26 of the 35 datasets with PBGI, and on 24, 26, and 28 datasets with LogEIPC, for $\lambda=10^{-3},10^{-4},10^{-5}$, respectively. The matched PBGI pair falls outside the top three at all three cost scales on eight datasets: Amazon\_employee\_access (32769), Australian (690), KDDCup09\_appetency (50000), christine (5418), cnae-9 (1080), fabert (8237), sylvine (5124), and vehicle (846). The other underperforming datasets vary with $\lambda$ and are highlighted in the corresponding plots.

Of these eight datasets, six have fewer than 10,000 instances; the exceptions are Amazon\_employee\_access and KDDCup09\_appetency. At individual cost scales, additional larger datasets can also fall outside the top three. This pattern motivates the validation--test mismatch diagnostic below, rather than establishing dataset size as the cause of underperformance.

\subsubsection{Diagnosing Underperforming LCBench Datasets via Validation--Test Rank Preservation.} \label{sec:val_test_diagnostic}
We examine whether weaker performance of our stopping rule on some LCBench datasets is associated with benchmark-intrinsic validation--test mismatch. We classify a dataset as ``underperforming'' if the cost-adjusted simple regret (at $\lambda = 10^{-4}$) of the matched PBGI pair does not rank among the top three PBGI-based acquisition function--stopping rule pairs formed by the seven non-trivial stopping rules we study. Formally, let
$\mathcal{S} = \{\text{PBGI/LogEIPC}, \text{LogEIPC-med}, \text{SRGap-med}, \text{UCB--LCB}, \text{PRB}, \text{GSS}, \text{Convergence}\}$
(excluding the Immediate and Hindsight baselines), and let
$r_d(s) := \mathcal{R}_c^{(\lambda=10^{-4})}(\text{PBGI}, s; d)$
denote the mean cost-adjusted regret of PBGI paired with rule $s$ on dataset $d$. With
$\rho_d := \big|\{s \in \mathcal{S} : r_d(s) \le r_d(\text{PBGI/LogEIPC})\}\big|$
the within-dataset rank of the matched rule (rank $1$ best), we set
\begin{equation}
    d \in
    \begin{cases}
        \mathcal{D}_{\mathrm{top}} & \text{if } \rho_d \le 3, \\
        \mathcal{D}_{\mathrm{under}} & \text{if } \rho_d \ge 4.
    \end{cases}
\end{equation}

The LCBench experiments optimize validation error but report test error. We quantify how faithfully validation ranks transfer to test ranks via the top-$k$ overlap fraction. For dataset $d$ with $n_d$ configurations, let $\pi^{\mathrm{val}}_d, \pi^{\mathrm{test}}_d \in S_{n_d}$ be the rankings of configurations by validation and test error (ascending, best first); then
\begin{equation}
    \mathrm{Top}\text{-}k(d) := \frac{1}{k}\,\big|\{\pi^{\mathrm{val}}_d(1), \dots, \pi^{\mathrm{val}}_d(k)\} \cap \{\pi^{\mathrm{test}}_d(1), \dots, \pi^{\mathrm{test}}_d(k)\}\big|.
\end{equation}
We take $k = 20$, comparable to the $200$-iteration cap in our experiments and to the top tail of configurations any adaptive method realistically considers as stopping candidates.

\begin{figure}[ht]
    \centering
    \includegraphics[width=\linewidth]{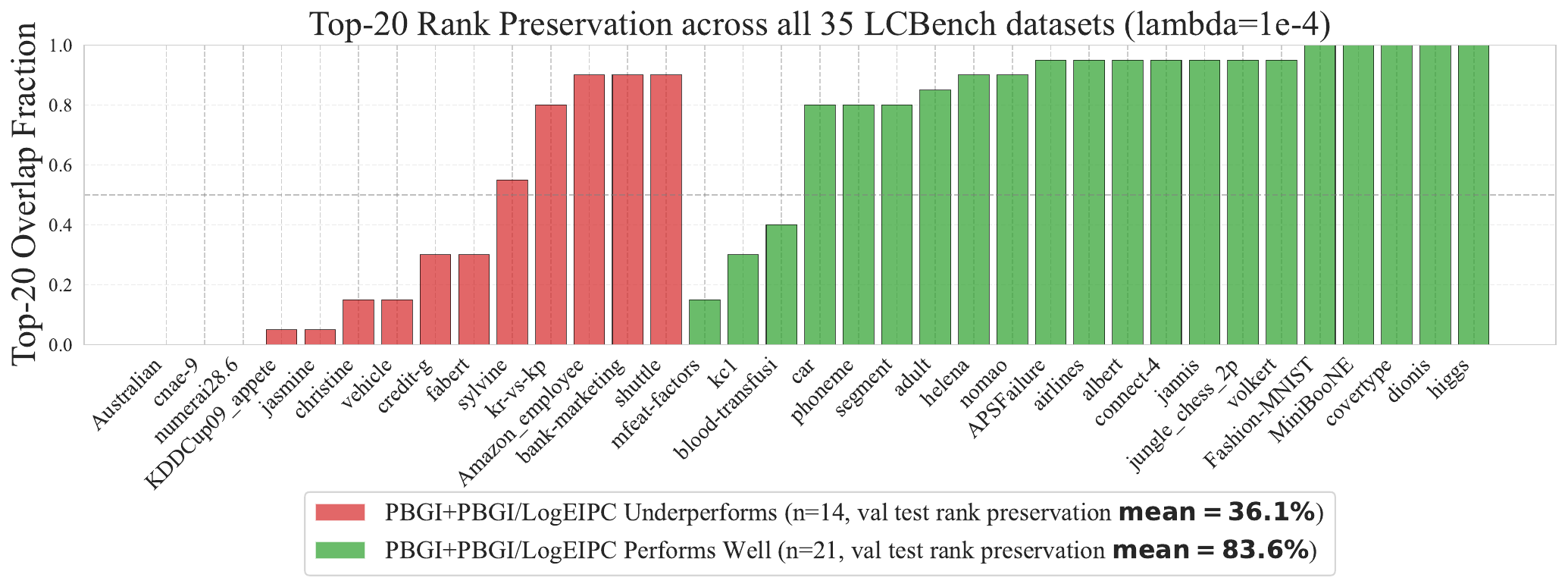}
    \caption{Top-20 validation--test rank preservation for every LCBench
    dataset. Datasets are split into $\mathcal{D}_{\mathrm{under}}$ (red) and
    $\mathcal{D}_{\mathrm{top}}$ (green) based on whether the matched PBGI pair
    ranks among the top three PBGI-based acquisition function--stopping rule
    pairs formed by the seven stopping rules in $\mathcal{S}$ at
    $\lambda = 10^{-4}$. The dashed line marks $\mathrm{Top}\text{-}20 = 0.5$.
    Datasets where the matched PBGI pair underperforms also have lower
    validation--test rank preservation. This association is consistent with
    benchmark mismatch contributing to the observed cost-adjusted regret,
    without establishing a causal explanation.}
    \label{fig:val_test_all_lcbench}
\end{figure}

Figure~\ref{fig:val_test_all_lcbench} reports $\mathrm{Top}\text{-}20(d)$ for every LCBench dataset, color-coded by the above classification. The $\mathcal{D}_{\mathrm{under}}$ datasets have \textit{substantially} lower val--test rank preservation than those in $\mathcal{D}_{\mathrm{top}}$:
\begin{equation*}
    \overline{\mathrm{Top}\text{-}20}\,\big|_{\mathcal{D}_{\mathrm{under}}}
    = \mathbf{36.1\%}
    \;(n = 14,\ \mathrm{std} = 35.7\%)
    \quad\ll\quad
    \overline{\mathrm{Top}\text{-}20}\,\big|_{\mathcal{D}_{\mathrm{top}}}
    = \mathbf{83.6\%}
    \;(n = 21,\ \mathrm{std} = 23.8\%).
\end{equation*}
Low overlap between validation-selected and test-optimal configurations can limit the test performance of a validation-driven optimization policy. The observed association is consistent with validation--test mismatch contributing to the matched PBGI pair's weaker performance, but does not by itself isolate the cause of underperformance.

\subsection{Additional Experiment Results: Bayesian Regret}

\begin{figure}[t!]
    \centering
    \includegraphics[width=0.9\textwidth]{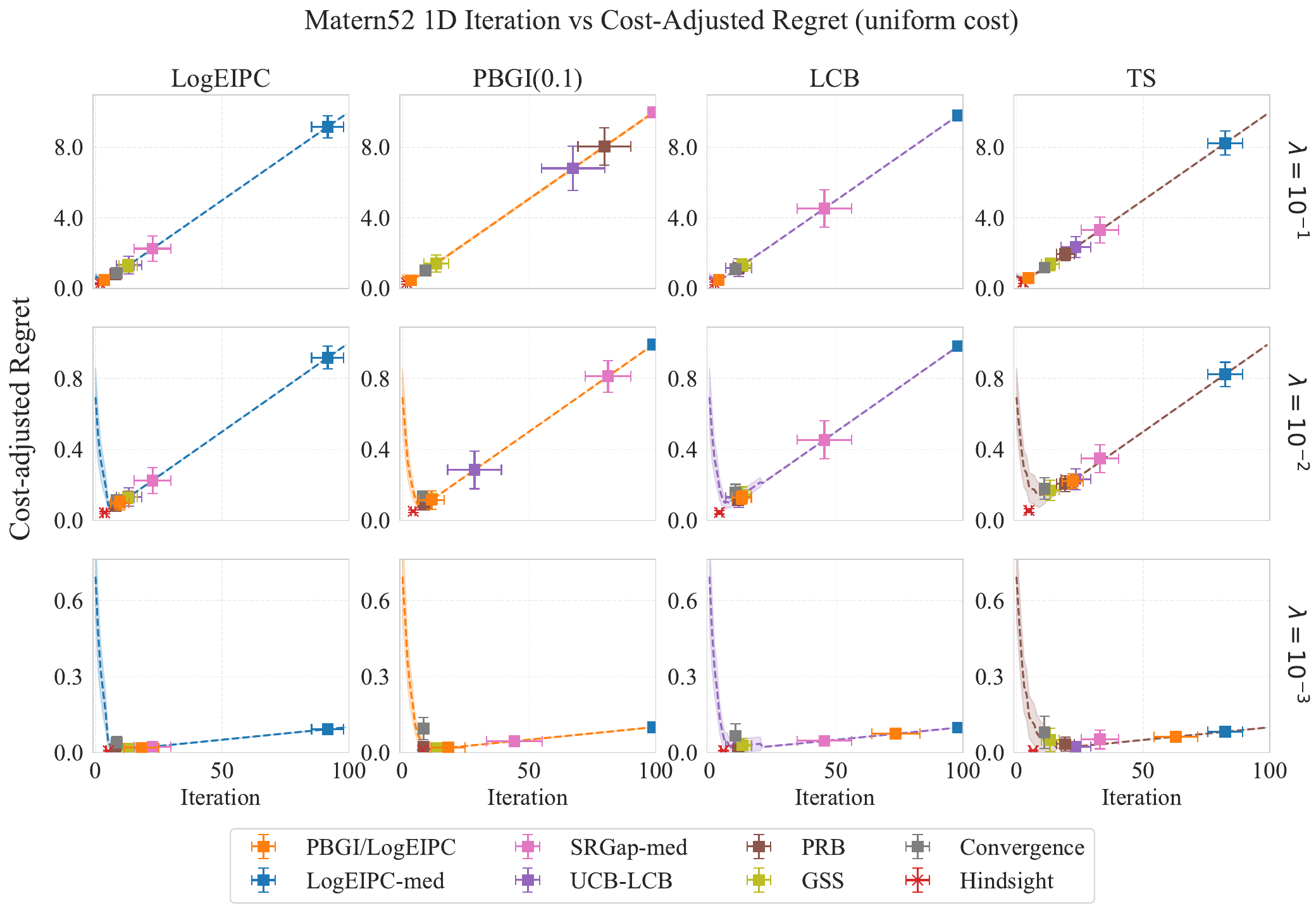}
    \caption{Comparison of cost-adjusted simple regret across acquisition function and stopping rule pairs in the 1D Bayesian regret experiments, with cost‐scaling factor $\lambda=10^{-1}, 10^{-2}, 10^{-3}$ and under uniform cost. The dashed line in each subplot represent fixed iteration stopping (e.g., the y-axis value of the line at iteration $50$ represent the cost-adjusted regret when always stop at iteration $50$).}
    \label{fig:Matern52_1D_uniform}
\end{figure}

\begin{figure}[t!]
    \centering
    \includegraphics[width=0.9\textwidth]{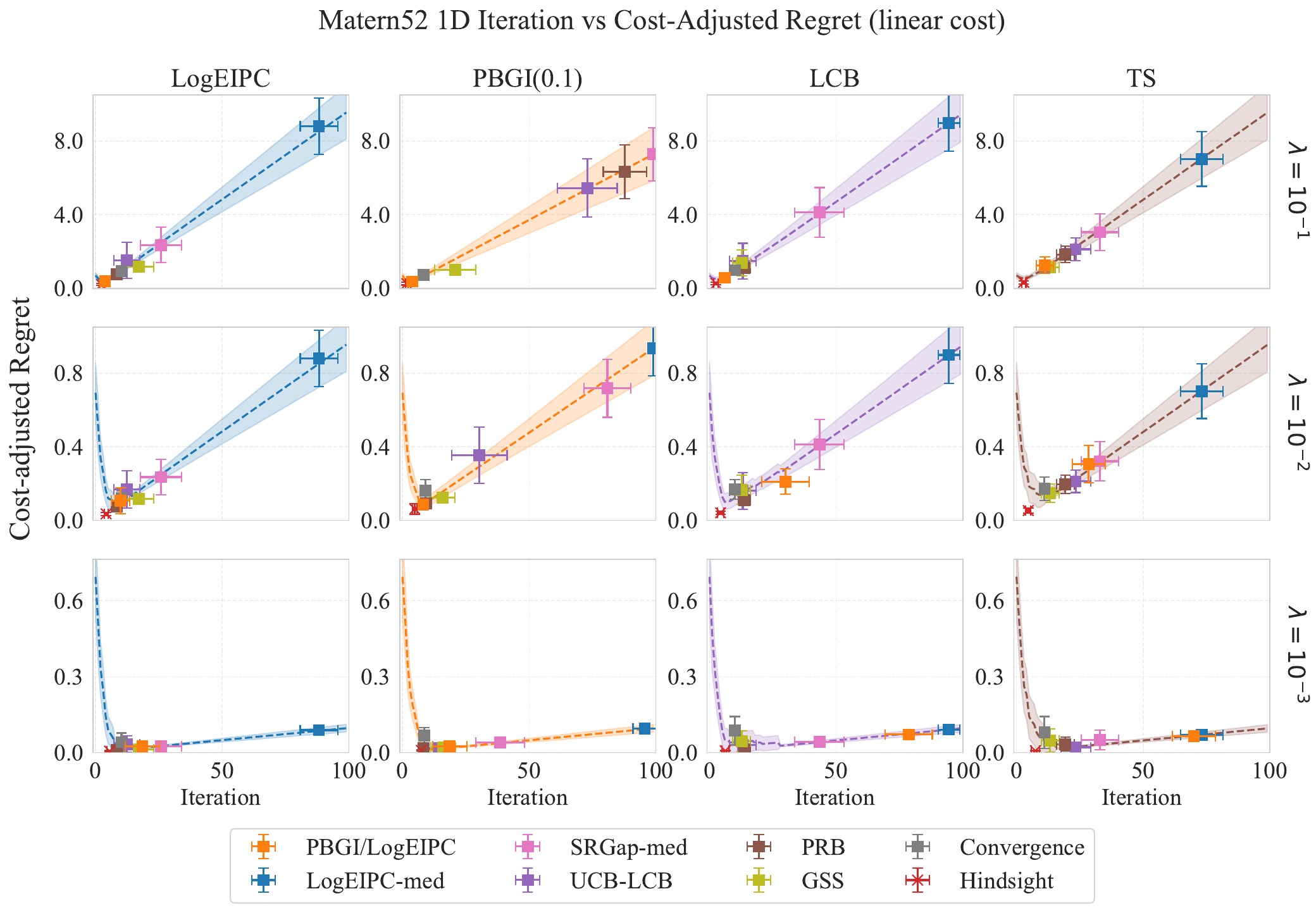}
    \caption{Comparison of cost-adjusted simple regret across acquisition function and stopping rule pairs in the 1D Bayesian regret experiments, with cost‐scaling factor $\lambda=10^{-1}, 10^{-2}, 10^{-3}$ and under linear cost. The dashed line in each subplot represent fixed iteration stopping (e.g., the y-axis value of the line at iteration $50$ represent the cost-adjusted regret when always stop at iteration $50$).}
    \label{fig:Mater52_1D_linear}
\end{figure}

\begin{figure}[t!]
    \centering
    \includegraphics[width=0.9\textwidth]{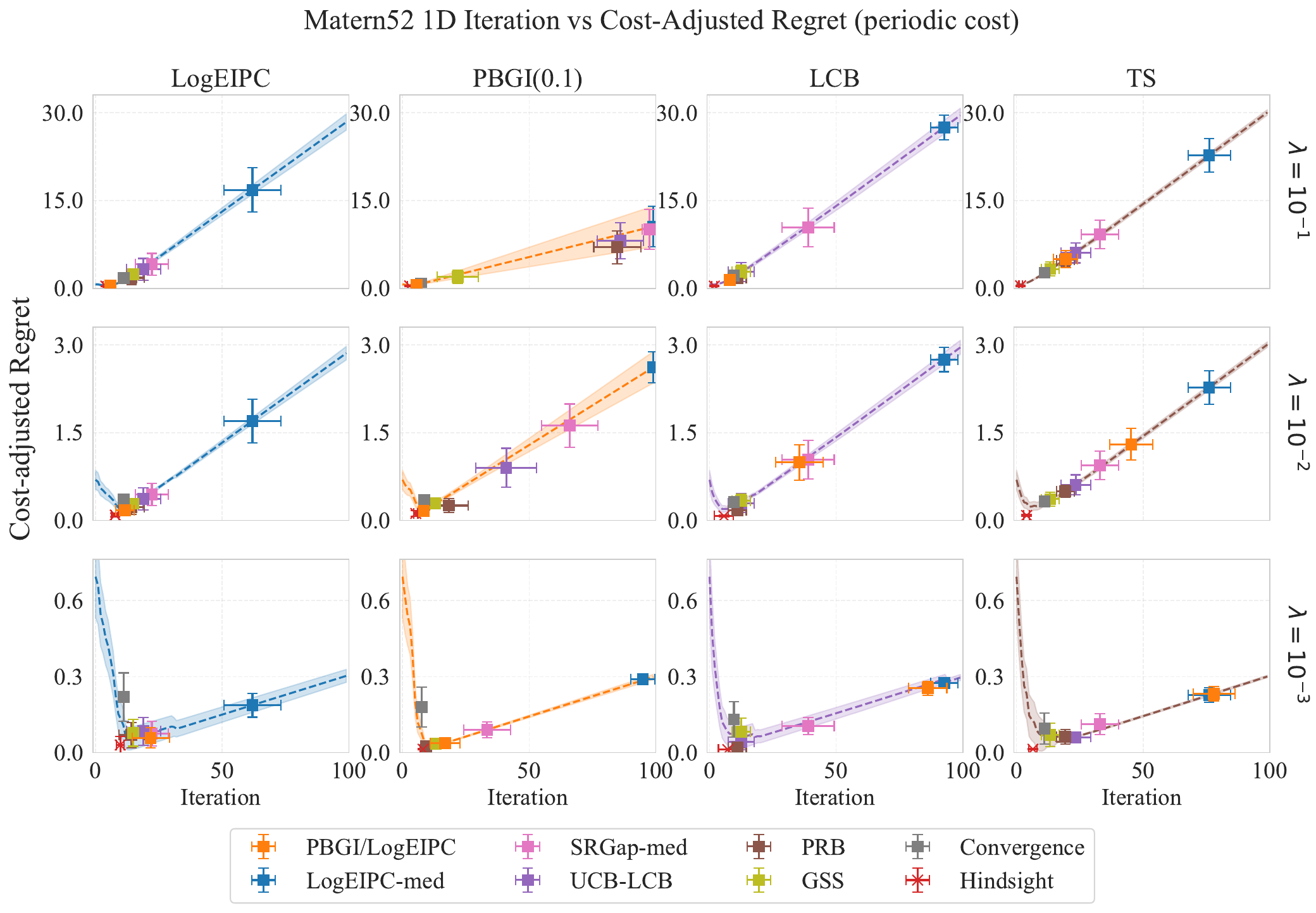}
    \caption{Comparison of cost-adjusted simple regret across acquisition function and stopping rule pairs in the 1D Bayesian regret experiments, with cost‐scaling factor $\lambda=10^{-1}, 10^{-2}, 10^{-3}$ and under periodic cost. The dashed line in each subplot represent fixed iteration stopping (e.g., the y-axis value of the line at iteration $50$ represent the cost-adjusted regret when always stop at iteration $50$).}
    \label{fig:Mater52_1D_periodic}
\end{figure}

\begin{figure}[t!]
    \centering
    \includegraphics[width=0.9\textwidth]{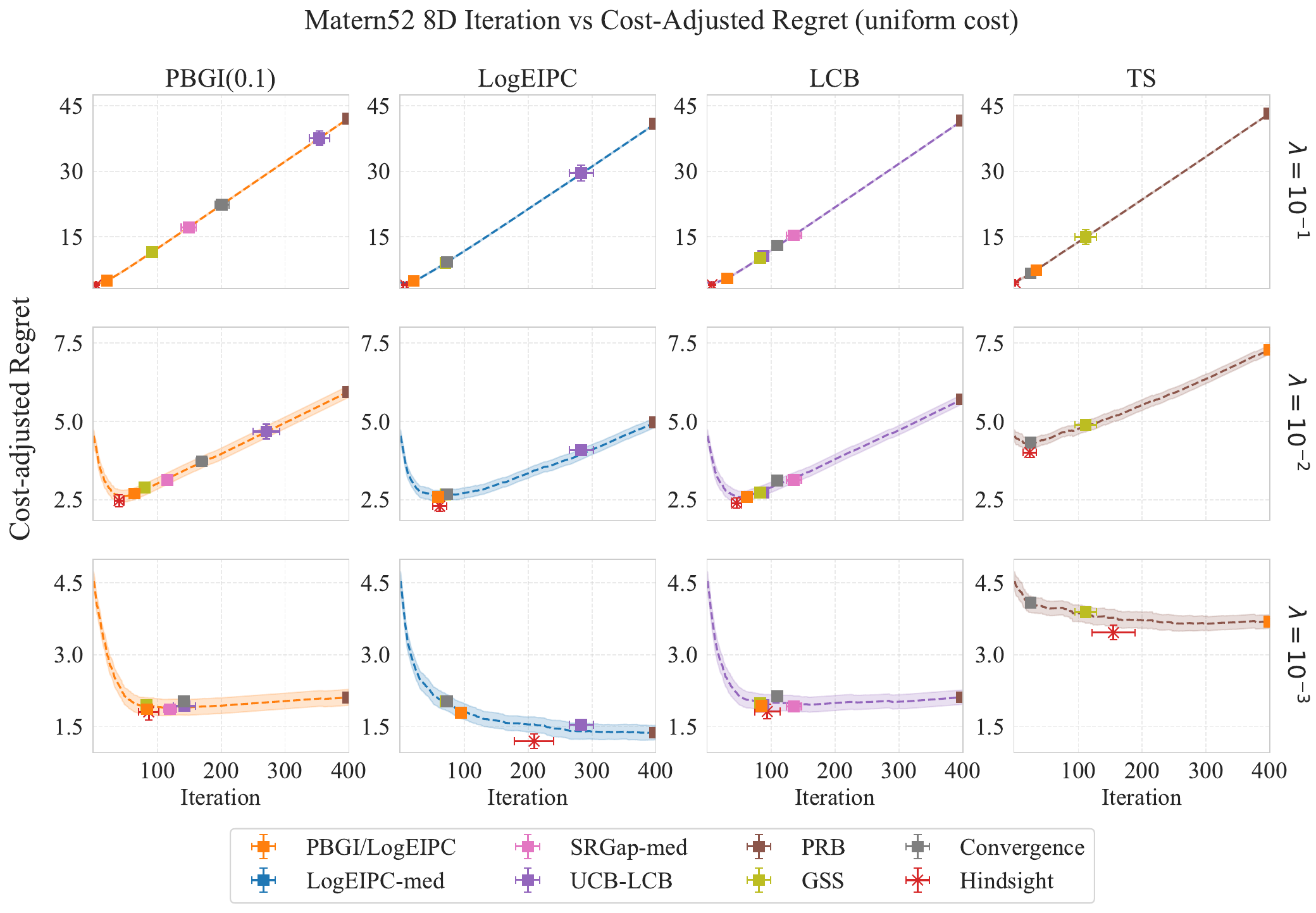}
    \caption{Comparison of cost-adjusted simple regret across acquisition function and stopping rule pairs in the 8D Bayesian regret experiments, with cost‐scaling factor $\lambda=10^{-1}, 10^{-2}, 10^{-3}$ and under uniform cost. The dashed line in each subplot represent fixed iteration stopping (e.g., the y-axis value of the line at iteration $50$ represent the cost-adjusted regret when always stop at iteration $50$).}
    \label{fig:Mater52_8D_uniform}
\end{figure}

\begin{figure}[t!]
    \centering
    \includegraphics[width=0.9\textwidth]{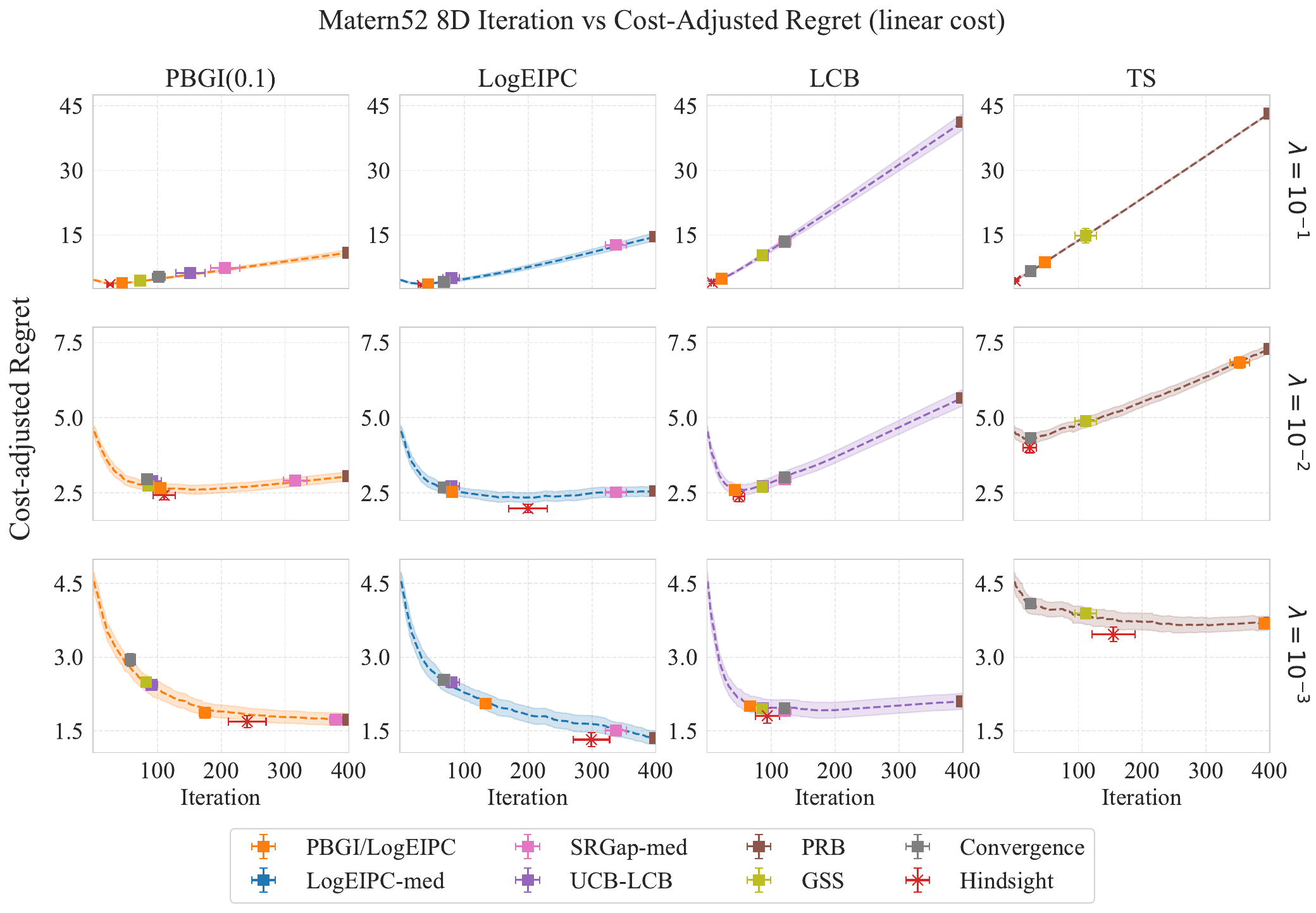}
    \caption{Comparison of cost-adjusted simple regret across acquisition function and stopping rule pairs in the 8D Bayesian regret experiments, with cost‐scaling factor $\lambda=10^{-1}, 10^{-2}, 10^{-3}$ and under linear cost. The dashed line in each subplot represent fixed iteration stopping (e.g., the y-axis value of the line at iteration $50$ represent the cost-adjusted regret when always stop at iteration $50$).}
    \label{fig:Mater52_8D_linear}
\end{figure}

\begin{figure}[t!]
    \centering
    \includegraphics[width=0.9\textwidth]{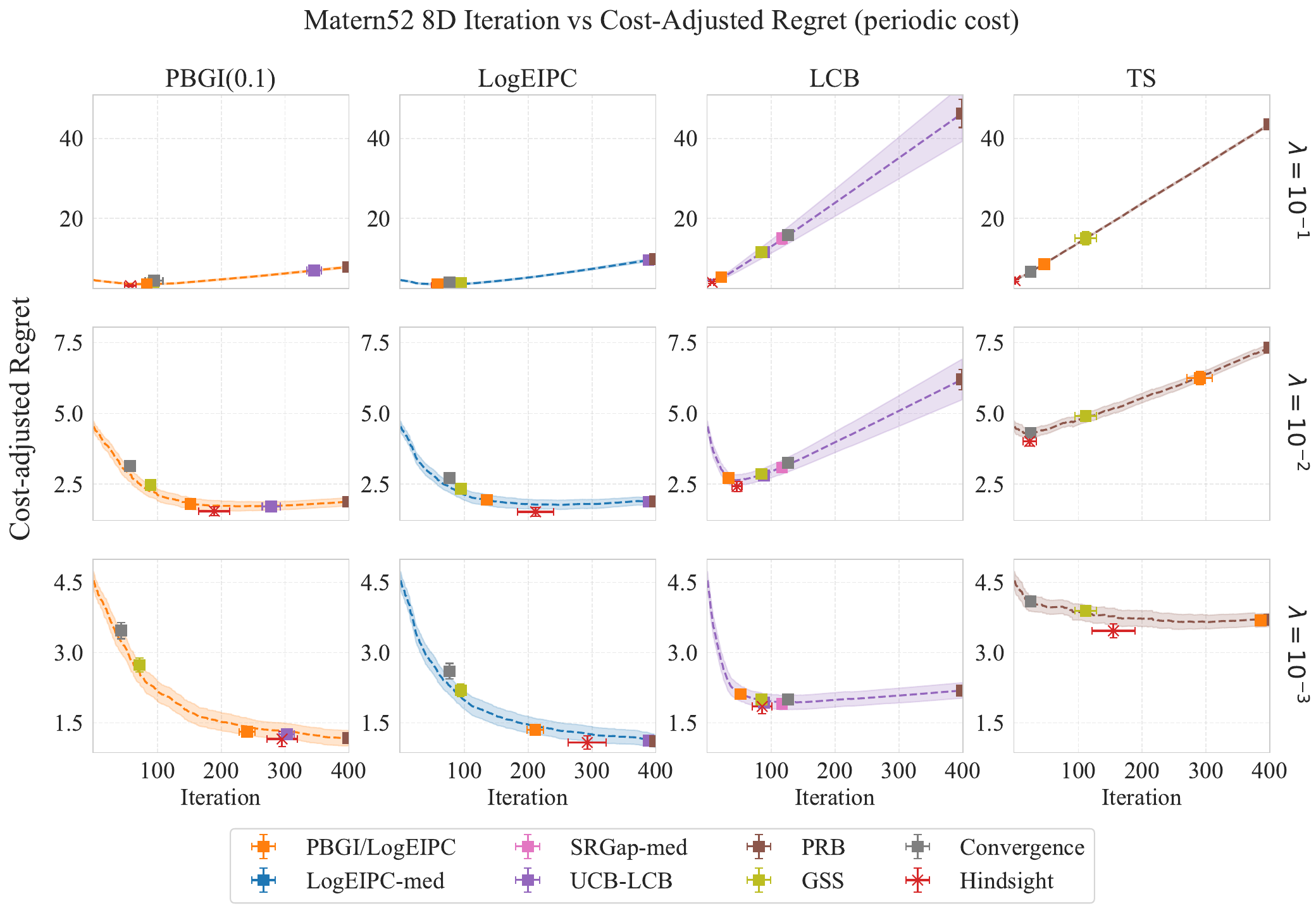}
    \caption{Comparison of cost-adjusted simple regret across acquisition function and stopping rule pairs in the 8D Bayesian regret experiments, with cost‐scaling factor $\lambda=10^{-1}, 10^{-2}, 10^{-3}$ and under periodic cost. The dashed line in each subplot represent fixed iteration stopping (e.g., the y-axis value of the line at iteration $50$ represent the cost-adjusted regret when always stop at iteration $50$).}
    \label{fig:Mater52_8D_periodic}
\end{figure}

In this section, we present the complete Bayesian regret results.  \Cref{fig:Matern52_1D_uniform,fig:Mater52_1D_linear,fig:Mater52_1D_periodic} show the 1D experiments, and \Cref{fig:Mater52_8D_uniform,fig:Mater52_8D_linear,fig:Mater52_8D_periodic} show the 8D experiments. Each figure corresponds to one cost setting (uniform, linear or periodic) and three values of the cost-scaling factor, $\lambda=10^{-1},10^{-2},10^{-3}$. In all of the experimental results, we observe that PBGI/LogEIPC acquisition function + PBGI/LogEIPC stopping achieves cost-adjusted regret that is not only competitive with the baselines, but is also competitive regarding the \textit{best in hindsight} fixed iteration stopping and often competitive even comparing to \textit{hindsight optimal} stopping. These results indicate that our automatic stopping rule can replace manual selection of stopping times without loss in performance. \Cref{fig:Matern52_1D_uniform,fig:Mater52_1D_linear,fig:Mater52_1D_periodic,fig:Mater52_8D_uniform,fig:Mater52_8D_linear,fig:Mater52_8D_periodic} also show that our PBGI/LogEIPC stopping rule outperforms other baselines when the cost-scaling factor $\lambda$ is large, indicating that it's an  especially suitable stopping criteria when evaluation is expensive. 


\subsubsection{Ablation: Moving-Average Window Size}
\label{sec:ma_ablation}

\begin{figure}[h]
    \centering
    \includegraphics[width=\linewidth]{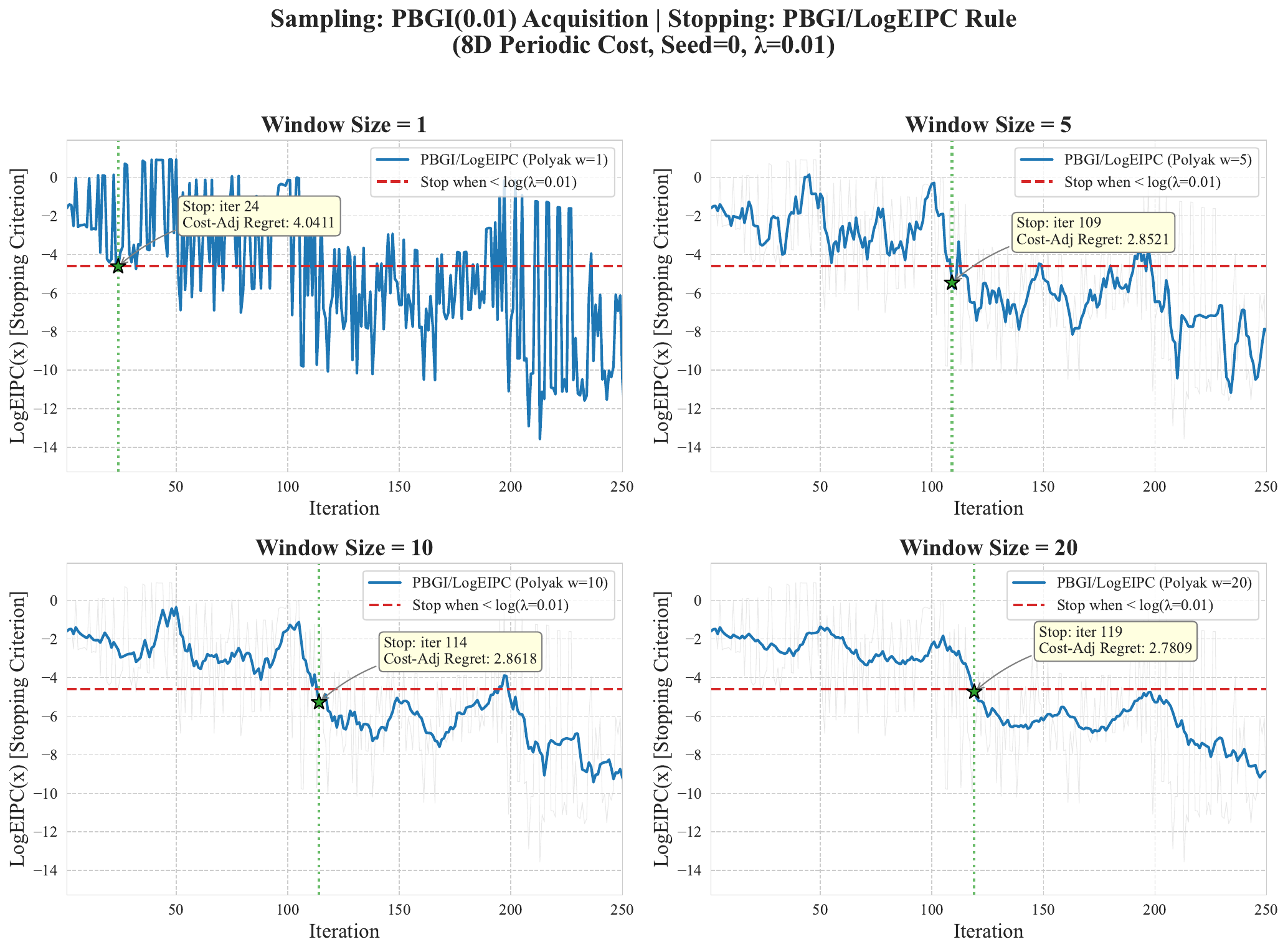}
    \caption{Single-trajectory illustration of how the moving-average
    window size $W$ affects the PBGI/LogEIPC stopping signal on the 8D
    Bayesian regret benchmark under periodic costs (seed 0,
    $\lambda = 10^{-2}$), with PBGI ($\lambda = 10^{-2}$) as the
    acquisition function. The light grey curve is the raw $\alpha_t^{\mathrm{LogEIPC}}$,
    the blue curve its $W$-iteration trailing average, the red dashed line
    the stopping threshold $\log\lambda$, and the green star the resulting
    stopping iteration. Without smoothing ($W=1$) the rule fires early
    (iter.~24, regret 4.04) on a noisy dip; smoothing defers the trigger
    and recovers low cost-adjusted regret (iter.~119, regret 2.78 at $W=20$).}
    \label{fig:ma_single_pbgi}
\end{figure}

\begin{figure}[h]
    \centering
    \includegraphics[width=\linewidth]{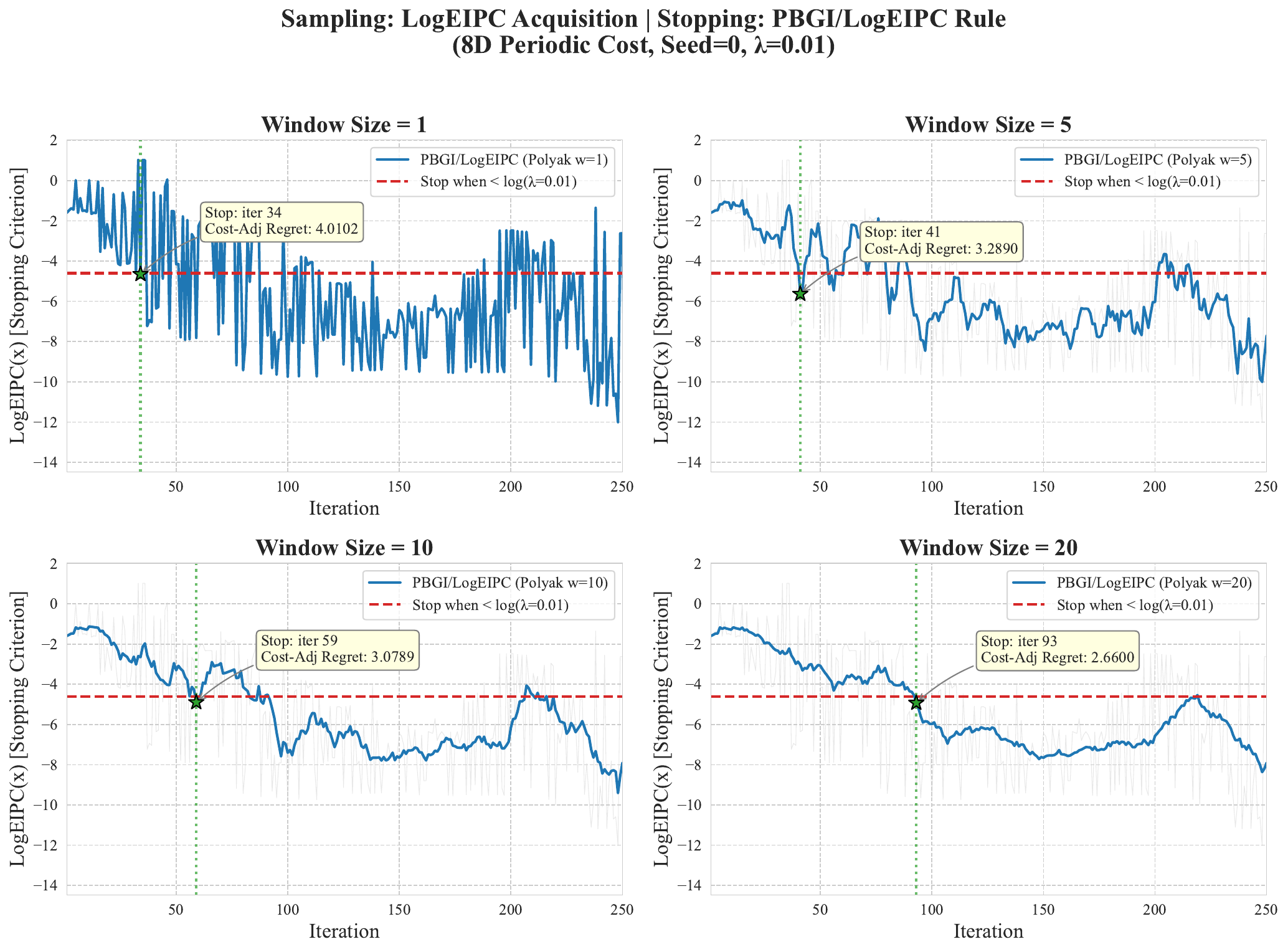}
    \caption{Single-trajectory illustration analogous to Figure~\ref{fig:ma_single_pbgi},
    with the LogEIPC acquisition function (same seed, same cost regime,
    $\lambda = 10^{-2}$). Without smoothing the rule fires at iter.~34 with
    regret $4.01$; with $W=20$ it fires at iter.~93 with regret $2.66$.}
    \label{fig:ma_single_logeipc}
\end{figure}

\begin{figure}[h]
    \centering
    \includegraphics[width=\linewidth]{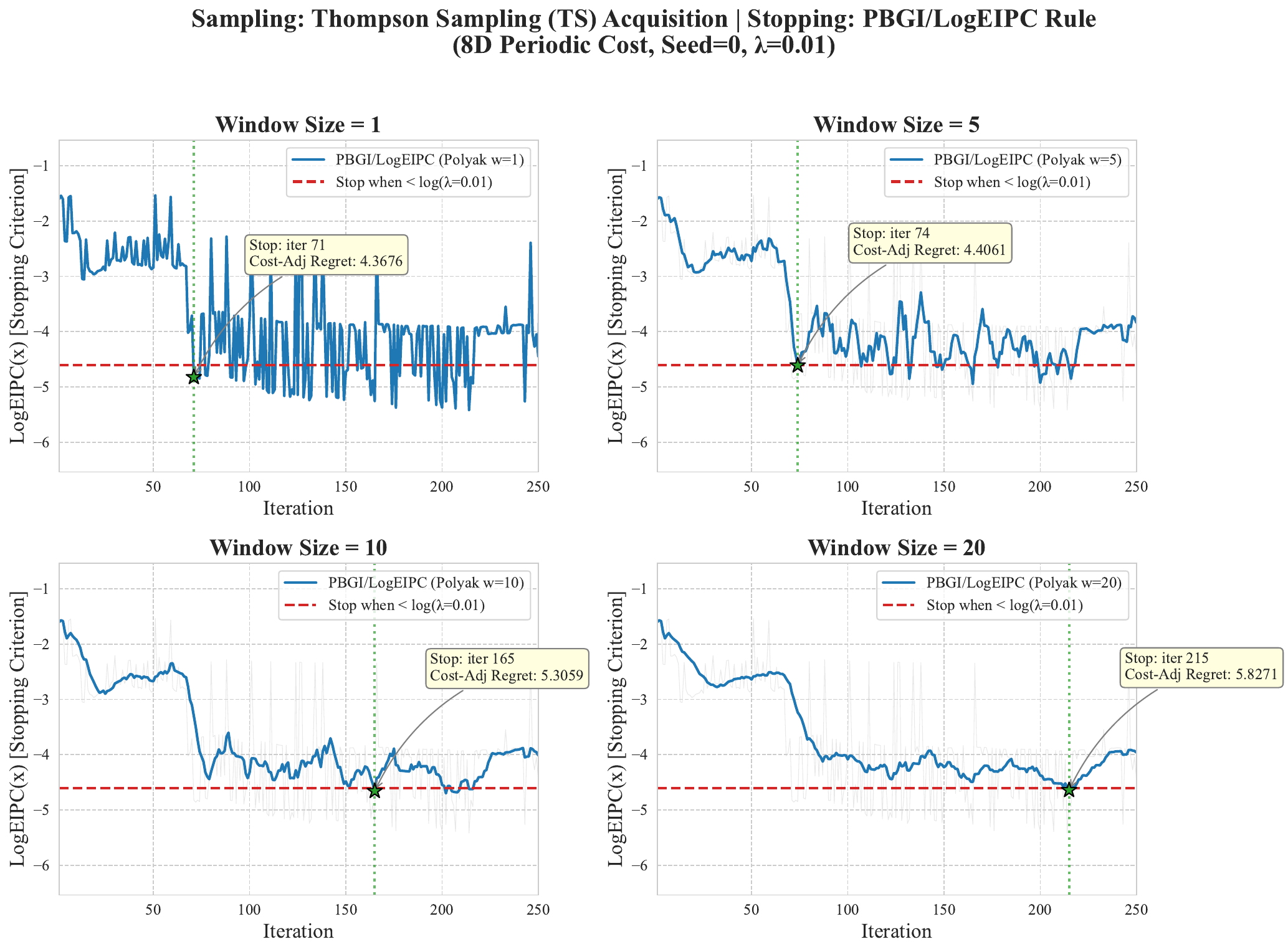}
    \caption{Single-trajectory illustration analogous to Figure~\ref{fig:ma_single_pbgi},
    with Thompson sampling (TS) as the acquisition function. Unlike the
    matched-acquisition cases, smoothing here pushes the stopping iteration
    later (iter.~71 at $W=1$ vs.~iter.~215 at $W=20$) without a corresponding
    drop in cost-adjusted regret -- TS does not satisfy the
    expected-improvement-exceeds-cost property of Lemma~\ref{lem:EI_lower_bound},
    so the PBGI/LogEIPC stopping rule has no matched guarantee for it.}
    \label{fig:ma_single_ts}
\end{figure}

\begin{figure}[h]
    \centering
    \includegraphics[width=\linewidth]{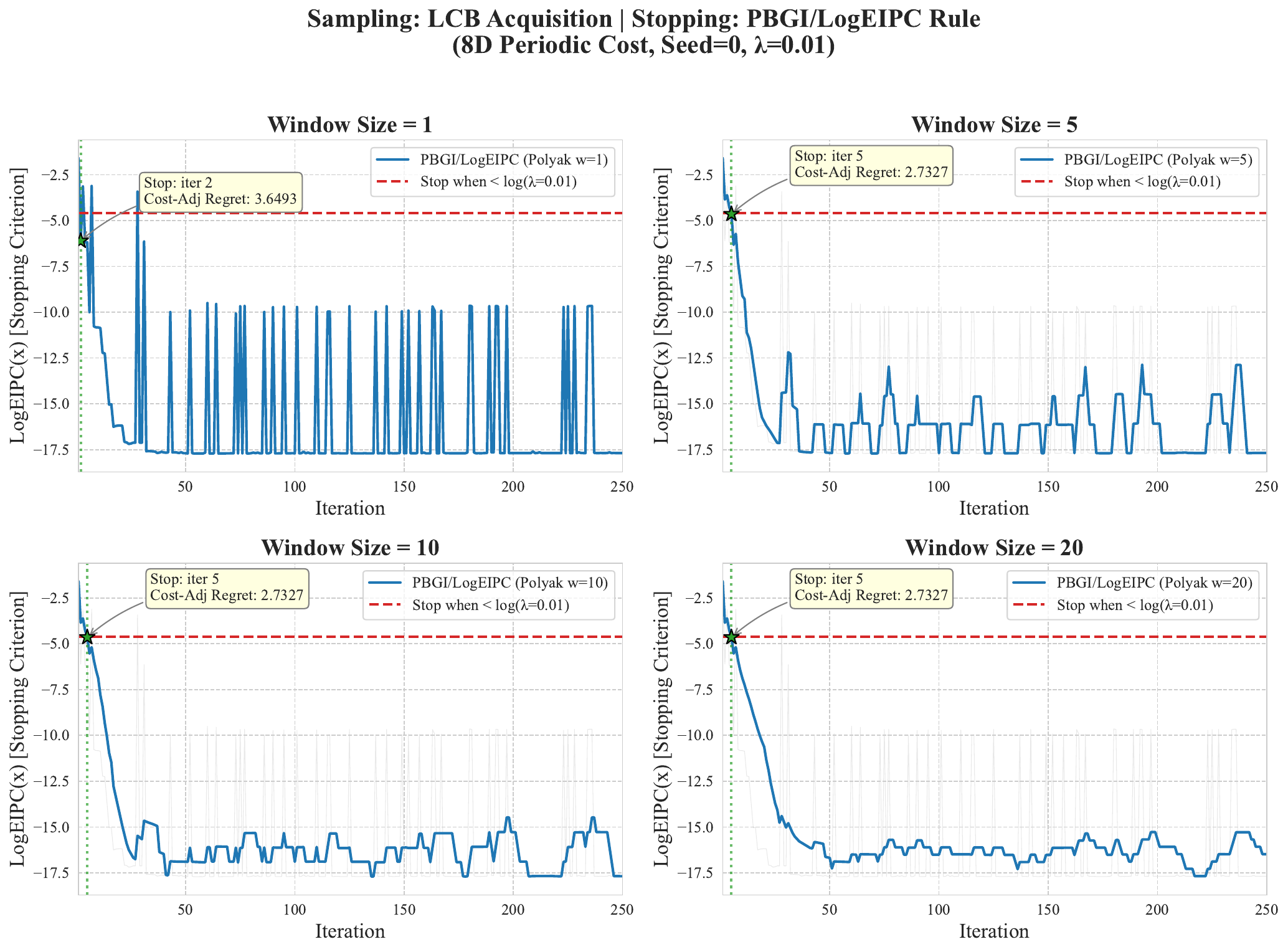}
    \caption{Single-trajectory illustration analogous to Figure~\ref{fig:ma_single_pbgi},
    with LCB as the acquisition function. The raw LogEIPC signal has a
    clean transient: smoothing changes the stopping iteration only slightly
    (iter.~2 at $W=1$ vs.~iter.~5 at $W\ge 5$) and the cost-adjusted regret
    stabilises at $2.73$ for all $W \ge 5$.}
    \label{fig:ma_single_lcb}
\end{figure}

\begin{figure}[h]
    \centering
    \includegraphics[width=\linewidth]{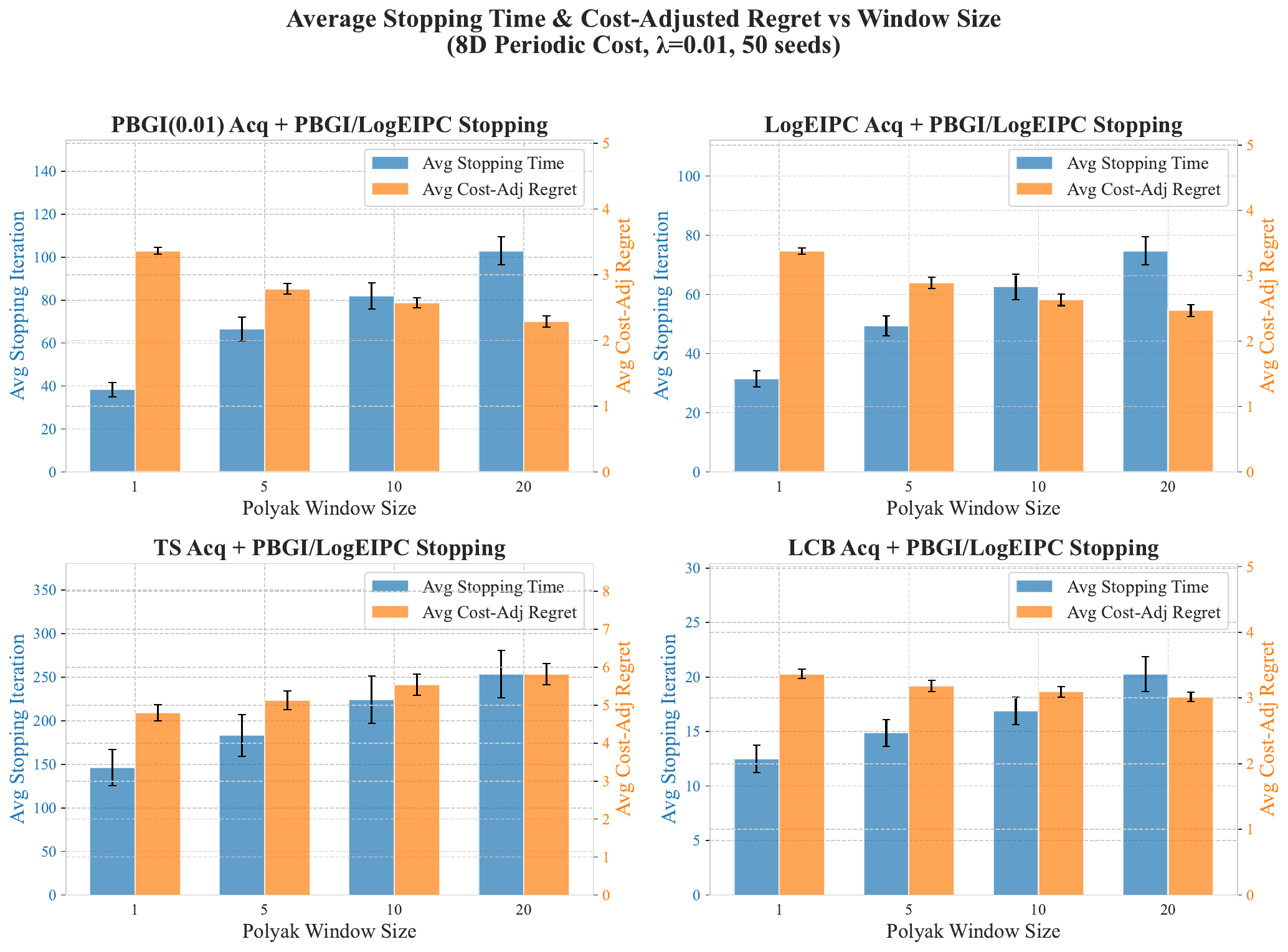}
    \caption{Effect of the moving-average window size $W \in \{1, 5, 10, 20\}$
    on the average stopping iteration (blue, left axis) and the average
    cost-adjusted simple regret (orange, right axis) for the PBGI/LogEIPC
    stopping rule on the 8D Bayesian regret benchmark under
    \emph{periodic} costs ($\lambda = 10^{-2}$, $50$ seeds, error bars are
    $\pm 2$~standard errors). Each subplot fixes one acquisition function.
    Smoothing yields the largest cost-adjusted regret reductions for the
    matched PBGI and LogEIPC pairs, a smaller reduction for LCB, and no
    benefit for TS.}
    \label{fig:ma_agg_periodic}
\end{figure}

\begin{figure}[h]
    \centering
    \includegraphics[width=\linewidth]{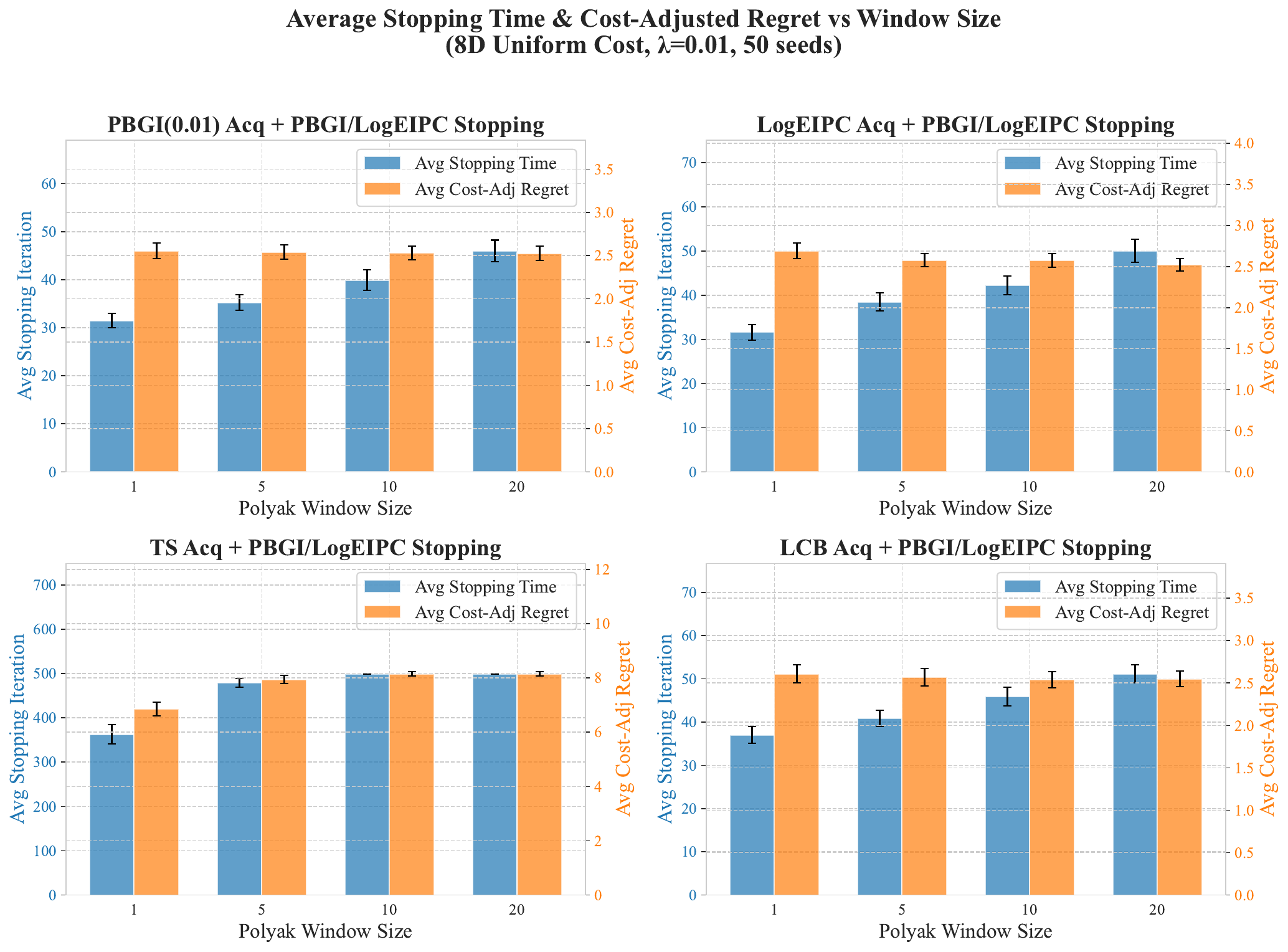}
    \caption{Aggregate ablation analogous to Figure~\ref{fig:ma_agg_periodic}
    under \emph{uniform} costs. Smoothing slightly reduces cost-adjusted
    regret for the matched PBGI and LogEIPC pairs and is roughly neutral
    for LCB and TS.}
    \label{fig:ma_agg_uniform}
\end{figure}

\begin{figure}[h]
    \centering
    \includegraphics[width=\linewidth]{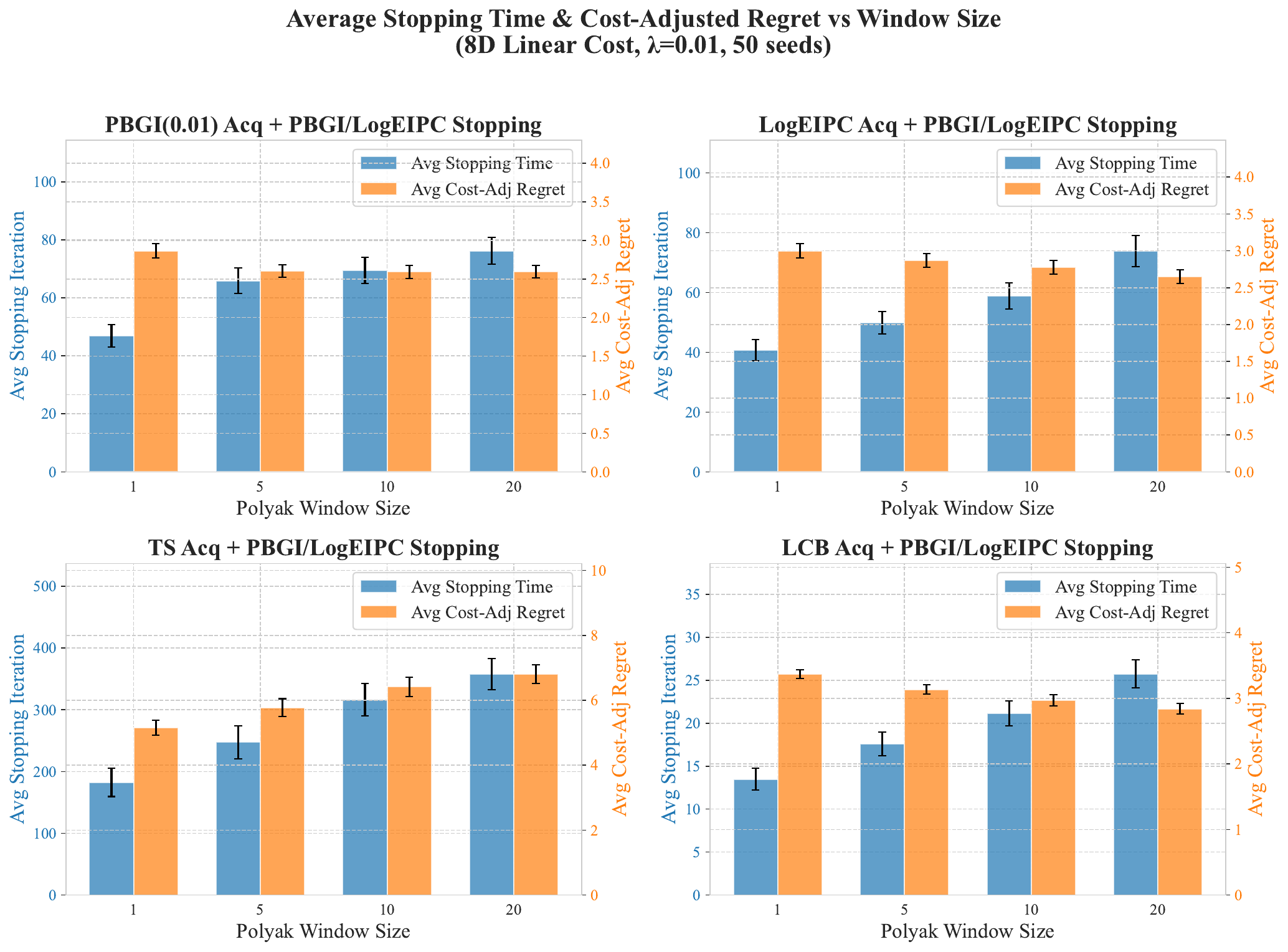}
    \caption{Aggregate ablation analogous to Figure~\ref{fig:ma_agg_periodic}
    under \emph{linear} costs. Smoothing reduces cost-adjusted regret for
    the matched PBGI and LogEIPC pairs and for LCB, with no benefit for TS.}
    \label{fig:ma_agg_linear}
\end{figure}

We ablate the window size $W \in \{1, 5, 10, 20\}$ for the PBGI/LogEIPC
stopping rule on the 8D Bayesian regret benchmark, where $W = 1$
corresponds to no smoothing. \Cref{fig:ma_single_pbgi,fig:ma_single_logeipc,fig:ma_single_ts,fig:ma_single_lcb} show how the window size affects the
stopping iteration and the resulting cost-adjusted regret along a single
acquisition-function path under each cost regime. Aggregate results
across 50 seeds for the PBGI/LogEIPC stopping rule paired with its
matched acquisition function under the three cost regimes (uniform,
linear, periodic) are summarized in \cref{fig:ma_agg_uniform,fig:ma_agg_linear,fig:ma_agg_periodic}.

Across all three cost regimes, applying a moving average consistently
improves cost-adjusted regret when the PBGI/LogEIPC stopping rule is
paired with its matched acquisition function, with the largest gains
observed under periodic costs.

\end{document}